%% file: main.tex
\documentclass[11pt,letterpaper]{article}
\usepackage{arxiv-format}

\usepackage{amsmath,amsfonts}
\usepackage{array}
\usepackage{textcomp}
\usepackage{stfloats}
\usepackage{url}
\usepackage{verbatim}
\usepackage{graphicx}
\usepackage{cite}

\usepackage{multirow}
\usepackage{booktabs} 
\usepackage{tabularx} 
\usepackage{array} 
\usepackage[table,dvipsnames]{xcolor} 

\definecolor{revisionblue}{RGB}{0,0,0}
\newcommand{\revise}[1]{{\color{revisionblue}#1}}
\usepackage{mathtools}
\usepackage{amsfonts}
\usepackage{hyperref} 
\usepackage{wrapfig}
\usepackage{caption} 

\usepackage{subcaption}
\usepackage[ruled,vlined]{algorithm2e}
\usepackage{float}

\usepackage{titlesec}                                               
\titleformat{\paragraph}[runin]                                     
{\normalfont\normalsize\bfseries}      
{}                     
{0pt}      
{}           
\titlespacing*{\paragraph}                                          
{0pt}                                          
{6pt}                                 
{0.6em}

\SetKwInput{KwIn}{Input}
\SetKwInput{KwOut}{Output}

\DeclareMathOperator*{\argmax}{arg\,max} %

\renewcommand{\Return}{\textbf{return}~}

\usepackage{graphicx}

\usepackage[most]{tcolorbox}
\newtcolorbox{stageonebox}{
    colback=blue!5,    
    colframe=blue!40!black, 
    boxrule=0.5pt,     
    arc=3mm,           
    left=1mm, right=1mm, top=1mm, bottom=1mm, 
    boxsep=1mm,      
    width=\dimexpr\linewidth-5em\relax,
}

\newtcolorbox{stagetwobox}{
    colback=teal!5,   
    colframe=teal!40!black, %
    boxrule=0.5pt,
    arc=3mm,
    left=1mm, right=1mm, top=1mm, bottom=1mm,
    boxsep=1mm,
    width=\dimexpr\linewidth-5em\relax,
}

\newtcolorbox{stageoneinline}{
    colback=blue!5,
    colframe=blue!40!black,
    boxrule=0.3pt,
    arc=3mm,                
    left=1mm, right=1mm, 
    top=0.5mm, bottom=0.5mm,
    boxsep=0.5mm,
    before skip=5pt, after skip=5pt,
}

\newtcolorbox{stagetwoinline}{
    colback=teal!5,
    colframe=teal!40!black,
    boxrule=0.3pt,
    arc=3mm,                
    left=1mm, right=1mm, 
    top=0.5mm, bottom=0.5mm,
    boxsep=0.5mm,
    before skip=5pt, after skip=5pt,
}

\newtcbox{\envname}{on line,
    colback=gray!8,
    colframe=gray!8,
    boxrule=0pt,
    arc=2pt,
    left=2pt,
    right=2pt,
    top=1pt,
    bottom=1pt,
    fontupper=\ttfamily\small
}

\usepackage[capitalize,noabbrev]{cleveref}

\usepackage{amsthm}
\theoremstyle{plain}
\newtheorem{theorem}{Theorem}

\theoremstyle{definition}
\newtheorem{definition}{Definition}

\theoremstyle{remark}

\newcolumntype{X}{c}
\RenewDocumentEnvironment{tabularx}{m m +b}{%
\resizebox{\textwidth}{!}{\begin{tabular}{#2}#3\end{tabular}}%
}{}
\usepackage{etoolbox}
\let\UniOMAoriginalgraphics\includegraphics
\renewcommand{\includegraphics}[2][]{%
  \ifstrequal{#2}{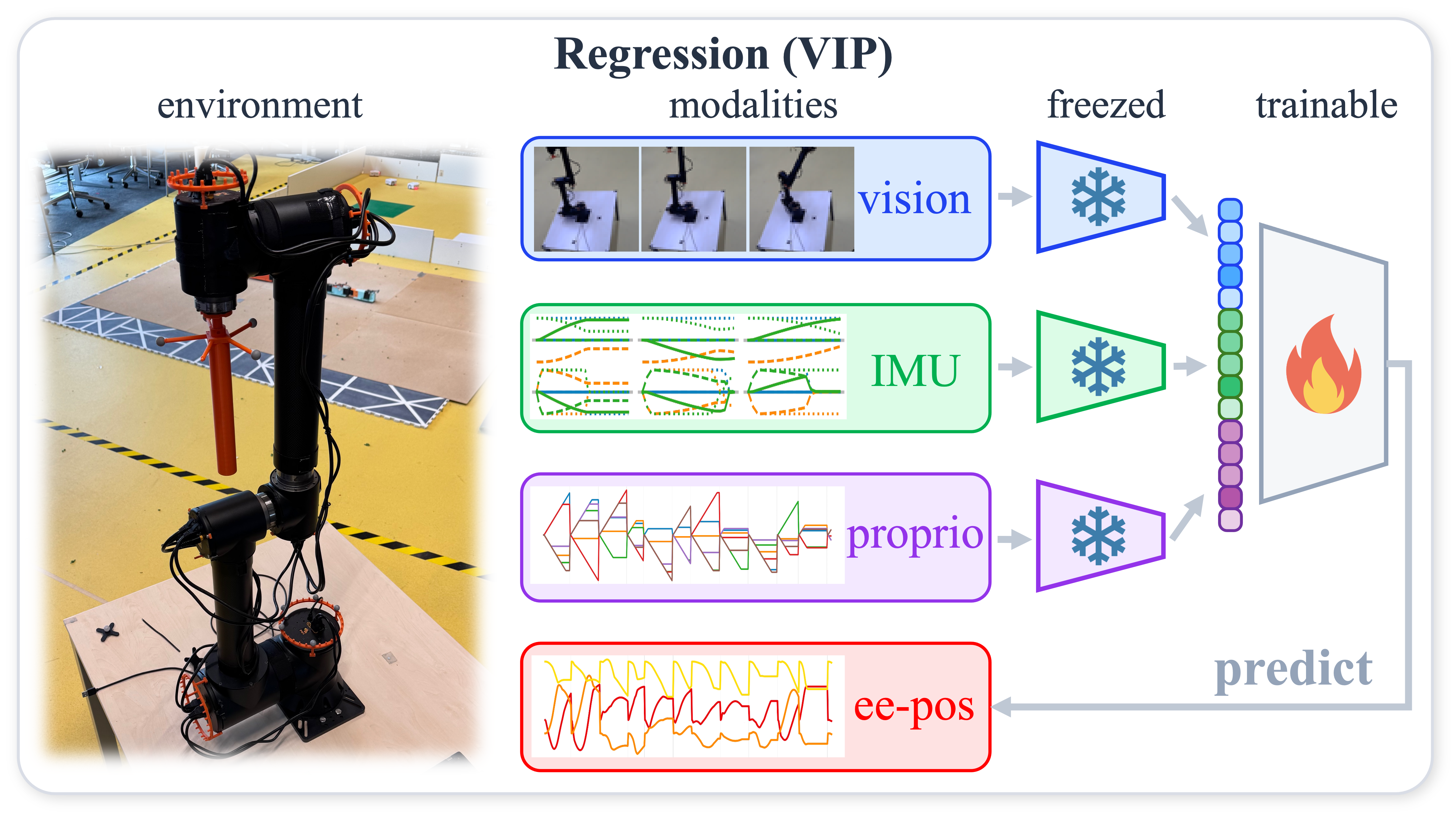}{%
    \UniOMAoriginalgraphics[width=0.68\linewidth]{#2}%
  }{\ifstrequal{#2}{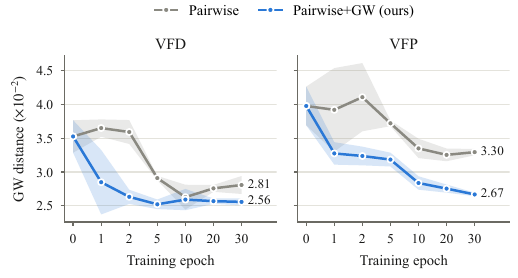}{%
    \UniOMAoriginalgraphics[width=0.65\linewidth]{#2}%
  }{\UniOMAoriginalgraphics[#1]{#2}}}%
}
\let\UniOMAoriginalresizebox\resizebox
\newcommand{\UniOMAtableresize}[3]{%
  \ifnum\value{table}=3\relax
    \UniOMAoriginalresizebox{0.624\linewidth}{#2}{#3}%
  \else
    \UniOMAoriginalresizebox{#1}{#2}{#3}%
  \fi
}
\AddToHook{env/table/begin}{\let\resizebox\UniOMAtableresize}

\usepackage{amssymb,thmtools,thm-restate}
\makeatletter
\newcommand{\StartSupplementAppendix}{%
  \setlength{\textfloatsep}{14pt plus 2pt minus 4pt}%
  \setcounter{section}{0}\setcounter{subsection}{0}\setcounter{subsubsection}{0}%
  \setcounter{equation}{0}\setcounter{figure}{0}\setcounter{table}{0}\setcounter{footnote}{0}%
  \renewcommand{\thesection}{\Roman{section}}%
  \renewcommand{\theHsection}{supp.\arabic{section}}%
  \renewcommand{\theHsubsection}{supp.\arabic{section}.\arabic{subsection}}%
  \renewcommand{\theHsubsubsection}{supp.\arabic{section}.\arabic{subsection}.\arabic{subsubsection}}%
  \renewcommand{\theHequation}{supp.\arabic{equation}}%
  \renewcommand{\theHfigure}{supp.\arabic{figure}}%
  \renewcommand{\theHtable}{supp.\arabic{table}}%
  \let\UniOMAtableresize\UniOMAoriginalresizebox
  \renewcommand{\envname}[1]{\texttt{##1}}%
  \renewcommand{\revise}[1]{{\leavevmode\color{revisionblue}##1}}%
  \def\title##1{\def\@title{##1}}%
  \def\author##1{\def\@author{##1}}%
  \def\maketitle{%
    \phantomsection\addcontentsline{toc}{section}{Appendix B: Supplementary Material}%
    \begin{center}\Large\bfseries Appendix B\par\medskip
    \large\@title\par\medskip\normalsize\normalfont\@author\par\end{center}\medskip}%
  \RenewDocumentEnvironment{tabular*}{m m +b}{%
    \begin{tabular}{@{}p{.22\linewidth}p{.43\linewidth}p{.29\linewidth}@{}}##3\end{tabular}%
  }{}%
}
\makeatother

\begin{document}
\input{main-body}

\clearpage
\phantomsection\addcontentsline{toc}{section}{References}
\input{main-references.bbl}

\clearpage
\begingroup
\renewcommand{\refname}{Supplementary References}
\phantomsection\addcontentsline{toc}{section}{Supplementary References}
\input{supplement-references.bbl}

\endgroup
\clearpage
\appendices

\input{appendix-main}

\clearpage
\begingroup
\StartSupplementAppendix
\input{appendix-supplement}

\clearpage
\endgroup
\end{document}

%% file: main-body.tex
\bstctlcite{IEEEtranBSTCTL:NoDash}

\title{UniOMA: Unified Optimal-Transport Multi-Modal Structural Alignment for Robot Perception}

\author{Xinrui Zu, Mark Hoogendoorn, Kevin Sebastian Luck, Shujian Yu\\[0.4em]\normalsize Vrije Universiteit Amsterdam, the Netherlands}


\markboth{IEEE Transactions on Robot Learning Submission}%
{Shell \MakeLowercase{\textit{et al.}}: A Sample Article Using IEEEtran.cls for IEEE Journals}


\maketitle

\begin{abstract}
Achieving generalizable and well-aligned multimodal representations remains a core challenge in robot learning. While recent approaches have attempted to align modalities by modeling conditional or higher-order statistical dependencies, they \revise{do not explicitly consider} the structural coherence across modalities. 
In this work, we propose a novel multimodal alignment method that augments existing contrastive losses with a geometry-aware Gromov--Wasserstein (GW) distance-based regularization. 
To this end, we encode intra-modal geometry for each modality and reduce cross-modal structural discrepancies through a dynamically learned GW barycenter, \revise{thereby encouraging structural consistency in addition to instance-level contrastive alignment}.
We apply this optimal-transport-based alignment strategy to robot perception tasks involving underexplored modalities such as force and tactile signals, where modality data often exhibit varying sample densities. 
Extensive experiments on five robotic benchmarks show that our method \revise{improves structural alignment and downstream performance} over state-of-the-art multimodal contrastive baselines across a wide range of downstream tasks, including robot and environment state prediction.
Moreover, our \revise{GW-based regularization can be} seamlessly integrated into most InfoNCE-like objectives.
\end{abstract}


\section{Introduction}
\label{sec:intro}

\IEEEPARstart{T}{he} integration of information from diverse sources or modalities has received increasing attention across a wide range of AI applications, including image/video/text generation \cite{Rombach_2022_CVPR,Mirza2014}, healthcare \cite{Acosta2022}, autonomous systems \cite{9000872}, scientific discovery \cite{Steyaert2023}, and robotics, particularly robot learning and perception \cite{lee2019making,Calandra2017}.
Recent advances in contrastive self-supervised learning (CSSL) \cite{he2020momentum,chen2020simple,10.5555/3495724.3497510,chen2021exploring}, particularly those leveraging InfoNCE losses \cite{Oord2018}, have shown strong performance in aligning heterogeneous modalities into a shared representation space \cite{radford2021learning}. 
Such alignment has enabled zero-shot cross-modal retrieval, transfer, generation, and completion \cite{radford2021learning,girdhar2023imagebind,chen2023vast,zhu2023languagebind,luo2022clip4clip}, while also providing a foundation for learning multimodal representations in robotic systems. 
In this setting, heterogeneous observations, such as visual, tactile, and proprioceptive modalities, are aligned for tasks like manipulation, imitation learning, reward learning, and perception \cite{Nair2022,karamcheti2023voltron,lee2019making}.
By maximizing agreement between paired modalities of the same instance while minimizing similarity between distinct instances, CSSL encourages the learning of invariant and semantically meaningful features.

Despite their empirical success, existing CSSL methods fundamentally cast multimodal alignment as a binary classification problem \cite{10.5555/3524938.3525859}: paired samples are encouraged to be close, while unpaired samples are pushed apart, without explicitly modeling the continuous intra-modal distance geometry. As a result, multimodal representations may appear statistically aligned at the population level, yet disagree in how samples relate to one another within each modality.
We refer to this phenomenon as the \emph{structural alignment gap} in contrast to the \emph{modality gap}~\cite{10.5555/3600270.3601550}, which characterizes the instance-level separation of modalities in the embedding space.

\begin{figure*}
    \centering
    \includegraphics[width=\textwidth]{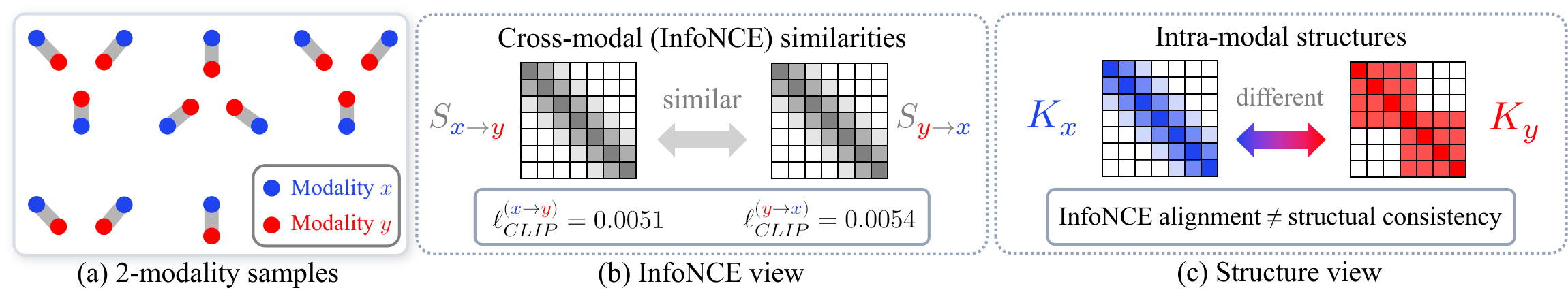}
    \caption{\textbf{Structural alignment gap.}
(a) Samples from two modalities (\textcolor{black}{$x$} and \textcolor{red}{$y$}) show strong instance-level correspondence, while the intra-modal structures are different;
(b) the cross-modal InfoNCE similarity matrices ($S_{x\to y}$ and $S_{y\to x}$, whose $(i,j)$-th element is $\text{sim}(x_i,y_j)$ and $\text{sim}(y_i,x_j)$ respectively) are similar and both lead to a small InfoNCE loss ($\ell_{\text{CLIP}}^{(x\to y)}$ and $\ell_{\text{CLIP}}^{(y\to x)}$ as defined in Eq.~\ref{eq:infonce}), indicating that existing contrastive alignment losses cannot distinguish modality-specific structural differences;
(c) the intra-modal similarity (kernel) structures differ ($K_x$ vs.\ $K_y$), where $K_y$ exhibits cluster structure absent from $K_x$.
Thus, alignment with InfoNCE loss does not imply structural consistency across modalities.}
    \label{fig:synthetic}
\end{figure*}

Crucially, the structural alignment gap is not an implementation artifact but a theoretical limitation of InfoNCE-style objectives. 
InfoNCE and related contrastive losses can be interpreted as variational lower bounds on mutual information \cite{Kraskov2004,poole2019variational}.
However, mutual information is invariant under certain invertible reparameterizations of the variables \cite{Cover2005,Kinney2014}, and is therefore underdetermined with respect to modality-specific geometry.
Consequently, while effective at capturing population-level statistical dependence, they \revise{do not by themselves constrain the intra-modal geometry to be consistent across modalities}. 

Fig.~\ref{fig:synthetic} illustrates this gap: the cross-modal InfoNCE similarities remain compatible, while the intra-modal kernels $K_x$ and $K_y$ (Sec.~\ref{subsec:k}) differ.
Thus, instance-level correspondence is preserved, yet the intra-modal relational geometry is inconsistent, which InfoNCE-like objectives alone do not penalize.

This \emph{structural alignment gap} is particularly critical in robotics, where multimodal sensor streams are neither i.i.d. nor structureless: trajectories form subclusters \cite{Sermanet2017TimeContrastiveNS}, contact events induce discontinuities \cite{pmlr-v211-guo23b}, and proprioceptive signals follow physical constraints \cite{Lee2020,10.5555/897831}. 
\revise{These robotic settings motivate us to test whether explicit relational alignment complements instance-level correspondence.}
To address the identified \emph{structural alignment gap}, we introduce a \textbf{Uni}fied \textbf{O}ptimal-transport \textbf{M}ulti-modal structural \textbf{A}lignment (\textbf{UniOMA}) framework that \revise{supports alignment across three or more modalities}.
UniOMA augments contrastive learning with a structure-aware regularization based on Gromov--Wasserstein (GW) distances and barycenters \cite{peyre2016gromov,Gong2022}. 
In our formulation, observations from each modality are represented as a metric space through intra-modal similarity matrices. 
A dynamic GW barycenter is then computed as the structural consensus across modalities, and each modality is softly aligned to this consensus by minimizing weighted GW distances.
To our knowledge, UniOMA is the first method to employ a GW barycenter as a structural-consensus regularizer for contrastive representation learning.
The modality weights are optimized end-to-end alongside encoder parameters, enabling adaptive contributions of different modalities to the structural consensus.
\revise{By aligning each modality to a shared structural consensus rather than every other modality, this barycentric formulation reduces structural couplings from $O(M^2)$ to $O(M)$, where $M$ is the number of modalities, and thus is well suited to settings with three or more modalities.}

In summary, our main contributions are:

\begin{itemize}
    \item[(C1)] We identify and formalize the structural alignment gap, demonstrating why InfoNCE-style objectives \revise{do not explicitly enforce intra-modal structural consistency}, supported by synthetic analysis.
    \item[(C2)] We propose UniOMA, a structure-aware multimodal alignment framework based on Gromov--Wasserstein distance and barycenters, which \revise{is well suited to settings with} 3+ modalities.
    \item[(C3)] We conduct extensive evaluations on five robotic benchmarks (including a new real-world robot-centric dataset with a 6-DoF manipulator), spanning heterogeneous modalities, 3+ modality settings, and complementary downstream tasks, showing that UniOMA \revise{improves downstream performance in most evaluated settings and achieves better structural alignment.}
\end{itemize}

\section{Background and Related Work}



\subsection{Alignment via InfoNCE and Extensions to More than Two Modalities}
\label{subsec:contrastive}
Unlike multimodal fusion \cite{10.5555/3454287.3454289,Li2019}, which typically requires all modalities to be present at inference, alignment into a shared embedding space remains functional even if some modalities are missing, enabling zero-shot retrieval, generation, and modality completion \cite{pmlr-v139-jia21b}. 
A representative example is CLIP \cite{radford2021learning}, which trains modality-specific encoders $f_{\theta}^{(1)}, f_{\theta}^{(2)}$ using an InfoNCE-style objective to align paired samples from two modalities.
Given a batch of $N$ paired samples $\{(\mathbf{x}_i^{(1)}, \mathbf{x}_i^{(2)})\}_{i=1}^{N}$, where $\mathbf{x}_i^{(m)}$ denotes the input of the $i$-th sample from modality $m$, each encoder maps its input to an embedding $\mathbf{z}_i^{(m)} = f_{\theta}^{(m)}(\mathbf{x}_i^{(m)})$.
CLIP then uses an InfoNCE-style objective to identify the correct cross-modal pair among $N$ candidates:
\begin{equation}
\ell_{\text{CLIP}}^{(1\to2)}(\theta) = -\frac{1}{N} \sum_{i=1}^{N}\log \frac{\exp\left(\text{sim}(\mathbf{{z}}_i^{(1)},\mathbf{{z}}_i^{(2)})/ \tau \right)}{\sum_{j=1}^N \exp\left( \text{sim}(\mathbf{{z}}_i^{(1)},\mathbf{{z}}_j^{(2)}) / \tau \right)},
\label{eq:infonce}
\end{equation}
where $\text{sim}(\cdot,\cdot)$ is the similarity between the embeddings $\mathbf{{z}}_i^{(m)}=f_{\theta}^{(m)}(\mathbf{x}_i^{(m)}),m=1,2$, and $\tau$ denotes a temperature parameter. The final CLIP objective symmetrizes Eq.~(\ref{eq:infonce}) by taking the average via
\begin{equation}
\mathcal{L}_{\text{CLIP}}^{(1,2)}(\theta) = \frac{1}{2}(\ell_{\text{CLIP}}^{(1\to2)}(\theta)+\ell_{\text{CLIP}}^{(2\to1)}(\theta)),
\label{eq:clip}
\end{equation}
where $\revise{\ell}_{\text{CLIP}}^{(2\to1)}$ is the reverse direction $2\to1$.
In general, this InfoNCE-based objective captures the statistical correlation, providing a lower bound on the mutual information \cite{Kraskov2004,poole2019variational} between the anchor modality $\mathcal{X}^{(1)}$ and modality $\mathcal{X}^{(2)}$
\begin{equation}
I(\mathcal{X}^{(1)};\mathcal{X}^{(2)})\ \ge\ \log N\ -\ 2\,\mathcal{L}_{\text{CLIP}}^{(1,2)}(\theta).
\label{eq:mi}
\end{equation}

Despite their empirical success, InfoNCE-like objectives discretize the continuous similarity structure into a binary positive–negative signal, which can result in embedding spaces that are statistically aligned across modalities while failing to preserve intra-modal geometric relationships.

Real-world systems, particularly in robotics, often involve three or more modalities.
Existing approaches typically extend CLIP to three modalities by summing all pairwise contrastive losses \cite{tian2020contrastive,girdhar2023imagebind,akbari2021vatt,chen2023vast,alayrac2020self,chen2021multimodal,liu2024valor,huang2023clover,mai2022hybrid,moon2022imu2clip,shvetsova2022everything,xue2022clip,guzhov2022audioclip}:
\begin{equation}
\mathcal{L}_{\text{CMC}}^{(1,2,3)}(\theta) = \mathcal{L}_{\text{CLIP}}^{(1,2)}(\theta)+\mathcal{L}_{\text{CLIP}}^{(1,3)}(\theta)+\mathcal{L}_{\text{CLIP}}^{(2,3)}(\theta).
\label{eq:cmc}
\end{equation}


\begin{figure*}[t]
\centering
\includegraphics[width=\textwidth]{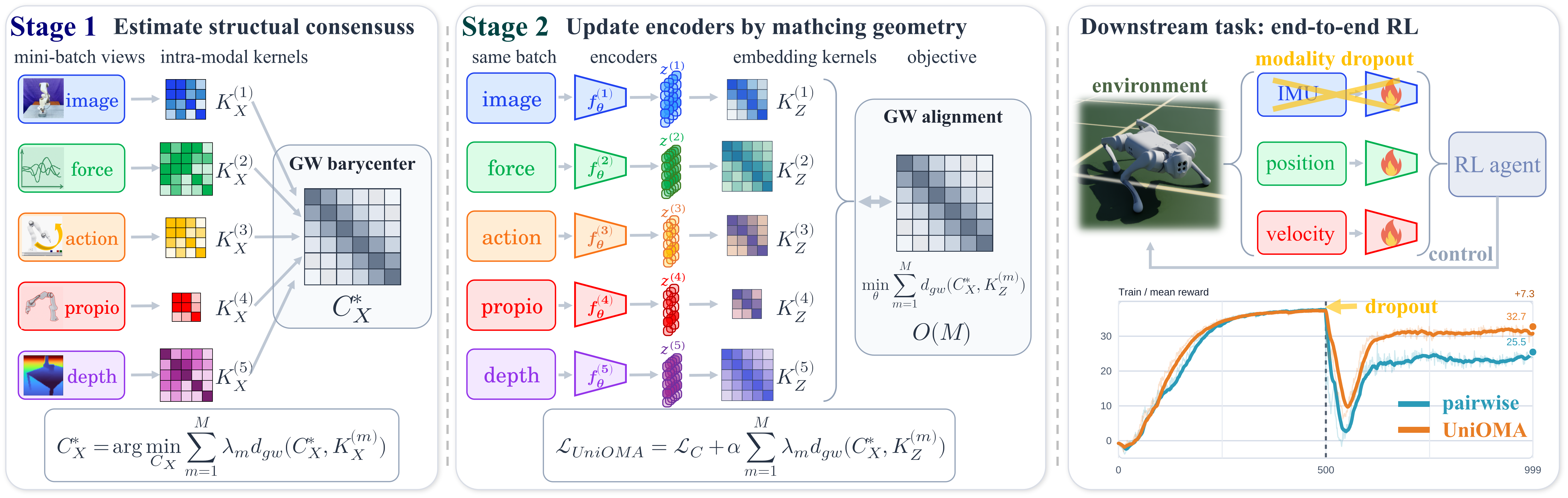}
\caption{\textbf{UniOMA in two stages.}
\emph{Stage 1} (left): given a mini-batch of heterogeneous observations, UniOMA builds one input-space similarity matrix \(\mathbf{K}_\mathbf{x}^{(m)}\) per modality and estimates a Gromov--Wasserstein (GW) barycenter \(\mathbf{C}_\mathbf{x}^{*}\), which serves as a batch-wise \emph{structural consensus}.
\emph{Stage 2} (middle): modality-specific encoders produce embeddings \(\mathbf{z}^{(m)}\) and the corresponding embedding-space kernels \(\mathbf{K}_\mathbf{z}^{(m)}\); the encoders are updated by matching each \(\mathbf{K}_\mathbf{z}^{(m)}\) to \(\mathbf{C}_\mathbf{x}^{*}\) through the UniOMA objective.
Aligning each modality to a single consensus avoids pairwise \(O(M^2)\) couplings\revise{, so the number of alignment terms grows linearly with \(M\)}.
\emph{Right}: UniOMA applied to the observation encoders of an end-to-end RL policy under modality dropout (Sec.~\ref{subsec:task} and Table~\ref{tab:go1_rl_robustness}).}
\label{fig:method}
\end{figure*}

Such pairwise extensions neglect higher-order dependencies among modalities. To address this issue, Symile \cite{saporta2024contrasting} formulates triple-wise contrastive objectives as:
\begin{equation}
\mathcal{L}_{\text{Symile}}^{(1,2,3)}(\theta) = \frac{1}{3}[\ell^{(1\to2,3)}(\theta)+\ell^{(2\to1,3)}(\theta)+\ell^{(3\to1,2)}(\theta)].
\label{eq:symile}
\end{equation}
Here, $\ell^{(1\to2,3)}$ is the InfoNCE-like loss for one positive triple and $N-1$ negative triples given by
\begin{equation}
\ell^{(1\to2,3)}(\theta) = 
- \frac{1}{N} \sum_{i=1}^{N}\log \frac{\exp(\langle \mathbf{z}^{(1)}_i, \mathbf{z}^{(2)}_i, \mathbf{z}^{(3)}_i \rangle / \tau)}{\sum_{j=1}^{N} \exp(\langle \mathbf{z}^{(1)}_i, \mathbf{z}^{(2)}_j, \mathbf{z}^{(3)}_j \rangle / \tau)},
\label{eq:symile-item}
\end{equation}
where each term $\ell^{(1\to2,3)}$ compares one positive triple against $N-1$ negatives, 
$\langle\cdot,\cdot,\cdot \rangle$ is the coordinate-wise sum of the element-wise product. 
More recently, GRAM \cite{cicchetti2024gramian} replaces the \revise{cosine} similarity with the Gramian volume $\text{Vol}(\cdot,\dots,\cdot)$ spanned by embeddings from modalities, providing a higher-order, groupwise compatibility score:
\[
\begin{alignedat}{1}
&\mathcal{L}_{\text{GRAM}}^{(1,\dots,M)}(\theta)
=
\frac{1}{2}\!\left(
\ell_{\text{D2A}}^{(1\to2,\dots,M)}(\theta)
+
\ell_{\text{A2D}}^{(1\to2,\dots,M)}(\theta)
\right),
\end{alignedat}
\]
\[
\scalebox{0.95}{$
\begin{alignedat}{1}
&\ell_{\text{D2A}}^{(1\to2,\dots,M)}(\theta) \\
&= 
-\frac{1}{N} \sum_{i=1}^{N} \log
\frac{
\exp\!\left(-\text{Vol}(\mathbf{z}^{(1)}_i,\mathbf{z}^{(2)}_i,\dots,\mathbf{z}^{(M)}_i)/\tau \right)
}{
\sum_{j=1}^{N}
\exp\!\left(-\text{Vol}(\mathbf{z}^{(1)}_j,\mathbf{z}^{(2)}_i,\dots,\mathbf{z}^{(M)}_i)/\tau \right)
},
\end{alignedat}
$}
\]
\[
\scalebox{0.95}{$
\begin{alignedat}{1}
&\ell_{\text{A2D}}^{(1\to2,\dots,M)}(\theta) \\
&= 
-\frac{1}{N} \sum_{i=1}^{N} \log
\frac{
\exp\!\left(-\text{Vol}(\mathbf{z}^{(1)}_i,\mathbf{z}^{(2)}_i,\dots,\mathbf{z}^{(M)}_i)/\tau \right)
}{
\sum_{j=1}^{N}
\exp\!\left(-\text{Vol}(\mathbf{z}^{(1)}_i,\mathbf{z}^{(2)}_j,\dots,\mathbf{z}^{(M)}_j)/\tau \right)
},
\end{alignedat}
$}
\]
where $\ell_{\text{D2A}}$ and $\ell_{\text{A2D}}$ are bidirectional cross-modal contrastive losses with modality~1 as the anchor (A):
The D2A loss matches the non-anchor modality group, consisting of modalities $2$ through $M$, to modality 1, whereas the A2D loss matches the modality 1 embedding to the non-anchor modality group.
$\text{Vol}(\cdot,\dots,\cdot)$ is the volume of the $M$-dimensional parallelotope formed by the embedding vectors $\mathbf{z}^{(m)}$, with smaller volumes corresponding to higher groupwise similarity.

These methods make progress on multimodal alignment but still remain limited to instance-level dependencies, overlooking intra-modal structure.
More advanced objectives (e.g., CoMM \cite{dufumier2024align} and TRIANGLE \cite{Cicchetti2025}) strengthen cross-modal matching, yet still do not explicitly align intra-modal relational geometry.
\cite{Zhu2024} introduces a standard optimal transport (OT) regularizer for cross-modal consistency; however, the matching is performed at the distributional level of the embeddings $\mathbf{z}$, thereby ignoring intra-modal structure in the observation space. 
In contrast, we align structures by encoding each modality by its intra-modal similarity (kernel) and enforcing a GW-based consensus in the input space $\mathcal{X}$, preserving centroid geometry rather than embedding-level OT matching.


\subsection{Multimodal Representation Learning in Robotics}

Robotics is inherently multimodal: vision, force–torque, tactile sensing, and proprioception provide complementary views of the robot–environment system. While multimodal representation learning has been extensively studied in vision–language settings, its exploration in robotics remains limited. Existing work, including the recent Vision–Language–Action (VLA) models, has primarily focused on modality fusion or transfer \cite{Shridhar2020,Brohan2022,10.5555/3618408.3618748,Kim2024,octo_2023,Intelligence2025}.

By comparison, alignment of robotic perception modalities into a shared representation space remains underexplored.
Recent efforts \cite{wojcik2024case,Dutta2024} demonstrate cross-modal retrieval and perception, while \cite{pmlr-v155-zambelli21a,Sermanet2017TimeContrastiveNS} show that cross-modal or cross-temporal alignment can yield transferable representations.
These developments highlight that robotic perception data is inherently structured, which is shaped by trajectories, contact events, and physical constraints, motivating alignment methods that preserve intra-modal geometry across modalities rather than relying solely on feature fusion.
Importantly, our work addresses a complementary setting to recent VLA models.
Rather than focusing on language-conditioned policy learning, we study multimodal alignment for robotic perception using physically grounded sensor modalities, such as force, tactile sensing, and proprioception, with the goal of learning structure-preserving multimodal representations.

\section{UniOMA: Unified Optimal-Transport Multi-Modal Structural Alignment}


Our proposed UniOMA aligns three or more heterogeneous modalities by combining instance-level statistical correspondence with structure-level geometric consistency. As illustrated in Fig.~\ref{fig:method}, UniOMA operates in two stages. Given a
mini-batch of multimodal observations, it first constructs modality-specific input-space kernels and estimates a GW barycenter as the batch-wise structural consensus. It then updates the modality-specific encoders by aligning their embedding-space kernels to this consensus while optimizing a contrastive objective.


\subsection{Problem Statement}

Consider a robot interacting with its environment over time. At each time step $t$, the robot receives a multimodal observation
$\mathbf{o}_t = (\mathbf{x}_t^{(1)}, \mathbf{x}_t^{(2)}, \ldots, \mathbf{x}_t^{(M)})$,
where $\mathbf{x}_t^{(m)} \in \mathcal{X}^{(m)}$ is the observation from modality $m$, such as vision, proprioception, tactile sensing, force--torque signals, or environment states. In some cases, the action $\mathbf{a}_t$ may also be included as one modality.
The goal of multimodal alignment is to learn modality-specific encoders
$f^{(m)}: \mathcal{X}^{(m)} \to \mathbb{R}^d,\ m=1,\ldots,M$,
that map each input $\mathbf{x}^{(m)}$ to a shared latent representation
$\mathbf{z}^{(m)}=f^{(m)}(\mathbf{x}^{(m)}) \in \mathbb{R}^d$.
For paired observations originating from the same underlying instance or physical event, the learned embeddings are expected to be semantically compatible and highly similar across modalities. At the same time, effective alignment should preserve the relational structure within each modality, so that the geometry among samples in one modality is consistently reflected in the shared representation space.


Fig.~\ref{fig:method} shows examples of such modalities and how UniOMA aligns them.

\subsection{Gromov--Wasserstein Distance}

The Gromov--Wasserstein (GW) distance \cite{peyre2016gromov,Gong2022} extends optimal transport (OT) \cite{villani2008optimal}, which seeks the minimum cost of transporting one probability measure into another within the same metric space, to measures supported on different metric spaces.
Since a cross-modal cost $c(\mathbf{x},\mathbf{y})$ is generally difficult to define, GW replaces it with a relational cost that only requires the two intra-modal similarity matrices.
\begin{definition}[Gromov--Wasserstein Distance]
Let $\mathcal{X}_{d_{\mathbf{x}},\mu}$ and $\mathcal{Y}_{d_{\mathbf{y}},\nu}$ be two metric–measure spaces, with distance metrics $d_{\mathbf{x}}$, $d_{\mathbf{y}}$ and probability measures $\mu,\nu$.
The GW distance between them is defined as:
\begin{equation}
\begin{aligned}
& d_{gw}(\mu,\nu)
:= \\
& \inf_{\pi \in \Pi(\mu,\nu)}
\int_{\mathcal{X}^2\times\mathcal{Y}^2}
c\!\left(
    d_{\mathbf{x}}(\mathbf{x},\mathbf{x}'),
    d_{\mathbf{y}}(\mathbf{y},\mathbf{y}')
\right)
\, d\pi(\mathbf{x},\mathbf{y})
\, d\pi(\mathbf{x}',\mathbf{y}')
\end{aligned}
\end{equation}
where $c\left(d_{\mathbf{x}}(\mathbf{x},\mathbf{x}'),d_{\mathbf{y}}(\mathbf{y},\mathbf{y}')\right)$ is a relational distance measuring the discrepancy between the sample pairs $(\mathbf{x},\mathbf{x}')$ and $(\mathbf{y},\mathbf{y}')$.
\label{def:gw}
\end{definition}

Minimizing the GW distance thus matches the relational geometry of two modalities rather than their raw coordinates.


For discrete samples, the continuous GW objective can be estimated from finite-sample intra-modal similarity matrices. 
\revise{We use the squared relational loss on symmetric similarity kernels and identify it as GW distance.}

\begin{theorem}[Kernelized empirical GW distance~\cite{Gong2022}]
Let $\mathcal{X}=\{x_i\}_{i=1}^{I}$ and $\mathcal{Y}=\{y_j\}_{j=1}^{J}$ be two sets of samples
with empirical measures \revise{$\hat{\mathbf{p}}_{\mathbf{x}}=\frac{1}{I}\mathbf{1}_I$} and \revise{$\hat{\mathbf{p}}_{\mathbf{y}}=\frac{1}{J}\mathbf{1}_J$}.
Let \revise{$\mathbf{K}_{\mathbf{x}}\in\mathbb R^{I\times I}$} and \revise{$\mathbf{K}_{\mathbf{y}}\in\mathbb R^{J\times J}$} denote their
intra-modal similarity matrices. 
The empirical GW distance between the two sample sets is defined as:
\[
\revise{\begin{aligned}
\hat d_{gw}(\mathbf K_{\mathbf x},\mathbf K_{\mathbf y})
&:=\min_{\mathbf T\in\Pi(\hat{\mathbf p}_{\mathbf x},\hat{\mathbf p}_{\mathbf y})}\;\\
&\quad\sum_{i,i',j,j'}\bigl((\mathbf K_{\mathbf x})_{ii'}-(\mathbf K_{\mathbf y})_{jj'}\bigr)^2 T_{ji}T_{j'i'}.
\end{aligned}}
\]
where $\mathbf{T}$ is the doubly-stochastic matrix that models the transport plan coupling the two empirical measures.
\label{thm:empirical-gw}
\end{theorem}
In practice, we estimate the optimal transport plan $\mathbf{T}^*$ via iterative OT solvers (see Alg.~\ref{alg:ot_estimation} in Appendix~\ref{app:implementation} for details) and compute the final GW distance between $\{\mathbf{x}_i\}_{i=1}^I$ and $\{\mathbf{y}_j\}_{j=1}^J$ as:
\begin{equation}
\label{eq:gw}
\revise{\begin{aligned}
\hat d_{gw}(\mathbf K_{\mathbf x},\mathbf K_{\mathbf y})
={}&\hat{\mathbf p}_{\mathbf x}^{\top}(\mathbf K_{\mathbf x}\odot\mathbf K_{\mathbf x})\hat{\mathbf p}_{\mathbf x}\\
&+\hat{\mathbf p}_{\mathbf y}^{\top}(\mathbf K_{\mathbf y}\odot\mathbf K_{\mathbf y})\hat{\mathbf p}_{\mathbf y}\\
&-2
\mathrm{tr}(\mathbf K_{\mathbf x}^{\top}\mathbf T^{*\top}\mathbf K_{\mathbf y}\mathbf T^*),
\end{aligned}}
\end{equation}
\revise{where $\odot$ denotes element-wise multiplication.}
\revise{The first two terms are independent of $\mathbf T$ only when the kernels and marginals are fixed. Thus minimizing the GW distance over $\mathbf T$ is equivalent to maximizing the trace term in Eq.~\ref{eq:gw}. During encoder optimization, we keep the embedding-dependent quadratic term.}
Eq.~(\ref{eq:gw}) enables cross-modal alignment directly from intra-modal similarity structures, without the need for an explicit cross-modal cost function or extra neural potential models \cite{korotin2022neural,korotin2022kernel,arjovsky2017wasserstein}.

\subsection{Structural Consensus}
\label{subsec:k}

To preserve intra-modal structure during alignment,  we represent each modality $\mathcal{X}^{(m)}$ by an intra-modal similarity matrix $\mathbf{K}_{\mathbf{x}}^{(m)}\in\mathbb{R}^{N_m \times N_m}$, where $(\mathbf{K}^{(m)}_{\mathbf{x}})_{ij}=\text{sim}(\mathbf{x}_i^{(m)},\mathbf{x}_j^{(m)})$ encodes the pairwise similarity between samples and $N_m$ is the number of observations from modality $m$, which can differ across modalities. 
Such kernel matrices provide a unified representation of relational structure across heterogeneous modalities, abstracting away differences in raw input dimensionality or temporal resolution.


We construct $\mathbf{K}_{\mathbf{x}}^{(m)}$ using modality-specific similarity functions. 
For high-dimensional modalities such as image tensors, we first extract feature representations with a pretrained encoder and then compute pairwise similarities in the resulting feature space, alleviating the sparsity and dimensionality issues associated with raw inputs.
For sequential or time-series modalities such as force--torque signals, we use the time-series clustering kernel (TCK; \cite{Mikalsen2018}) to more effectively capture temporal dynamics.
Additional implementation details are provided in Supplementary Material, Sec.~I.




The goal is to identify a shared structural consensus that captures the common relational patterns across modalities. \revise{We assume that the modalities share these relational patterns only partially, for example through common trajectories and contact events, rather than having identical geometry; the consensus represents the shared part, and the contrastive term retains modality-specific information.}
We define this consensus as a Gromov--Wasserstein barycenter~\cite{Gong2022} of the intra-modal similarity matrices.

\begin{definition}[Structural Consensus of Multimodal Data]
Given intra-modal kernel matrices $\{\mathbf{K}^{(m)}_{\mathbf{x}}\}_{m=1}^M$, the structural consensus is defined as the barycenter:
\begin{equation}
\mathbf{C}^*_{\mathbf{x}}=\mathrm{arg}\min_{\mathbf{C}_{\mathbf{x}}\in\mathcal{M}}\sum_{m=1}^M\lambda_m \cdot {\hat d_{gw}}(\mathbf{C}_{\mathbf{x}},\mathbf{K}^{(m)}_{\mathbf{x}}),
\end{equation}
where $\mathcal{M}$ denotes the space of symmetric positive definite (SPD) matrices, $\hat d_{gw}$ is the \revise{empirical} GW distance (Eq.~\ref{eq:gw}), and $\{\lambda_m\}$ are learnable modality weights satisfying $\lambda_m\geq0$ and $\sum_m \lambda_m=1$.
\label{def:bc}
\end{definition}

In practice, $\mathbf{C}^{*}_{\mathbf{x}}$ is estimated by an iterative barycenter optimization
scheme (Alg.~\ref{alg:bc} in Appendix~\ref{app:implementation}). 
The central idea of UniOMA is to use this batch-wise consensus as a structural anchor: during training, each modality is aligned by minimizing the GW distance between its embedding-induced kernel \revise{$\mathbf K^{(m)}_{\mathbf z}$, which is the embedding-space similarity matrix of $\mathbf z^{(m)}=f_{\theta}^{(m)}(\mathbf x^{(m)})$}, and the consensus \revise{$\mathbf C^*_{\mathbf x}$}.
\revise{The weights determine each modality's contribution to the consensus and serve as an empirical indicator of modality-wise agreement with the learned structural consensus, rather than a direct measure of modality importance.}
\revise{Detailed implementations are in the next section.}


\subsection{UniOMA Objective and Algorithm}

Given the batch-wise structural consensus $\mathbf{C}^*_{\mathbf{x}}$ in Sec.~\ref{subsec:k}, UniOMA augments a standard contrastive term $\mathcal{L}_{\text{c}}$ with a structure-aware regularizer
\begin{equation}
\mathcal{L}_{\text{UniOMA}}(\theta) = \mathcal{L}_{\text{c}}(\theta) +  \alpha \sum_{m=1}^M \lambda_m \cdot \revise{\hat d_{gw}}(\mathbf{C}^*_{\mathbf{x}}, \mathbf{K}^{(m)}_{\mathbf{z}}),
\label{eq:UniOMA_loss}
\end{equation}
where $\mathbf{K}^{(m)}_{\mathbf{z}}$ is the embedding-space similarity matrix of $\mathbf{z}^{(m)}=f_{\theta}^{(m)}(\mathbf{x}^{(m)})$.
\revise{With the plan $\mathbf T^{(m)*}$ of Alg.~\ref{alg:ot_estimation} held fixed and $\hat{\mathbf p}=\frac{1}{N}\mathbf 1_N$, the structural term of modality $m$ is}
\begin{equation}
\label{eq:structural-term}
\revise{\begin{aligned}
\hat d_{gw}(\mathbf C^*_{\mathbf x},\mathbf K^{(m)}_{\mathbf z}(\theta))
={}&\hat{\mathbf p}^{\top}(\mathbf C^*_{\mathbf x}\odot\mathbf C^*_{\mathbf x})\hat{\mathbf p}\\
&+\hat{\mathbf p}^{\top}(\mathbf K^{(m)}_{\mathbf z}(\theta)\odot\mathbf K^{(m)}_{\mathbf z}(\theta))\hat{\mathbf p}\\
&-2\,\mathrm{tr}\big(\mathbf C^{*\top}_{\mathbf x}\mathbf T^{(m)*\top}\mathbf K^{(m)}_{\mathbf z}(\theta)\mathbf T^{(m)*}\big),
\end{aligned}}
\end{equation}
\revise{where the first term is constant and the gradients change $\theta$ through $\mathbf K^{(m)}_{\mathbf z}$ in the second and third terms.}
\revise{We add an entropy term $\beta\sum_{m=1}^M\lambda_m\log(\lambda_m+\varepsilon)$ with $\beta=1$ to discourage the concentration of the weights on a single modality.}
The scalar \( \alpha \) balances contrastive discrimination and structural coherence, and the learnable weights \( \{\lambda_m\} \) \revise{determine} each modality’s contribution to the consensus.
Implementation details for estimating \( \mathbf{C}^{*}_{\mathbf{x}} \) and evaluating \( \revise{\hat d_{gw}}(\cdot,\cdot) \) are in Appendix~\ref{app:implementation} (see also Fig.~\ref{fig:method}).

\begin{algorithm}[htbp]
\small
\setlength{\algomargin}{0.6em}
\SetAlgoNlRelativeSize{-1}
\setlength{\interspacetitleruled}{2pt} 
\setlength{\algotitleheightrule}{0.2pt}
\setlength{\algomargin}{0.6em}
    \caption{UniOMA Training($\{\mathcal{X}^{(m)}\}_{m=1}^{M}, \revise{\eta}, \alpha$)}
    \label{alg:UniOMA}
    
    \KwIn{Multimodal dataset $\{\mathcal{X}^{(m)}\}_{m=1}^{M}$, learning rate $\revise{\eta}$, structural weight $\alpha$}
    
    Initialize $\{f^{(m)}(\cdot)\}_{m=1}^M$, modality weights $\{\lambda_m\}_{m=1}^M$\;
    
    \While{not converged}{
        \begin{stageonebox}
        \tcp{\textcolor{black}{Stage 1: consensus estimation}} 
        Sample batch $\{\mathbf{x}_i^{(m)}\}_{i=1}^{N_m}$ for modalities $\{\mathcal{X}^{(m)}\}_{m=1}^{M}$\;

        \For{$m \leftarrow 1$ \KwTo $M$}{ 
            Compute $\mathbf{K}^{(m)}_{\mathbf{x}}$ for the batch $\{\mathbf{x}_i^{(m)}\}_{i=1}^{N_m}$\;
        } 
        Estimate the structural consensus $\mathbf{C}^*_{\mathbf{x}}$ via Alg.~\ref{alg:bc}\;
        \end{stageonebox}

        \begin{stagetwobox}
        \tcp{\textcolor{teal}{Stage 2: alignment update}} 
        $\mathbf{z}_i^{(m)} \leftarrow f_{\theta}^{(m)}(\mathbf{x}_i^{(m)})$ for all $i, m$\;

        \For{$m \leftarrow 1$ \KwTo $M$}{
            ${\mathbf{T}^{(m)}}^*\leftarrow\texttt{OTPlan}(\mathbf{C}^*_{\mathbf{x}}, \mathbf{K}^{(m)}_{\mathbf{z}})$ via Alg. \ref{alg:ot_estimation}\;

            Compute $\hat{d}_{gw}(\mathbf{C}^*_{\mathbf{x}},\mathbf{K}^{(m)}_{\mathbf{z}})$ via Eq.~\revise{(\ref{eq:structural-term})};
        }
        
        Compute the contrastive learning loss $\mathcal{L}_{\text{c}}$;
        

        \revise{Compute $\mathcal L_{\text{UniOMA}}$ via Eq.~\eqref{eq:UniOMA_loss} and add the entropy term}\;
    
        $\theta \leftarrow \theta - \revise{\eta}\nabla_{\theta}\mathcal{L}_{\text{UniOMA}}$\;
        
        $\lambda_m \leftarrow \lambda_m - \revise{\eta}\nabla_{\lambda_m}\mathcal{L}_{\text{UniOMA}}$ for $m=1,\dots,M$\;
        \end{stagetwobox}
    }
\KwRet{$\{f^{(m)}_{\theta}\}_{m=1}^M, \{\lambda_m\}_{m=1}^M$}
\end{algorithm}

\paragraph{Why this design?}
(1) \textbf{scalable} to \(M\geq3\). Aligning every modality to one consensus avoids \(O(M^2)\) pairwise couplings.
(2) \textbf{flexible} to heterogeneous and asynchronous modalities. GW distance~\cite{peyre2016gromov} supports different dimensionalities and unequal sample counts across modalities.
This is particularly advantageous in robot perception, and we empirically validate \revise{this property} in Sec.~\ref{sec:ablation}.
(3) \textbf{Interpretable}. \revise{The modality weights $\{\lambda_m\}$ serve as an empirical indicator of modality-wise agreement with the learned structural consensus (Sec.~\ref{sec:weights}).}


The overall training procedure is in Alg.~\ref{alg:UniOMA}. At each iteration, UniOMA first estimates a batch-wise structural consensus $\mathbf{C}^*_{\mathbf{x}}$ from modality-specific input-space kernels, and then updates the encoders by aligning the embedding-induced kernels $\{\mathbf{K}^{(m)}_{\mathbf{z}}\}_{m=1}^M$ to this consensus while optimizing the contrastive objective. 
Both the encoder parameters $\theta$ and the modality weights $\{\lambda_m\}_{m=1}^M$ are updated by stochastic gradient descent.






A shared consensus could suppress information from a more informative modality; probes on the frozen encoders show that UniOMA does not reduce the task-relevant information of the information-rich modality and can increase that of the joint representation (Supplementary Material, Sec.~III-G).

\section{Experiments}

\begin{figure*}[ht!]
    \centering
    \includegraphics[width=\textwidth]{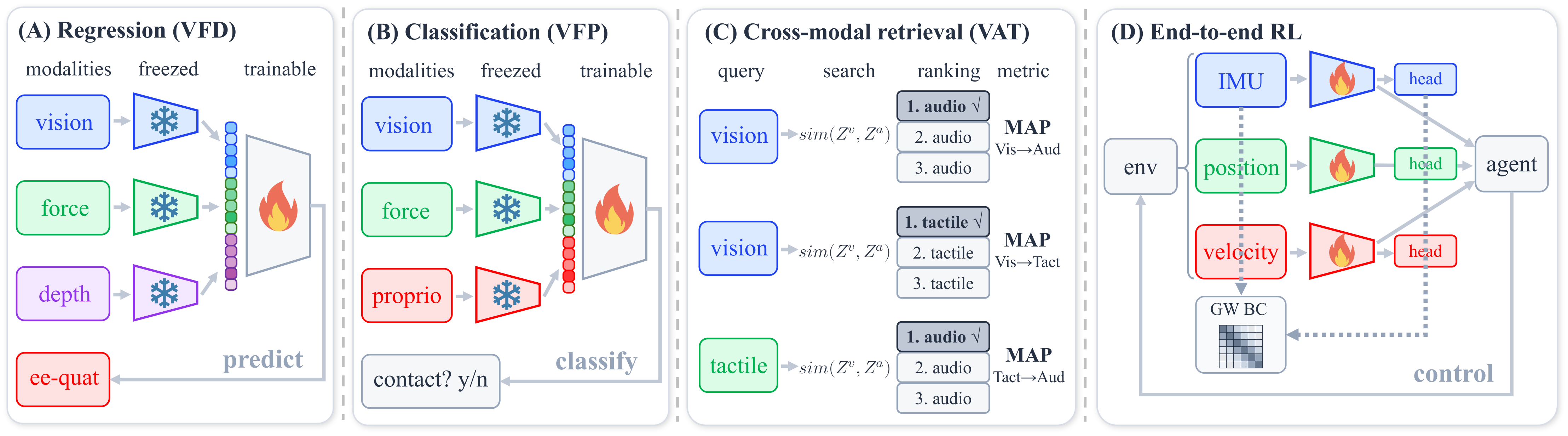}
\caption{\textbf{Downstream tasks.}
(A) \emph{Regression} (VFD): a head on the concatenated frozen embeddings predicts the next-step end-effector orientation.
(B) \emph{Classification} (VFP): a classifier on the frozen embeddings predicts binary contact.
(C) \emph{Cross-modal retrieval} (VAT): a query from one modality ranks items from another by embedding similarity (MAP for Vis\(\rightarrow\)Aud, Vis\(\rightarrow\)Tact, and Tact\(\rightarrow\)Aud).
(D) \emph{End-to-end RL}: the encoders are trained within the policy while UniOMA aligns their embeddings, and modalities are dropped to test robustness.}
    \label{fig:tasks}
\end{figure*}

We evaluate UniOMA on five robot-perception benchmarks with different sensor combinations: VFD (Vision--Force--Depth) and VFP (Vision--Force--Proprioception) from the Vision\&Touch dataset~\cite{lee2019making,Liang2021}, MuJoCo Push~\cite{Lee2020,Todorov2012}, VAT (Vision--Audio--Tactile) from ObjectFolder~2.0~\cite{gao2022objectfolder,wojcik2024case}, and our real-world VIP (Vision--IMU--Proprioception) dataset.
To assess scalability beyond three modalities, we further include a 4--7 modality classification task on the Vision\&Touch dataset (Sec.~\ref{sec:efficiency}).
All methods share the same backbones and training schedule (Supplementary Material, Sec.~I).

\subsection{Tasks and Datasets}
\label{subsec:task}
As illustrated in Fig.~\ref{fig:tasks}, the evaluation covers four downstream tasks: regression, classification, cross-modal retrieval, and end-to-end reinforcement learning (RL) under modality dropout.

\paragraph{VFD (Vision--Force--Depth).}
We evaluate two tasks on VFD: 
(1) next-step end-effector orientation prediction (regression, 4D), and 
(2) trajectory-pair discrimination (classification, binary). 
For both tasks, inputs are third-person RGB ($[b{\times}3{\times}128{\times}128]$), force--torque histories ($[b{\times}32{\times}6]$), and depth ($[b{\times}1{\times}128{\times}128]$). 
For trajectory-pair discrimination, given a pair of triplets (vision--force--depth), the model identifies whether the pair is from the same trajectory. 

\paragraph{VFP (Vision--Force--Proprioception).}
We evaluate next-step contact prediction (classification, binary). 
Inputs are RGB ($[b{\times}3{\times}128{\times}128]$), force--torque histories ($[b{\times}32{\times}6]$), and end-effector pose ($[b{\times}7]$). 
The task is to classify whether the end-effector is in contact with the object.

\paragraph{MuJoCo Push.}
We evaluate next-step object-position prediction in a planar pushing task where a Franka Emika Panda arm pushes a puck. 
Inputs are low-resolution grayscale images ($[b{\times}1{\times}32{\times}32]$), current force--torque with the contact flag ($[b{\times}7]$), and end-effector position ($[b{\times}3]$). 
The task is to predict the object's next-step 2D position on the table.

\paragraph{VAT (Vision--Audio--Tactile).}
We evaluate cross-modal retrieval using mean average precision (MAP). 
The dataset provides per-object visual, sound, and tactile observations. 
Queries and retrievals are built across $\{\text{Vis},\text{Aud},\text{Tact}\}$; we report direction-specific MAP, e.g., Vis$\to$Tact.

\begin{table*}[t]
\centering
\caption{Comparative results on downstream tasks (regression, classification, and cross-modal retrieval). Performance is measured by MSE ($\downarrow$), Acc.\ $(\%\!\uparrow)$, and MAP $(\uparrow)$ (mean $\pm$ std over 10 seeds\revise{; VFD and VFP use a trajectory-level 80/10/10 training/validation/test split}). Arrows denote retrieval direction. \colorbox{gray!10}{Gray rows} correspond to our method (UniOMA). \revise{Adding the GW term improves its contrastive baseline in most paired comparisons}, and all methods achieving the best performance for each task are UniOMA variants (highlighted in \textcolor{Bittersweet}{brown}).}
\vspace{2pt}
\label{tab:vat_results}
\setlength{\tabcolsep}{3pt}
\renewcommand{\arraystretch}{1.1}
\begin{tabularx}{\textwidth}{@{}l*{7}{>{\centering\arraybackslash}X}@{}}
\toprule
 & \multicolumn{2}{c}{Regression (MSE) $\downarrow$} & \multicolumn{2}{c}{Classification (\%) $\uparrow$} & \multicolumn{3}{c}{VAT retrieval (MAP) $\uparrow$} \\
\cmidrule(lr){2-3}\cmidrule(lr){4-5}\cmidrule(l){6-8}
Method & \mbox{V\&F\&D\,{\scriptsize($\times10^{-3}$)}} & MuJoCo & V\&F\&D & V\&F\&P & Vis$\rightarrow$Aud & Vis$\rightarrow$Tact & Tact$\rightarrow$Aud \\
\midrule
\multicolumn{8}{@{}l}{\textit{\revise{Methods with and without our GW regularizer}}}\\
\addlinespace[1pt]
Pairwise{\scriptsize~\cite{tian2020contrastive}}     & \textcolor{revisionblue}{1.34{\scriptsize$\pm$0.17}} & 0.44{\scriptsize$\pm$0.07} & \textcolor{revisionblue}{89.01{\scriptsize$\pm$0.04}} & \textcolor{revisionblue}{94.29{\scriptsize$\pm$0.02}} & 0.25{\scriptsize$\pm$0.07} & 0.41{\scriptsize$\pm$0.11} & 0.10{\scriptsize$\pm$0.01} \\
\rowcolor{gray!10}
Pairwise+GW{\scriptsize (ours)}  & \textcolor{Bittersweet}{\textbf{1.26}{\scriptsize$\pm$0.11}} & \textbf{0.38}{\scriptsize$\pm$0.09} & \textcolor{revisionblue}{\textbf{91.50}{\scriptsize$\pm$0.02}} & \textcolor{Bittersweet}{\textbf{94.71}{\scriptsize$\pm$0.02}} & \textbf{0.36}{\scriptsize$\pm$0.05} & \textcolor{Bittersweet}{\textbf{0.60}{\scriptsize$\pm$0.03}} & \textbf{0.12}{\scriptsize$\pm$0.02} \\
\addlinespace[3pt]
Symile{\scriptsize~\cite{saporta2024contrasting}}       & \textcolor{revisionblue}{2.89{\scriptsize$\pm$0.08}} & 0.28{\scriptsize$\pm$0.04} & \textcolor{revisionblue}{89.28{\scriptsize$\pm$0.05}} & \textcolor{revisionblue}{93.65{\scriptsize$\pm$0.05}} & 0.10{\scriptsize$\pm$0.02} & \textbf{0.21}{\scriptsize$\pm$0.05} & 0.08{\scriptsize$\pm$0.01} \\
\rowcolor{gray!10}
Symile+GW{\scriptsize (ours)}    & \textcolor{revisionblue}{\textbf{2.30}{\scriptsize$\pm$0.07}} & \textcolor{Bittersweet}{\textbf{0.23}{\scriptsize$\pm$0.02}} & \textcolor{revisionblue}{\textbf{91.93}{\scriptsize$\pm$0.02}} & \textcolor{revisionblue}{\textbf{93.72}{\scriptsize$\pm$0.03}} & \textbf{0.13}{\scriptsize$\pm$0.03} & 0.15{\scriptsize$\pm$0.03} & \textbf{0.14}{\scriptsize$\pm$0.03} \\
\addlinespace[3pt]
GRAM{\scriptsize~\cite{cicchetti2024gramian}}         & \textcolor{revisionblue}{3.43{\scriptsize$\pm$0.12}} & 0.52{\scriptsize$\pm$0.07} & \textcolor{revisionblue}{92.09{\scriptsize$\pm$0.03}} & \textcolor{revisionblue}{93.48{\scriptsize$\pm$0.04}} & 0.13{\scriptsize$\pm$0.02} & 0.34{\scriptsize$\pm$0.05} & 0.15{\scriptsize$\pm$0.01} \\
\rowcolor{gray!10}
GRAM+GW{\scriptsize (ours)}      & \textcolor{revisionblue}{\textbf{2.33}{\scriptsize$\pm$0.05}} & \textbf{0.30}{\scriptsize$\pm$0.06} & \textcolor{Bittersweet}{\textbf{92.91}{\scriptsize$\pm$0.03}} & \textcolor{revisionblue}{\textbf{93.80}{\scriptsize$\pm$0.03}} & \textcolor{Bittersweet}{\textbf{0.79}{\scriptsize$\pm$0.10}} & \textbf{0.58}{\scriptsize$\pm$0.04} & \textcolor{Bittersweet}{\textbf{0.16}{\scriptsize$\pm$0.01}} \\
\midrule
\multicolumn{8}{@{}l}{\textit{\revise{Independent methods (different architectures or objectives)}}}\\
\addlinespace[1pt]
CoMM{\scriptsize~\cite{dufumier2024align}}         & \textcolor{revisionblue}{1.75{\scriptsize$\pm$0.05}} & 0.26{\scriptsize$\pm$0.04} & \textcolor{revisionblue}{92.06{\scriptsize$\pm$0.01}} & \textcolor{revisionblue}{94.09{\scriptsize$\pm$0.02}} & 0.21{\scriptsize$\pm$0.10} & 0.40{\scriptsize$\pm$0.15} & 0.08{\scriptsize$\pm$0.03} \\
OTLC{\scriptsize~\cite{Zhu2024}}     & \textcolor{revisionblue}{1.44{\scriptsize$\pm$0.12}} & 0.40{\scriptsize$\pm$0.07} & \textcolor{revisionblue}{92.26{\scriptsize$\pm$0.02}} & \textcolor{revisionblue}{94.29{\scriptsize$\pm$0.03}} & 0.37{\scriptsize$\pm$0.05} & 0.58{\scriptsize$\pm$0.04} & 0.09{\scriptsize$\pm$0.01} \\
TRIANGLE{\scriptsize~\cite{Cicchetti2025}}         & \textcolor{revisionblue}{3.77{\scriptsize$\pm$0.11}} & 0.41{\scriptsize$\pm$0.06} & \textcolor{revisionblue}{92.89{\scriptsize$\pm$0.03}} & \textcolor{revisionblue}{93.76{\scriptsize$\pm$0.04}} & 0.17{\scriptsize$\pm$0.07} & 0.39{\scriptsize$\pm$0.10} & 0.13{\scriptsize$\pm$0.03} \\
\bottomrule
\end{tabularx}%
\end{table*}

\begin{figure}[ht!]
    \centering
    \includegraphics[width=\linewidth]{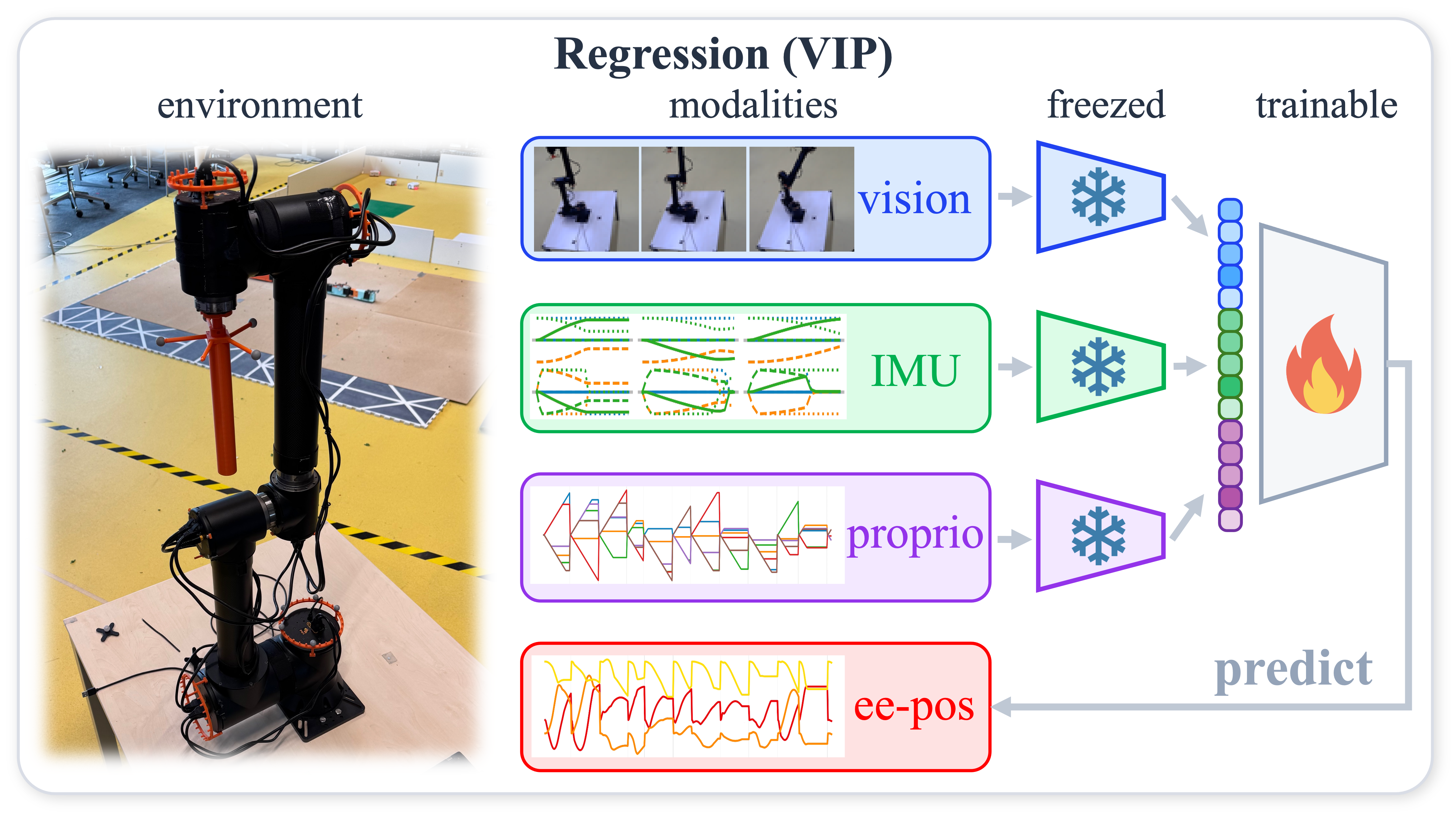}
\caption{\textbf{VIP setup.}
The VIP dataset is collected on a real 6-DoF robotic manipulator (\emph{environment}, left).
At test time, third-person \emph{vision}, per-joint \emph{IMU}, and \emph{proprioception} (joint angles) are embedded and concatenated to a single \emph{trainable} head that predicts the current end-effector position (\texttt{ee-pos}, 3D).
This applies the regression protocol of Fig.~\ref{fig:tasks}(A), and the corresponding results are in Table~\ref{tab:vfp_ee_pos}.}
    \label{fig:vip}
\end{figure}

\paragraph{VIP (Vision--IMU--Proprioception).}
To evaluate whether UniOMA remains effective on real-world robotic data beyond simulation and benchmark datasets, we introduce a new robot-centric dataset collected on a 6-DoF robotic manipulator (Fig.~\ref{fig:vip}).
\revise{In each trajectory, the robot performs a reaching motion toward a sampled end-effector target.}
The task is to predict the current end-effector position (regression, 3D).
Each sample consists of three synchronized modalities:
(1) third-person RGB ($[b{\times}3{\times}64{\times}64]$),
(2) proprioception given by the current joint angles ($[b{\times}6]$),
and (3) normalized inertial measurement unit (IMU) signals from the joints ($[b{\times}6{\times}6]$).
We report the \revise{mean Euclidean distance between the predicted and measured end-effector positions} for this task in Table~\ref{tab:vfp_ee_pos}. 
\revise{Data collection, dataset scale, sensor setup, ground truth, and the data split are described in Supplementary Material, Sec.~II, together with illustrative examples of the modalities.}

\paragraph{End-to-end RL.}
We further evaluate UniOMA \revise{on two closed-loop RL tasks in Isaac Lab~\cite{NVIDIA2025}: locomotion and contact-rich peg insertion, using Proximal Policy Optimization (PPO)~\cite{Schulman2017} for all methods. We assess policy robustness under modality dropout and manipulation performance under full observations.}
Detailed settings are provided in Supplementary Material, Sec.~I.
\revise{As illustrated in Fig.~\ref{fig:method} and Fig.~\ref{fig:tasks}(D), each task uses three observation modalities provided to the actor: torso IMU states ($[b{\times}6]$), joint positions ($[b{\times}12]$), and joint velocities ($[b{\times}12]$) for locomotion; and end-effector pose ($[b{\times}7]$), velocity ($[b{\times}6]$), and wrist force ($[b{\times}3]$) for peg insertion.}
\revise{For locomotion, we introduce joint-velocity dropout after 500 training iterations and continue training until iteration 1,000, with the resulting learning curves shown in Fig.~\ref{fig:method}. We evaluate single-modality dropout at test time using policies trained for 500 iterations. Table~\ref{tab:go1_rl_robustness} shows the average rewards of the locomotion task over 10 episodes and the average success rate of the peg insertion task over 512 episodes.}

\begin{table*}[t]
\centering
\caption{
\textbf{Scalability analysis with 4--7 modalities.}
We report trajectory-pair classification accuracy (mean $\pm$ std over 10 seeds)
together with wall-clock time per epoch\revise{, its ratio to Pairwise (in parentheses), and peak GPU memory} for all three methods:
Pairwise contrastive learning, OTLC, and Pairwise+GW (UniOMA, highlighted in \colorbox{gray!10}{gray}).
\revise{Peak GPU memory was measured on an NVIDIA H100 (batch size 32) and is the same for all three methods at each $M$.}
UniOMA achieves consistently higher accuracy and becomes faster than OTLC when $M\ge6$.
}
\label{tab:scalability-6col}
\small

\resizebox{\linewidth}{!}{ 
\begin{tabular}{lcccccc>{\columncolor{gray!10}}c>{\columncolor{gray!10}}c}
\toprule
\multirow{2}{*}{Modality Combination} &
\multirow{2}{*}{$M$} &
\multirow{2}{*}{\revise{Mem.\ (GiB)}} &
\multicolumn{2}{c}{Pairwise} &
\multicolumn{2}{c}{OTLC} &
\multicolumn{2}{>{\columncolor{gray!10}}c}{Pairwise+GW{\scriptsize (ours)}} \\
\cmidrule(lr){4-5} \cmidrule(lr){6-7} \cmidrule(lr){8-9}
& & & Acc. & Time & Acc. & Time & Acc. & Time \\
\midrule

V+F+P+D           
& 4 & \revise{1.20} & 89.94{\scriptsize$\pm$0.03} & 110.38{\scriptsize$\pm$1.74}s 
& 92.07{\scriptsize$\pm$0.03} & 135.57{\scriptsize$\pm$2.92}s \revise{\scriptsize(1.23$\times$)}
& \textbf{92.39}{\scriptsize$\pm$0.02} & 201.36{\scriptsize$\pm$7.61}s \revise{\scriptsize(1.82$\times$)} \\

V+F+P+D+A      
& 5 & \revise{1.22} & 90.72{\scriptsize$\pm$0.03} & 129.44{\scriptsize$\pm$1.92}s
& 92.51{\scriptsize$\pm$0.03} & 178.63{\scriptsize$\pm$3.11}s \revise{\scriptsize(1.38$\times$)}
& \textbf{93.04}{\scriptsize$\pm$0.02} & 225.89{\scriptsize$\pm$5.44}s \revise{\scriptsize(1.75$\times$)} \\

V+F+P+D+A+C   
& 6 & \revise{1.24} & 89.12{\scriptsize$\pm$0.04} & 150.77{\scriptsize$\pm$2.51}s
& 91.03{\scriptsize$\pm$0.03} & 268.41{\scriptsize$\pm$6.83}s \revise{\scriptsize(1.78$\times$)}
& \textbf{92.11}{\scriptsize$\pm$0.03} & 248.52{\scriptsize$\pm$6.12}s \revise{\scriptsize(1.65$\times$)} \\

V+F+P+D+A+C+O 
& 7 & \revise{1.76} & 87.95{\scriptsize$\pm$0.05} & 171.42{\scriptsize$\pm$3.12}s
& 89.84{\scriptsize$\pm$0.04} & 382.77{\scriptsize$\pm$10.44}s \revise{\scriptsize(2.23$\times$)}
& \textbf{91.02}{\scriptsize$\pm$0.03} & 273.36{\scriptsize$\pm$7.40}s \revise{\scriptsize(1.59$\times$)} \\

\bottomrule
\end{tabular}
}
\end{table*}

\subsection{Implementation Details}
\label{sec:imple-brief}

Encoders, optimizer, temperature, and schedules are shared across methods (fusion heads differ in CoMM). 
\revise{VFD and VFP are split at the trajectory level into training, validation, and test trajectories at a ratio of 80/10/10; validation data are used for early stopping and checkpoint selection, and test data only for the final evaluation.} 
We compute input-space kernels $\{\mathbf{K}^{(m)}_{\mathbf{x}}\}$ (RBF for images \revise{with the median-heuristic bandwidth}; TCK for time-series/force; RBF for other signals) and estimate the batch-wise consensus $\mathbf{C}^{*}_{\mathbf{x}}$ using iterative barycenter updates (Appendix~\ref{app:implementation}). 
Hyperparameters, TCK settings, convergence diagnostics, and end-to-end RL PPO settings are detailed in Supplementary Material, Sec.~I \revise{and Sec.~III-B}.








\begin{table}[t]
\centering
\caption{
\textbf{End-effector position prediction (regression) on Vision + IMU + Proprioception (VIP).}
We report the \revise{mean Euclidean position error ($\times10^{-2}$\,m$\,\downarrow$)}; mean $\pm$ std over 10 seeds).
\colorbox{gray!10}{The gray row} denotes our method (UniOMA).
\revise{For reference, the final two rows report a proprioception-only MLP and calibrated forward kinematics, which uses the robot's kinematic model and therefore indicates the accuracy attainable with privileged model knowledge; both are evaluated against motion-capture ground truth.}
}
\label{tab:vfp_ee_pos}
\small

\resizebox{\linewidth}{!}{
\begin{tabular}{lcc}
\toprule
\multirow{2}{*}{Method} &
\multirow{2}{*}{Modality} &
\multicolumn{1}{c}{\revise{Position error} $\downarrow$ ($\times10^{-2}$\revise{\,m})} \\
\cmidrule(lr){3-3}
& & End-effector Pos. \\
\midrule

Pairwise{\scriptsize~\cite{tian2020contrastive}}
& V+I+P & 6.32{\scriptsize$\pm$0.15} \\

Symile{\scriptsize~\cite{saporta2024contrasting}}
& V+I+P & 4.59{\scriptsize$\pm$0.12} \\

GRAM{\scriptsize~\cite{cicchetti2024gramian}}
& V+I+P & 6.10{\scriptsize$\pm$0.18} \\

TRIANGLE{\scriptsize~\cite{Cicchetti2025}}
& V+I+P & 5.87{\scriptsize$\pm$0.09} \\

\rowcolor{gray!10}
Pairwise+GW{\scriptsize (ours)}
& V+I+P & \textbf{4.02{\scriptsize$\pm$0.10}} \\

\midrule
\revise{Proprioception-only MLP} & \revise{P} & \revise{5.24{\scriptsize$\pm$0.15}} \\
\revise{Calibrated forward kinematics} & \revise{P} & \revise{4.18{\scriptsize$\pm$0.14}} \\
\bottomrule
\end{tabular}
}
\end{table}

\subsection{Results: Comparisons on Downstream Tasks}

We compare UniOMA against state-of-the-art multimodal alignment methods.
First, we consider a family of flexible \emph{InfoNCE-style} frameworks:
(i) \textbf{Pairwise} (CMC)~\cite{tian2020contrastive}, which sums pairwise contrastive losses;
(ii) \textbf{Symile}~\cite{saporta2024contrasting}, extending InfoNCE to triple-wise contrastive objectives; and
(iii) \textbf{GRAM}~\cite{cicchetti2024gramian}, which replaces cosine similarity with Gramian volume similarity for $M\!\geq\!3$ modalities.
These methods share a contrastive backbone, allowing our GW regularizer to be incorporated in a plug-and-play manner; we therefore also report ``\,+GW'' variants for (i), (ii), and (iii).

In addition, we compare against three recent, more specialized multimodal objectives:
(iv) \textbf{OTLC}~\cite{Zhu2024}, which augments pairwise InfoNCE with a standard OT regularizer;
(v) \textbf{TRIANGLE}~\cite{Cicchetti2025}, which aligns three modalities via triangle area similarity; and
(vi) \textbf{CoMM}~\cite{dufumier2024align}, a strong fusion-based baseline with a dedicated multimodal architecture.
All methods are trained with matched optimizer, batch size, temperature, and learning budget (see Supplementary Material, Sec.~I).

\revise{In Table~\ref{tab:vat_results}, adding the GW term improves its contrastive baseline in most paired comparisons, and the best method for each task is a GW-regularized variant; on VIP (Table~\ref{tab:vfp_ee_pos}), Pairwise+GW has the lowest error.
The gains are large for VFD regression with Symile and GRAM and for Vis$\to$Aud retrieval with GRAM, and small for VFP classification, where all methods reach \revise{93.5--94.7\%}.
}


\begin{figure*}[ht!]
    \centering
    \includegraphics[width=\linewidth]{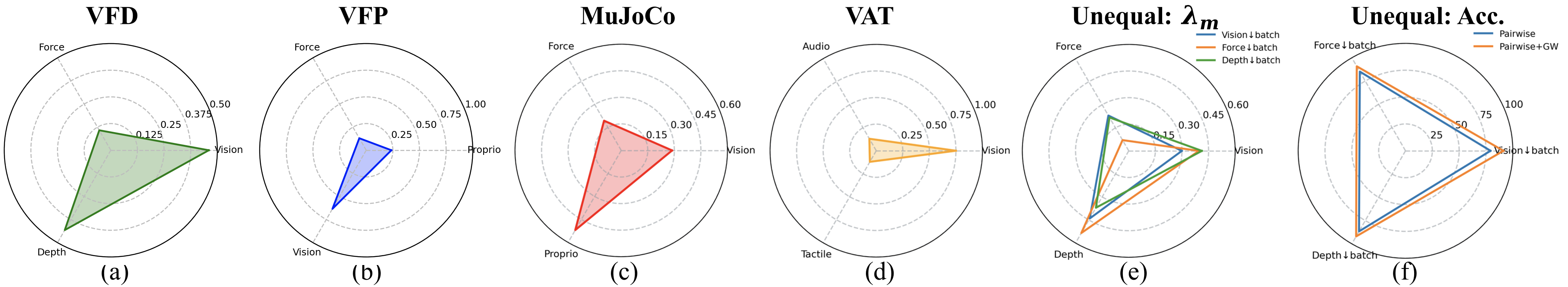}
    \caption{(a–d) Final learned modality weights $\{\lambda_m\}$ for each task (VFD, VFP, MuJoCo Push, VAT). Each radar chart shows per-modality weights that sum to $1$, \revise{showing which modalities agree most with the learned structural consensus in each dataset}. 
(e) Ablation on VFD. One modality is downsampled by $\times\frac{1}{2}$ per batch. The plot shows UniOMA’s adaptive redistribution of $\{\lambda_m\}$ toward intact modalities while retaining signal from the undersampled one. 
(f) Accuracy under the same ablation (Top-1, \%). Pairwise vs.\ Pairwise+GW (UniOMA). The outer polygon indicates consistent gains from the GW regularizer across all downsampled cases.}
    \label{fig:radar-datasets}
\end{figure*}

Table~\ref{tab:go1_rl_robustness} further evaluates policy robustness in the end-to-end RL setting under observation modality dropout. 
We compare the unaligned PPO baseline with Pairwise and Pairwise+GW (UniOMA), where the baseline uses the same training pipeline without aligning the modality encoders. 
The training-time dropout curves in Fig.~\ref{fig:method} show that when the joint-velocity modality is removed after 500 iterations, alignment-based policies retain higher reward than the unaligned baseline.
The test-time dropout in Table~\ref{tab:go1_rl_robustness} shows that
UniOMA achieves the highest mean reward under all dropout cases. 
These results suggest that UniOMA alignment improves RL policy robustness when observations are missing.
\revise{On peg insertion, Pairwise+GW reaches a success rate of 96.6\% under full observations, compared with 92.2\% for Pairwise and 90.8\% for Baseline PPO; the learning curves are in Supplementary Material, Sec.~III-F.}

\begin{table}[t]
\centering
\caption{
\textbf{RL policy \revise{performance under full observations and }robustness under modality dropout.}
\revise{Locomotion:} we report test time average 10-episode reward on an Isaac Lab task: \protect\envname{Isaac-Velocity-Flat-Unitree-Go1-v0} ($\uparrow$; mean $\pm$ std over 3 seeds).
`\revise{IMU}' removes the IMU-related modality, `\revise{Pos.}' removes joint-position modalities, and `\revise{Vel.}' removes joint-velocity modalities.
\revise{Peg insertion: success rate on the Factory task (\%, $\uparrow$; mean $\pm$ std over 5 seeds), with end-effector pose, end-effector velocity, or wrist force removed at test time.}
\colorbox{gray!10}{The gray \revise{rows}} \revise{denote} UniOMA.
}
\vspace{4pt}
\label{tab:go1_rl_robustness}
\small
\setlength{\tabcolsep}{3pt}
\begin{tabular*}{\linewidth}{@{\extracolsep{\fill}}lcccc@{}}
\toprule
Method & \revise{Full obs.} & \multicolumn{3}{c}{\revise{Modality dropped at test time}} \\
\midrule
\textit{\revise{Locomotion (reward)}} & & IMU & Pos. & Vel. \\
Baseline PPO & \revise{\textbf{40.79}{\scriptsize$\pm$0.35}} & 9.98{\scriptsize$\pm$2.39} & 15.29{\scriptsize$\pm$2.16} & 13.86{\scriptsize$\pm$2.41} \\
Pairwise & \revise{40.50{\scriptsize$\pm$0.35}} & 26.50{\scriptsize$\pm$1.76} & 32.74{\scriptsize$\pm$0.93} & 28.31{\scriptsize$\pm$1.57} \\
\rowcolor{gray!10}
Pairwise+GW{\scriptsize~(ours)} & \revise{40.70{\scriptsize$\pm$0.41}} & \textbf{31.51{\scriptsize$\pm$0.94}} & \textbf{34.73{\scriptsize$\pm$0.71}} & \textbf{29.04{\scriptsize$\pm$1.19}} \\
\midrule
\textit{\revise{Peg insertion (\%)}} & & \revise{Pose} & \revise{Vel.} & \revise{Force} \\
\revise{Baseline PPO} & \revise{90.8{\scriptsize$\pm$5.2}} & \revise{3.6{\scriptsize$\pm$1.8}} & \revise{72.3{\scriptsize$\pm$10.5}} & \revise{91.7{\scriptsize$\pm$6.6}} \\
\revise{Pairwise} & \revise{92.2{\scriptsize$\pm$2.4}} & \revise{2.9{\scriptsize$\pm$1.9}} & \revise{66.0{\scriptsize$\pm$10.6}} & \revise{92.3{\scriptsize$\pm$4.0}} \\
\rowcolor{gray!10}
\revise{Pairwise+GW{\scriptsize~(ours)}} & \revise{\textbf{96.6{\scriptsize$\pm$1.5}}} & \revise{\textbf{4.5{\scriptsize$\pm$1.3}}} & \revise{\textbf{78.5{\scriptsize$\pm$7.4}}} & \revise{\textbf{93.9{\scriptsize$\pm$3.9}}} \\
\bottomrule
\end{tabular*}
\end{table}

\subsection{\revise{Results: Structural Alignment and Cross-modal Transfer}}
\label{sec:structure}
\revise{Table~\ref{tab:structure-overview} measures structural alignment directly on VFD and VFP.
We compare each embedding-space kernel of 256 held-out test samples with \revise{a fixed input-kernel reference, the arithmetic mean $\bar{\mathbf K}$ of their input-space kernels,} using pairwise similarity rank correlation, 10-nearest-neighbor overlap, squared-loss GW distance, and the adjusted Rand index of spectral clusterings.
\revise{Unlike the GW barycenter $\mathbf C^*_{\mathbf x}$ used in training, $\bar{\mathbf K}$ does not depend on the trained encoders or the modality weights and is the same for all methods.}
Cross-modal transfer freezes the encoders, fits a ridge probe for next-step end-effector orientation on one modality, and applies it unchanged to another \revise{(illustrated in Supplementary Material, Fig.~5)}; we report the average $R^2$ over modality pairs.
Adding the GW term to Pairwise improves most structural measures and cross-modal transfer on both datasets.
Relational controls that use the same input kernels (Own, Uniform, CKA, and pairwise GW) improve some of the measures, but none matches our GW term on rank correlation, GW distance, and transfer together.}

\begin{figure}[t]
\centering
\includegraphics[width=\columnwidth]{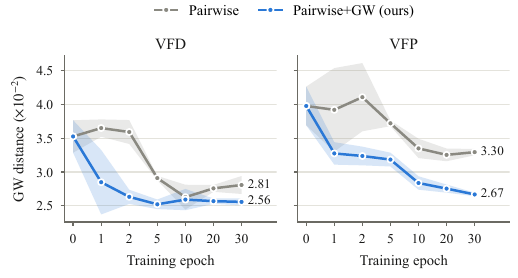}
\caption{\revise{\textbf{GW distance to the \revise{input-kernel reference} during training.}
Lines show the mean $\pm$ standard deviation over three seeds.}}
\label{fig:gw-training}
\end{figure}

\revise{Fig.~\ref{fig:gw-training} tracks the GW distance to \revise{the input-kernel reference} during training: with the GW term, it falls within the first epochs and ends lower than with Pairwise on both datasets.
Fig.~4 of the Supplementary Material shows the corresponding similarity matrices on VFP: after training with the GW term, the structure of each modality is closer to \revise{the input-kernel reference}, with rank correlations of 0.51, 0.67, and 0.56 for vision, force, and proprioception, against 0.35, 0.50, and 0.36 with Pairwise.}

\begin{table*}[t]
\centering
\caption{\revise{\textbf{Structural alignment and cross-modal transfer on VFD and VFP.} \revise{For each modality, the embedding-space kernel of 256 held-out test samples is compared with the input-kernel reference (mean input-space kernel) by the rank correlation of pairwise similarities, 10-nearest-neighbor overlap, GW distance, and the adjusted Rand index of spectral clusterings, averaged over modalities.} Transfer is the $R^2$ of a ridge probe fitted on one modality and applied to another (Sec.~\ref{sec:structure}); \revise{all values are} mean $\pm$ std over five seeds. Own, Uniform, and CKA add a relational loss to the own input kernel, to the mean input kernel, or a CKA loss to a weighted reference; Pairwise GW aligns the embedding kernels of modality pairs without a consensus. The gray rows denote our method; bold marks the best value in each column.}}
\label{tab:structure-overview}
{\color{revisionblue}\footnotesize
\setlength{\tabcolsep}{3pt}
\renewcommand{\arraystretch}{1.05}
\begin{tabularx}{\textwidth}{@{}l*{5}{>{\centering\arraybackslash}X}@{}}
\toprule
 & \multicolumn{4}{c}{Direct structural-alignment measurements} & Cross-modal transfer \\
\cmidrule(lr){2-5}\cmidrule(l){6-6}
Method & Rank correlation\,$\uparrow$ & 10-nearest-neighbor overlap\,$\uparrow$ & GW distance\,$\downarrow$ {\scriptsize($\times10^{-2}$)} & Adjusted Rand index\,$\uparrow$ & $R^2$\,$\uparrow$ \\
\midrule
\multicolumn{6}{@{}l}{\textbf{VFD}} \\
Pairwise{\scriptsize~\cite{tian2020contrastive}} & 0.318{\scriptsize$\pm$0.045} & 0.205{\scriptsize$\pm$0.006} & 2.78{\scriptsize$\pm$0.10} & 0.146{\scriptsize$\pm$0.008} & 0.228{\scriptsize$\pm$0.258} \\
\quad + Own & 0.408{\scriptsize$\pm$0.013} & 0.213{\scriptsize$\pm$0.010} & 2.79{\scriptsize$\pm$0.30} & 0.159{\scriptsize$\pm$0.025} & $-$0.208{\scriptsize$\pm$0.209} \\
\quad + Uniform & 0.417{\scriptsize$\pm$0.024} & 0.218{\scriptsize$\pm$0.004} & 2.63{\scriptsize$\pm$0.03} & 0.160{\scriptsize$\pm$0.017} & 0.194{\scriptsize$\pm$0.188} \\
\quad + CKA & 0.432{\scriptsize$\pm$0.030} & \textbf{0.223}{\scriptsize$\pm$0.006} & 2.72{\scriptsize$\pm$0.08} & 0.156{\scriptsize$\pm$0.019} & 0.076{\scriptsize$\pm$0.152} \\
\quad + Pairwise GW & 0.305{\scriptsize$\pm$0.019} & 0.208{\scriptsize$\pm$0.004} & 2.82{\scriptsize$\pm$0.05} & 0.141{\scriptsize$\pm$0.008} & 0.366{\scriptsize$\pm$0.034} \\
\rowcolor{gray!10}
\quad + GW{\scriptsize~(ours)} & \textbf{0.481}{\scriptsize$\pm$0.042} & 0.219{\scriptsize$\pm$0.009} & \textbf{2.59}{\scriptsize$\pm$0.04} & \textbf{0.163}{\scriptsize$\pm$0.010} & \textbf{0.396}{\scriptsize$\pm$0.175} \\
\midrule
\multicolumn{6}{@{}l}{\textbf{VFP}} \\
Pairwise{\scriptsize~\cite{tian2020contrastive}} & 0.392{\scriptsize$\pm$0.035} & 0.246{\scriptsize$\pm$0.006} & 3.25{\scriptsize$\pm$0.06} & 0.150{\scriptsize$\pm$0.011} & 0.283{\scriptsize$\pm$0.130} \\
\quad + Own & 0.551{\scriptsize$\pm$0.009} & \textbf{0.259}{\scriptsize$\pm$0.004} & 3.45{\scriptsize$\pm$0.18} & 0.154{\scriptsize$\pm$0.020} & $-$0.889{\scriptsize$\pm$0.201} \\
\quad + Uniform & 0.564{\scriptsize$\pm$0.019} & 0.248{\scriptsize$\pm$0.005} & 2.74{\scriptsize$\pm$0.05} & 0.157{\scriptsize$\pm$0.008} & 0.373{\scriptsize$\pm$0.044} \\
\quad + CKA & 0.554{\scriptsize$\pm$0.019} & 0.256{\scriptsize$\pm$0.003} & 2.88{\scriptsize$\pm$0.10} & 0.162{\scriptsize$\pm$0.008} & 0.320{\scriptsize$\pm$0.127} \\
\quad + Pairwise GW & 0.420{\scriptsize$\pm$0.026} & 0.248{\scriptsize$\pm$0.004} & 3.30{\scriptsize$\pm$0.05} & 0.159{\scriptsize$\pm$0.015} & 0.314{\scriptsize$\pm$0.028} \\
\rowcolor{gray!10}
\quad + GW{\scriptsize~(ours)} & \textbf{0.575}{\scriptsize$\pm$0.020} & 0.257{\scriptsize$\pm$0.007} & \textbf{2.70}{\scriptsize$\pm$0.05} & \textbf{0.164}{\scriptsize$\pm$0.007} & \textbf{0.390}{\scriptsize$\pm$0.090} \\
\bottomrule
\end{tabularx}
}
\end{table*}

\subsection{Results: Efficiency and Scalability}
\label{sec:efficiency}

We evaluate scalability on a challenging trajectory-consistency classification task with up to seven modalities (vision, force, proprioception, depth, action, contact, and optical flow).
The task determines whether two multimodal samples originate from the same trajectory; details are provided in Supplementary Material, Sec.~II.
Results for $M\in\{4,5,6,7\}$ modalities are reported in Table~\ref{tab:scalability-6col}.

As the number of modalities increases, distributional heterogeneity grows and alignment becomes more difficult.
Consequently, pairwise contrastive and OT-based baselines decrease in classification accuracy \revise{for $M\geq6$}.
In contrast, UniOMA consistently achieves the highest accuracy across all modality counts, demonstrating scalability in high-modality settings.

Table~\ref{tab:scalability-6col} shows that UniOMA’s wall-clock time per epoch grows approximately linearly with $M$, \revise{while that of OTLC grows superlinearly}.
For $M\ge 6$, UniOMA is strictly faster per epoch than the OTLC baseline, with identical peak memory usage.

We further observe stable convergence of the GW barycenter with a small number of solver iterations ($T_{\max}=5$), as shown in \revise{Supplementary Material, Fig.~6}.
Ablations over \revise{$T_{\max}\in\{3,\dots,7\}$ (Supplementary Material, Table~II)} confirm that performance is insensitive to this choice.
Overall, mini-batch GW inference introduces only moderate overhead and remains practical when scaling modalities.

\subsection{Results: Modality Weights}
\label{sec:weights}

UniOMA learns modality weights $\{\lambda_m\}$ that \revise{determine} each modality’s contribution to the consensus (Appendix~\ref{app:implementation}). 
\revise{Fig.~\ref{fig:radar-datasets} shows the modality weights on each dataset; these values are an empirical indicator of modality-wise agreement with the learned structural consensus rather than a measure of the importance of a modality for the task.} 

\paragraph{\revise{Weight sensitivity and kernel bandwidth.}}
\revise{To better understand the learned modality weights, we tested their sensitivity to cross-modal disagreement and kernel bandwidth on VFD and VFP. 
First, we destroyed cross-modal pairing by replacing images in 75\% of the training samples with images from other samples. 
The vision weight decreased from 0.46 to 0.23 on VFD and from 0.63 to 0.38 on VFP. 
Second, we varied the kernel bandwidth of one modality at a time by factors of $0.25$, $0.5$, $2$, and $4$, while keeping the data unchanged. 
The corresponding weight changes remained below 0.05.}

\revise{These results suggest that the weights reflect the tested cross-modal corruption and show limited sensitivity to kernel bandwidth. 
We further compared learned weights with fixed uniform weights ($\lambda_m=1/M$) under the same GW objective. 
Learned weights showed higher cross-modal transfer $R^2$ under corruption ($-0.26$ vs.\ $-0.58$ on VFD; $-1.44$ vs.\ $-2.17$ on VFP). Thus, the weights serve as an empirical indicator of modality-wise agreement with the learned structural consensus.}

\subsection{Ablation Study: Unequal Modality Sampling}
\label{sec:ablation}

To evaluate robustness to asynchronous robot perception, we ablate UniOMA on the VFD classification by downsampling one modality per mini-batch by $\times\frac{1}{2}$ (vision/force/depth $\downarrow$ batch). 
Specifically, we downsample one modality (e.g., $b{=}32$) while keeping others at $b{=}64$, which yields unequal sample frequencies and removes strict one-to-one pairing across modalities.
We compare UniOMA (Pairwise+GW) with a contrastive-only baseline (Pairwise).
As shown in Fig.~\ref{fig:radar-datasets}(f), UniOMA (Pairwise+GW) consistently outperforms Pairwise under all downsampling cases, showing that UniOMA remains effective without \revise{strict one-to-one} cross-modal pairing.

\paragraph{Interpretability (weights).} Fig.~\ref{fig:radar-datasets}(e) reports the learned modality weights under each downsampling setting: UniOMA shifts the weight toward intact modalities while still retaining non-trivial mass on the undersampled one.
Across datasets (Fig.~\ref{fig:radar-datasets}(a--d)), the aggregated weights further \revise{show that the modality agreeing most with the learned structural consensus differs between tasks}.

\paragraph{Interpretability (structure).} We visualize the mini-batch GW barycenter in Supplementary Material, \revise{Sec.~III-D, which} shows coherent geometric patterns shared across modalities, indicating that UniOMA aligns intra-modal \emph{structure} rather than raw features.

\section{Conclusions, Limitations, and Future Work}
\label{sec:discuss}

We revisit multimodal alignment through \emph{structural} consistency: while pointwise correspondences are statistically strong in existing methods, intra-modal geometries can still disagree across modalities. 
UniOMA \revise{reduces} this gap by combining contrastive learning with a GW-barycenter regularizer that aligns 3+ modalities to a shared structural consensus. 
Concretely, UniOMA represents each modality as an intra-modal similarity (kernel) space, computes a dynamic GW barycenter as the structural consensus, and aligns each modality to this consensus with learned modality weights. 
This barycentric formulation avoids pairwise couplings across modalities, and \revise{is well suited to scaling to} more modalities.
Across VFP, VFD, MuJoCo Push, VAT, VIP, and 4--7 modality settings, UniOMA \revise{improves regression, classification, and cross-modal retrieval in most paired comparisons}, \revise{brings the similarity structure of each modality closer to \revise{a fixed input-kernel reference},} and learns dataset-specific modality weights \revise{that serve as an empirical indicator of modality-wise agreement with the learned structural consensus}.

\paragraph{Limitations.}
UniOMA introduces additional computation due to the barycentric GW updates, but we explicitly quantify this overhead in Table~\ref{tab:scalability-6col}. 
Importantly, the runtime scales \emph{linearly} with the number of modalities, rather than superlinearly, which makes UniOMA practical for 3+ modalities. 
A second consideration is the choice of intra-modal similarity kernels used to form \revise{$\mathbf{K}^{(m)}_{\mathbf{x}}$}. 
While kernel mismatch can affect the learned consensus geometry, we partially address this sensitivity through \revise{the kernel-bandwidth test of Sec.~IV-F and the feature-extractor swap of Supplementary Material, Sec.~III-H}, which empirically validate the robustness of UniOMA under reasonable kernel choices.

\revise{The shared structural consensus is most appropriate when common physical factors induce related relational patterns across modalities. 
If the modalities capture largely distinct information, alignment may ignore useful modality-specific structure. 
In addition, the learned weights are an empirical indicator of modality-wise agreement with the learned structural consensus and do not directly measure modality importance.}
\revise{The entropy term discourages but does not prevent the concentration of the weights on one modality.}
\revise{Better structural alignment does not necessarily yield better downstream performance: the gains in Table~\ref{tab:vat_results} are small on tasks close to saturation, and \revise{one comparison shows} a decrease.
Simpler relational losses that use the same input-space kernels can be comparable to or better than the GW term on the structural measures (Table~\ref{tab:structure-overview}).
Finally, the GW term increases the training time per epoch by 1.6--1.8$\times$ relative to Pairwise (Table~\ref{tab:scalability-6col}), which may not be justified when the modalities share little structure or when the contrastive baseline is already close to saturation.}

\paragraph{Future work.}
Building on these findings, a promising direction is large-scale real-robot alignment under heterogeneous sampling rates, missing segments, and asynchronous sensor streams, which calls for more robust couplings and barycenter estimation in the presence of temporal misalignment. 
Another direction is extending the framework beyond symmetric similarity kernels to \emph{asymmetric} or \emph{directed} similarities (e.g., temporal precedence or causal structure), enabling the consensus geometry to capture directional relations that arise in contact-rich robotics data.




%% file: main-references.bbl

%% file: supplement-references.bbl

%% file: appendix-main.tex
\section{Implementation Details}
\label{app:implementation}
\subsection{Optimal Transport Plan Estimation}
\label{app:ot-plan}

\begin{algorithm}[htbp]
    \caption{\texttt{OTPlan}$(\hat{\mathbf{K}}, \mathbf{K})$}
    \label{alg:ot_estimation}
    \KwIn{Kernel matrices $\hat{\mathbf{K}} \in \mathbb{R}^{\hat N \times \hat N}, \mathbf{K} \in \mathbb{R}^{N \times N}$}
    \KwOut{Optimal transport matrix $\mathbf{T}^*$}
    
    \BlankLine
    Initialize $\mathbf{p} \leftarrow \frac{1}{N}\mathbf{1}_N, \quad \hat{\mathbf{p}} \leftarrow \frac{1}{\hat N}\mathbf{1}_{\hat N}, \quad \mathbf{T} \leftarrow \hat{\mathbf{p}} \mathbf{p}^{\top}$\;
    
    \While{not converged}{
        \tcp{\textcolor{gray}{Apply network simplex algorithm:}}
        $\hat{\mathbf{T}} \leftarrow \argmax_{\revise{\mathbf{T}'} \in \Pi(\hat{\mathbf{p}}, \mathbf{p})} \operatorname{tr}(\hat{\mathbf{K}}^{\top} \revise{\mathbf{T}'^{\top}} \mathbf{K} \mathbf{T})$\;
        
        \tcp{\textcolor{gray}{Line search to find the minimum:}}
        $a \leftarrow -2 \operatorname{tr}(\hat{\mathbf{K}}^{\top} \revise{(\hat{\mathbf{T}}-\mathbf{T})}^{\top} \mathbf{K} \revise{(\hat{\mathbf{T}}-\mathbf{T})})$\;
        
        $b \leftarrow \operatorname{tr}\big(\big((\hat{\mathbf{K}} \odot \hat{\mathbf{K}}) \hat{\mathbf{p}} \revise{\mathbf{1}^{\top}} + \revise{\mathbf{1}\mathbf{p}^{\top}} (\mathbf{K} \odot \mathbf{K})\big)^{\top}\revise{(\hat{\mathbf{T}}-\mathbf{T})}\big)$\;
        
        $c \leftarrow \begin{aligned}[t]-2\big(&\operatorname{tr}(\hat{\mathbf{K}}^{\top} \mathbf{T}^{\top} \mathbf{K} \revise{(\hat{\mathbf{T}}-\mathbf{T})})\\&+\operatorname{tr}(\hat{\mathbf{K}}^{\top} \revise{(\hat{\mathbf{T}}-\mathbf{T})}^{\top} \mathbf{K} \mathbf{T})\big);\end{aligned}$\DontPrintSemicolon\;\PrintSemicolon
        
        \If{$a > 0$}{
            $\tau \leftarrow \min(1, \max(0, -\frac{b+c}{2a}))$\;
        }
        \Else{
            $\tau \leftarrow \begin{cases} 1, & \text{if } a+b+c < 0, \\ 0, & \text{otherwise.} \end{cases}$\;
        }
        
        $\mathbf{T} \leftarrow (1-\tau) \mathbf{T} + \tau \hat{\mathbf{T}}$\;
    }
    \Return $\mathbf{T}$\;
    
\end{algorithm}

Algorithm~\ref{alg:ot_estimation} computes an empirical OT plan $\mathbf{T}$ by
\revise{minimizing the GW distance of Eq.~(\ref{eq:gw}) for fixed kernels; since its first two terms do not depend on $\mathbf{T}\in\Pi(\hat{\mathbf{p}},\mathbf{p})$, this is equivalent to} solving the quadratic program
\[
\max_{\mathbf{T}\in\Pi(\hat{\mathbf{p}},\mathbf{p})}\;
f(\mathbf{T})\ :=\ \mathrm{tr}\!\big(\hat{\mathbf{K}}^{\top}\,\mathbf{T}^{\top}\,\mathbf{K}\,\mathbf{T}\big),
\]
where $\hat{\mathbf{K}},\mathbf{K}\in\mathbb{R}^{N\times N}$ are intra-domain
similarity (or distance) matrices and
$\Pi(\hat{\mathbf{p}},\mathbf{p})=\{\mathbf{T}\!\ge\!0\mid
\mathbf{T}\mathbf{1}=\hat{\mathbf{p}},\ \mathbf{T}^{\top}\mathbf{1}=\mathbf{p}\}$
is the transportation polytope (doubly-stochastic when
$\hat{\mathbf{p}}=\tfrac{1}{\hat N}\mathbf{1}_{\hat N},\mathbf{p}=\tfrac{1}{N}\mathbf{1}_N$).
Here $\odot$ is the Hadamard product, so $(\hat{\mathbf{K}}\odot\hat{\mathbf{K}})$
and $(\mathbf{K}\odot\mathbf{K})$ are elementwise squares of the corresponding kernels,
which makes $b$ compact.
We initialize with the independent coupling $\mathbf{T}=\hat{\mathbf{p}}\mathbf{p}^{\top}$
and iterate a Conditional Gradient (Frank--Wolfe; FW) update.

\revise{The linear oracle (network simplex) and the closed-form line search are given in Supplementary Material, Sec.~I-F.}

\subsection{GW Barycenter Estimation}
\label{app:gw-barycenter}

\begin{algorithm}[]
\caption{GW Barycenter Estimation (mini-batch)}
\label{alg:bc}
\KwIn{Intra-modal similarity matrices $\{\mathbf{K}^{(m)}_{\mathbf{x}}\}_{m=1}^{M}$ (batch size $N_m$ per modality with $\text{min}\{N_m\}=N$), modality weights $\{\lambda_m\}_{m=1}^{M}$ with $\lambda_m\!\ge\!0,\ \sum_m\lambda_m\!=\!1$, uniform marginal $\hat{\mathbf{p}}=\tfrac{1}{N}\mathbf{1}_N$, max iters $T_{\max}$}
\KwOut{Batch-wise structural consensus (GW barycenter) $\mathbf{C}^*_{\mathbf{x}}\in\mathbb{R}^{N\times N}$}

\BlankLine
Initialize $\mathbf{C}_{\mathbf{x}}$ as the weighted average of $\mathbf{K}^{(m)}_{\mathbf{x}}$ \;

\For{$t \leftarrow 0$ \KwTo $T_{\max}-1$}{
  \For{$m \leftarrow 1$ \KwTo $M$}{
    $\mathbf{T}^{(m)} \leftarrow \texttt{OTPlan}\!\left(\mathbf{C}_{\mathbf{x}},\,\mathbf{K}^{(m)}_{\mathbf{x}}\right)$ \tcp*{Alg.~\ref{alg:ot_estimation}}
  }
  $\widetilde{\mathbf{C}} \leftarrow \sum_{m=1}^{M}\lambda_m\,\mathbf{T}^{(m)}\,\mathbf{K}^{(m)}_{\mathbf{x}}\,\mathbf{T}^{(m)\top}$
  \;
  
  $\mathbf{C}_{\mathbf{x}} \leftarrow \widetilde{\mathbf{C}} \oslash \left(\hat{\mathbf{p}}\,{\hat{\mathbf{p}}}^{\top}\right)$
  \;
}
\Return $\mathbf{C}^*_{\mathbf{x}} \leftarrow \mathbf{C}_{\mathbf{x}}$
\end{algorithm}

\revise{For fixed couplings $\mathbf{T}^{(m)}$, the update of $\mathbf{C}_{\mathbf{x}}$ in Alg.~\ref{alg:bc} is the closed-form minimizer of the barycenter objective of Def.~\ref{def:bc}, where $\oslash$ denotes the element-wise division; the derivation is given in Supplementary Material, Sec.~I-G.}

%% file: appendix-supplement.tex
                                                    
\title{Supplementary Material for\\                             
\textit{UniOMA: Unified Optimal-Transport Multi-Modal Structural    
Alignment for Robot Perception}}
\author{Xinrui Zu, Mark Hoogendoorn, Kevin Sebastian Luck, Shujian Yu\\[0.4em]\normalsize Vrije Universiteit Amsterdam, the Netherlands}                                          
\maketitle                                                          

\section{Implementation Details}
\label{supp:appendix:imple}

\subsection{Implementation: Time-Series Cluster Kernel}
\label{supp:appendix:kernels}
We use the Time-series Cluster Kernel (TCK; \cite{supp:Mikalsen2018}) to build intra-modal similarity matrices for time-series modalities (e.g., force/torque). TCK fits an ensemble of diagonal covariance Gaussian mixture models (GMMs) with informative priors and computes a posterior membership vector per sample
\[
\boldsymbol{\Pi}_i(q)\;=\;\bigl(\pi^{(i)}_1(q),\,\ldots,\,\pi^{(i)}_{G_q}(q)\bigr)^\top,\qquad
\sum_{g=1}^{G_q}\pi^{(i)}_g(q)=1,
\]
where each component $\pi^{(i)}_g(q)$ is the posterior responsibility of mixture $g$ for sequence $i$ under the $q$-th GMM, i.e.
\[
\pi^{(i)}_g(q)\;=\;p\!\left(z{=}g\;\middle|\;\mathbf{x}_i^{(q)};\,\hat\theta_q\right),
\]
where $z$ is the latent mixture index, $\hat\theta_q$ is the MAP-EM estimate of the $q$-th model parameters, and $\mathbf{x}_i^{(q)}$ is the subsequence of $i$ restricted to the time window and variable subset chosen by that ensemble member.

The final kernel is the sum of posterior inner products over the ensemble:
\[
(K_{\text{TCK}})_{ij}\ \leftarrow\ \sum_{q\in\mathcal{Q}} \boldsymbol{\Pi}_i(q)^\top \boldsymbol{\Pi}_j(q),
\]
which is positive semidefinite as a sum of linear kernels. 
In practice, for time-critical training, we \revise{fit the TCK ensemble (the GMM parameters) on the training trajectories only. With this ensemble fixed, we compute the posteriors of the training, validation, and test samples separately and precompute one symmetric matrix per split,} $\mathbf{K}_{\text{TCK}}\in\mathbb{R}^{N\times N}$ \revise{over the $N$ samples of that split; no kernel entry couples samples of different splits}.
During mini-batch training, the intra-modal similarity submatrix for an index set $\mathcal{I}\subseteq\{1,\dots,N\}$ is obtained by simple indexing
\[
\mathbf{K}_{\text{batch}}\;=\;K_{\text{TCK}}[\mathcal{I},\,\mathcal{I}],
\]
thus avoiding repeated TCK fits inside the inner learning loop. 
\revise{Training uses only the training-split matrix, and the test-split matrix is used only for the final evaluation; no test trajectory enters TCK fitting, training, or model selection.}
We follow the original TCK protocol to induce ensemble diversity (random time windows and variable subsets, random initializations, and varying mixture counts), and we cache per-member posteriors to enable fast posterior lookups at test time.
See \S4.1--4.4 of \cite{supp:Mikalsen2018} for modeling details.
In practice, we set the maximal number of mixtures $C$ and the number of randomizations $Q$ as the only user-set hyperparameters. We set $C{=}30$ and $Q{=}15$ for force/torque signals in the VFP and VFD settings.

\subsection{Implementation: Pre-trained Features}
\label{supp:appendix:x}
We use pre-trained feature extractors for some modalities to produce modality-specific features whose pairwise similarities form the input-space kernels \(\{\mathbf{K}^{(m)}_{\mathbf{x}}\}_{m=1}^M\) used by our structural consensus $\mathbf{C}^*_{\mathbf{x}}$. For the time-series modality (e.g., force/torque), we directly use the TCK method to obtain the input-space kernels.

\textit{Pre-trained feature extractors (frozen).} 
\begin{itemize}
\item \textbf{Vision / Depth / Tactile / Optical Flow:} Vision Transformer (ViT-B/16) \cite{supp:Dosovitskiy2020}, taking the final [CLS] embedding. Single-channel inputs (e.g., depth) are replicated to 3 channels before preprocessing.
\item \textbf{Force:} Time-Series Cluster Kernel (TCK; \cite{supp:Mikalsen2018}) directly forms $\mathbf{K}^{(\text{force})}_{\mathbf{x}}$ (Sec.~\ref{supp:appendix:kernels}).
\item \textbf{Audio (VAT):} A frozen Audio Spectrogram Transformer (AST-B, AudioSet-pretrained; \cite{supp:gong21ast}) on log-mel spectrograms; we take the [CLS] embedding and build $\mathbf{K}^{(\text{aud})}_{\mathbf{x}}$ with \revise{an RBF kernel (median heuristic, Sec.~\ref{supp:appendix:nn})}.
\item \textbf{Other modalities:} RBF kernel on frozen features.
\end{itemize}

\begin{figure}[h!]
    \centering
    \includegraphics[width=0.7\columnwidth]{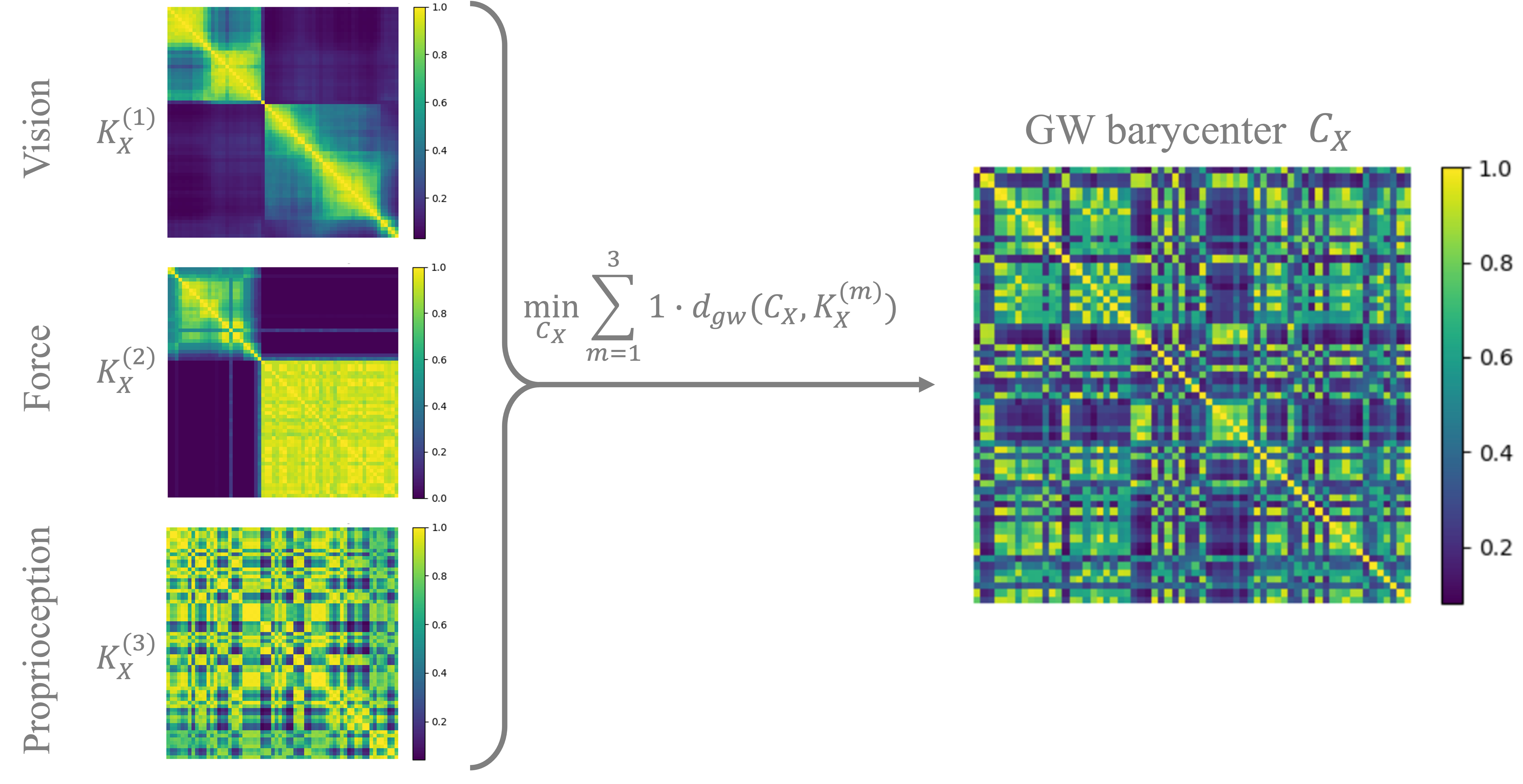} 
    \caption{GW barycenter of input-space kernels on the VFP dataset. 
Left: intra-modal similarity matrices $K^{(1)}_{\mathbf{x}}$ (Vision), $K^{(2)}_{\mathbf{x}}$ (Force), and $K^{(3)}_{\mathbf{x}}$ (Proprioception), each min–max normalized for display. 
Right: the batch-wise structural consensus $\mathbf{C}^*_{\mathbf{x}}$ obtained by solving 
$\displaystyle \min_{\mathbf{C}_{\mathbf{x}}}\sum_{m=1}^{3}\lambda_m\, d_{gw}\!\big(\mathbf{C}_{\mathbf{x}},K^{(m)}_{\mathbf{x}}\big)$ 
(with \revise{$\lambda_m{=}1/3$} here). 
The barycenter preserves recurrent block/trajectory patterns shared across modalities while smoothing modality-specific artifacts, and is later used to regularize the embedding-space geometry in Stage~2. The batch size is 64.}
    \label{supp:fig:VFP}
\end{figure}

\subsection{Implementation: Modality Encoders}
\label{supp:appendix:nn}
To avoid architectural confounds, all methods share identical backbones and training schedules. In UniOMA (Stage~2), each modality encoder $\mathcal{E}^{(m)}_\theta$ produces a feature $\mathbf{h}^{(m)}\!\in\!\mathbb{R}^{d_h}$, which is passed through a \emph{modality-specific MLP projector} $g^{(m)}_\theta$ to a \emph{shared} embedding size $d{=}256$:
\[
{\mathbf{z}}^{(m)} \;=\; g^{(m)}_\theta\!\big(\mathcal{E}^{(m)}_\theta(\mathbf{x}^{(m)})\big)\in\mathbb{R}^{d}, \quad f_{\theta}^{(m)}=g_{\theta}^{(m)}\revise{\circ} \mathcal{E}_{\theta}^{(m)}
\]
and the embedding-space kernel within a mini-batch is
\[
\big({\mathbf{K}^{(m)}_{\mathbf{z}}}\big)_{ij}
=\exp\!\Big(-\gamma\,\|{\mathbf{z}}^{(m)}_i-{\mathbf{z}}^{(m)}_j\|_2^2\Big),
\qquad
\revise{\gamma \;=\; \Big(2\,\operatorname{median}_{a,b}\|{\mathbf{z}}^{(m)}_a-{\mathbf{z}}^{(m)}_b\|_2^2\Big)^{-1}},
\]
\revise{where the median is taken over the squared distances of the mini-batch (median heuristic); the input-space RBF kernels use the same rule.} (Stage~1 input-space kernels $\{\mathbf{K}^{(m)}_{\mathbf{x}}\}$ are computed independently using frozen extractors; see Sec.~\ref{supp:appendix:x}.)

\textit{Backbones.}
\begin{itemize}
\item \textbf{Vision / Depth / Tactile (image-based).} A 2D CNN (ResNet-18). Single-channel inputs (e.g., depth, some tactile images) are replicated to 3 channels before feeding into the backbone.
\item \textbf{Force (time series).} A 1D temporal ConvNet built from stacked causal Conv1D layers (kernel size 2, stride 2) with LeakyReLU activations; the final feature map is flattened to obtain a fixed-length embedding.
\item \textbf{Proprioception.} A 3-layer MLP with LeakyReLU activations, mapping the low-dimensional pose / joint vector to the shared embedding space.
\item \textbf{Audio (VAT).} A 1D CNN with three convolutional blocks (channels $1\!\to\!64\!\to\!128\!\to\!256$, kernel size $5$, stride $2$), each followed by ReLU, then \texttt{AdaptiveAvgPool1d(1)}, \texttt{flatten()}, and a final \texttt{Linear}(256$\!\to\!d_h$).
\item \textbf{Action.} A small MLP mapping the action vector to a compact embedding, implemented as
      \texttt{Linear}($d_a\!\to\!32$)–LeakyReLU–\texttt{Linear}($32\!\to\!32$)–LeakyReLU.
\item \textbf{Contact.} A lightweight MLP applied to the binary contact state, using a structure analogous to the action branch (two \texttt{Linear} layers with LeakyReLU) to obtain a 32-dimensional embedding.
\item \textbf{Optical Flow.} We take one channel of the dense optical-flow field (e.g., horizontal component or magnitude), resize it to $128{\times}128$, replicate it to 3 channels, and feed it through the same ResNet-18 image encoder as for RGB and depth.

\end{itemize}

We fix the projector output to $d{=}256$, use the same temperature $\tau$ for the contrastive term, and share optimizer, batch size, and schedule across methods. UniOMA augments the contrastive loss with a GW-barycenter regularizer (weight $\alpha$) and learnable modality weights $\{\lambda_m\}$ (softmax-parameterized to enforce $\lambda_m\!\ge\!0$ and $\sum_m\lambda_m\!=\!1$). Encoders and projectors are trained end-to-end with the UniOMA objective; the structure-aware term is computed on $\{{\mathbf{K}^{(m)}_{\mathbf{z}}}\}$, while Stage~1 kernels $\{\mathbf{K}^{(m)}_{\mathbf{x}}\}$ remain fixed within each epoch.

\subsection{Implementation: Hyper-parameters}
\label{supp:appendix:hparam}

\textit{Shared training.}
Unless otherwise noted, all methods use the same backbone--projector settings.
We optimize with AdamW (learning rate \(3{\times}10^{-4}\), weight decay \(10^{-4}\), \(\beta_1{=}0.9,\beta_2{=}0.999\)), batch size 64, and temperature \(\tau{=}0.1\).
Each modality head outputs a \(d{=}256\)-dimensional embedding via a lightweight MLP projector (shared width across modalities).
We train for 200 epochs with early stopping on the validation metric when applicable, and report mean\(\pm\)std over 10 independent seeds\revise{, unless otherwise noted (the structural and weight analyses use 30 epochs and five seeds; Sec.~\ref{supp:supp:structural})}.

\textit{Stage-1 input-space kernels.}
Pre-trained feature extractors for vision/depth/tactile (ViT-B/16 via \texttt{timm}) are \emph{frozen} to compute \(\{\mathbf{K}^{(m)}_{\mathbf{x}}\}\).
For force/torque we use TCK with max mixtures \(C{=}30\) and randomizations \(Q{=}15\) following Sec.~\ref{supp:appendix:kernels}.
For VAT audio, we use AST-B as in Sec.~\ref{supp:appendix:x} to form features and then an RBF kernel.
To avoid repeated online estimation during Stage~2, we compute \revise{the force kernel of each split once with the TCK ensemble fitted on the training trajectories} and cache it; mini-batch kernels $\mathbf{K}_{\text{batch}}^{(\text{force})}$ are obtained by submatrix indexing.

\textit{Stage-2 embedding-space kernels.}
All modalities use the same Gaussian kernel
\[
\big({\mathbf{K}^{(m)}_{\mathbf{z}}}\big)_{ij}
=\exp\!\big(-\gamma\,\|\mathbf{z}^{(m)}_i-\mathbf{z}^{(m)}_j\|^2_2\big),
\]
\revise{with $\gamma$ set by the median heuristic of Sec.~\ref{supp:appendix:nn}.}

\textit{UniOMA-specific.}
The GW regularization weight \(\alpha=1000\).
Modality weights \(\{\lambda_m\}\) are learnable with a softmax parameterization (\(\lambda_m\!\ge\!0,\ \sum_m\lambda_m{=}1\)) and initialized uniformly\revise{; the entropy term of the training objective uses $\beta=1$ and $\varepsilon=10^{-9}$}.
For the coupling oracle in \texttt{OTPlan} we use a Frank--Wolfe linearization; the linear subproblem is solved with a network-simplex transportation solver.
The line search on the FW segment uses the closed-form quadratic coefficients \((a,b,c)\) derived in \revise{Sec.~\ref{supp:supp:otplan}}.
GW barycenter iterations are run with a maximum of \(T_{\max}{=}5\) per inner-loop (we analyze this choice in Sec.~\ref{supp:appendix:hparam_analysis}).

To further justify these design choices, we provide a hyper-parameter ablation in Sec.~\ref{supp:appendix:hparam_ablation}, where we evaluate the effects of \revise{$\alpha$ and $T_{\max}$}.

\subsection{PPO Settings for End-to-End RL}
\label{supp:app:go1-ppo}
\revise{Table~\ref{supp:tab:go1_ppo_settings} lists the PPO and rollout settings of the two RL tasks. Within each task, the baseline, pairwise alignment, and UniOMA share these settings and differ only in the alignment loss.}
\revise{The peg-insertion task uses the 8\,mm peg and hole of Isaac Lab Factory (diameters 7.99\,mm and 8.1\,mm, i.e., about 0.11\,mm clearance) with a Franka arm. An episode lasts 10\,s (150 policy steps), and a trial counts as successful when the peg is centered within 2.5\,mm of the hole axis and seated within 1\,mm of the target depth.}

\begin{table*}[t]
\centering
\caption{
\textbf{PPO and rollout settings for the \revise{RL experiments (Go1 locomotion and peg insertion)}.}
These settings are shared by the baseline, pairwise alignment, and UniOMA variants unless otherwise stated.
}
\label{supp:tab:go1_ppo_settings}
\small
\begin{tabular*}{0.9\textwidth}{@{\extracolsep{\fill}}lll}
\toprule
Setting & \revise{Go1 locomotion} & \revise{Peg insertion} \\
\midrule
Environment & \envname{Isaac-Velocity-Flat-Unitree-Go1-v0} & \revise{Factory peg insertion (+ wrist force)} \\
RL algorithm & PPO & \revise{PPO} \\
Parallel environments & 4096 & \revise{512} \\
Rollout length & 24 steps/env & \revise{64 steps/env} \\
Transitions per PPO iteration & 98{,}304 & \revise{32{,}768} \\
Mini-batch size & 24{,}576 transitions & \revise{4{,}096 transitions} \\
Learning iterations & 500 for test-time dropout; 1000 for training dropout & \revise{300} \\
Training-dropout activation & iteration 500 & \revise{none} \\
Optimizer & Adam & \revise{Adam} \\
Learning rate & $1.0{\times}10^{-3}$, adaptive schedule & \revise{$1.0{\times}10^{-4}$, adaptive schedule} \\
Desired KL & 0.01 & \revise{0.008} \\
Discount factor $\gamma$ & 0.99 & \revise{0.995} \\
GAE parameter $\lambda$ & 0.95 & \revise{0.95} \\
PPO epochs per iteration & 5 & \revise{4} \\
Mini-batches per iteration & 4 & \revise{8} \\
PPO clip parameter & 0.2 & \revise{0.2} \\
Value loss coefficient & 1.0 & \revise{2.0} \\
Entropy coefficient & 0.01 & \revise{0.0} \\
Max gradient norm & 1.0 & \revise{1.0} \\
Clipped value loss & enabled & \revise{enabled} \\
Observation normalization & disabled & \revise{enabled} \\
Actor/critic hidden layers & $[128,128,128]$ & \revise{$[512,256,128]$} \\
Activation & ELU & \revise{ELU} \\
Action distribution & Gaussian, initial std 1.0 & \revise{Gaussian, initial std 1.0} \\
\revise{Training seeds} & \revise{3} & \revise{5} \\
\revise{Evaluation} & \revise{average 10-episode reward} & \revise{average 512-episode success rate} \\
\bottomrule
\end{tabular*}
\end{table*}

\subsection{\revise{Optimal Transport Plan Estimation: Linear Oracle and Line Search}}
\label{supp:supp:otplan}
{\color{revisionblue}
This section details Alg.~2 (\texttt{OTPlan}) of the main paper, which maximizes $f(\mathbf{T})=\mathrm{tr}(\hat{\mathbf{K}}^{\top}\mathbf{T}^{\top}\mathbf{K}\mathbf{T})$ over $\Pi(\hat{\mathbf{p}},\mathbf{p})$, which is equivalent to minimizing the GW distance of Eq.~(8) of the main paper for fixed kernels.

\paragraph{Network simplex algorithm.}
At each iteration, we linearize $f$ and solve
\[
\hat{\mathbf{T}}
\ \in\ \arg\max_{\mathbf{T}'\in\Pi(\hat{\mathbf{p}},\mathbf{p})}
\ \big\langle \mathbf{T}',\ \nabla f(\mathbf{T})\big\rangle.
\]
For $f(\mathbf{T})=\mathrm{tr}(\hat{\mathbf{K}}^{\top}\mathbf{T}^{\top}\mathbf{K}\mathbf{T})$,
we use the gradient form
\[
\nabla f(\mathbf{T})=\mathbf{K}\,\mathbf{T}\,\hat{\mathbf{K}}
\ +\ \mathbf{K}^{\top}\,\mathbf{T}\,\hat{\mathbf{K}}^{\top},
\]
which reduces to $2\,\mathbf{K}\,\mathbf{T}\,\hat{\mathbf{K}}$ when
$\mathbf{K},\hat{\mathbf{K}}$ are symmetric, so that $\langle \mathbf{T}',\nabla f(\mathbf{T})\rangle=2\,\mathrm{tr}(\hat{\mathbf{K}}^{\top}\mathbf{T}'^{\top}\mathbf{K}\mathbf{T})$ as in Alg.~2.
The oracle is a linear transportation problem.
We implement it using a network simplex \cite{supp:JMLR:v22:20-451,supp:Bonneel2011}.

\paragraph{Line search.}
Define the search segment
$\mathbf{T}(\tau)=(1-\tau)\mathbf{T}+\tau\hat{\mathbf{T}}$, $\tau\in[0,1]$.
\revise{Substituting $\mathbf{T}(\tau)$ into the GW distance of Eq.~(8) of the main paper yields the univariate quadratic
$d(\tau)=d(0)+a\,\tau^2+(b+c)\,\tau$, where $d(\tau)$ is the GW distance at $\mathbf{T}(\tau)$,
with the coefficients of Alg.~2 of the main paper: $a$ and $c$ come from the trace term and $b$ from the two quadratic terms ($b=0$ when $\mathbf{T}$ and $\hat{\mathbf{T}}$ satisfy the marginal constraints exactly).
The code computes $(a,b,c)$ and picks the minimizer on $[0,1]$:}
$\tau^\star=\min\!\bigl(1,\max(0,-(b+c)/(2a))\bigr)$ if $a>0$,
otherwise $\tau^\star\in\{0,1\}$ by comparing endpoints.
We then set $\mathbf{T}=\mathbf{T}(\tau^\star)$.
}

\subsection{\revise{GW Barycenter Update}}
\label{supp:supp:bary}
{\color{revisionblue}
Consider the barycenter objective (Def.~2 of the main paper):
\[
\begin{aligned}
\mathbf{C}^*_{\mathbf{x}}
&= \operatorname*{arg\,min}_{\mathbf{C}_{\mathbf{x}}\in\mathcal{M}}
\sum_{m=1}^M \lambda_m \cdot \revise{\hat d_{gw}}\!\left(
\mathbf{C}_{\mathbf{x}},
\mathbf{K}^{(m)}_{\mathbf{x}}
\right),\\
&\quad \lambda_m\ge 0,\quad
\sum_{m=1}^M \lambda_m=1.
\end{aligned}
\]
\revise{By Eq.~(8) of the main paper, each term consists of the quadratic term of $\mathbf{C}_{\mathbf{x}}$, the quadratic term of $\mathbf{K}^{(m)}_{\mathbf{x}}$, and a trace term, and the barycenter minimizes their weighted sum.}
Fix couplings $\{\mathbf{T}^{(m)}\}_{m=1}^M$ with
$\mathbf{T}^{(m)}\!\in\!\Pi(\hat{\mathbf{p}},\mathbf{p}^{(m)})$ for the current
consensus $\mathbf{C}_{\mathbf{x}}$, and define
\[
\mathbf{A}^{(m)}\ :=\ \mathbf{T}^{(m)}\,\mathbf{K}^{(m)}_{\mathbf{x}}\,{\mathbf{T}^{(m)}}^{\top}\ \in\ \mathbb{R}^{N\times N},
\]
as $\mathbf{C}_{\mathbf{x}}$-independent constants\revise{; since $\sum_m\lambda_m=1$,} the objective reduces to
\[
\revise{\begin{aligned}\mathcal{J}(\mathbf{C}_{\mathbf{x}})\ ={}&\hat{\mathbf{p}}^{\top}(\mathbf{C}_{\mathbf{x}}\odot\mathbf{C}_{\mathbf{x}})\hat{\mathbf{p}}\\
&-2\,\sum_{m=1}^M \lambda_m\,\mathrm{tr}\!\bigl(\mathbf{C}_{\mathbf{x}}^{\top}\mathbf{A}^{(m)}\bigr)+\text{const}.\end{aligned}}
\]
Following the standard GW-barycenter normalization (as in Eq.~(8) of \cite{supp:Gong2022}), we take the derivative with respect to $\mathbf{C}_{\mathbf{x}}$ and set it to zero
\[
\begin{aligned}
&\frac{\partial \mathcal{J}(\mathbf{C}_{\mathbf{x}})}{\partial \mathbf{C}_{\mathbf{x}}}=\revise{2\,(\hat{\mathbf{p}}\hat{\mathbf{p}}^{\top})\odot\mathbf{C}_{\mathbf{x}}-2\sum_{m}\lambda_m\mathbf{A}^{(m)}=}\ \mathbf{0}\\
&\Rightarrow \quad \mathbf{C}_{\mathbf{x}}
\ =\ \Bigl(\sum_{m=1}^M \lambda_m\,\mathbf{A}^{(m)}\Bigr)\ \oslash\ \bigl(\hat{\mathbf{p}}\,\hat{\mathbf{p}}^{\top}\bigr),
\end{aligned}
\]
i.e.
\begin{equation}
\label{supp:eq:gw-bary-update-detailed}
\boxed{
\begin{aligned}
\mathbf{C}_{\mathbf{x}}
&\leftarrow
\widetilde{\mathbf{C}}\oslash
\bigl(\hat{\mathbf{p}}\,\hat{\mathbf{p}}^{\top}\bigr),\\
\widetilde{\mathbf{C}}
&= \sum_{m=1}^M \lambda_m\,
\mathbf{T}^{(m)}\mathbf{K}^{(m)}_{\mathbf{x}}{\mathbf{T}^{(m)}}^{\top}.
\end{aligned}}
\end{equation}
Here $\oslash$ denotes the element-wise division.
}

\section{Datasets and Preprocessing}
\label{supp:appendix:exp}

We detail the exact splitting, windowing, and per-modality preprocessing used in our experiments. 
Unless specified, all randomization uses a fixed seed (\texttt{seed}=42), and splits are performed at the file/trajectory level to avoid leakage.


\paragraph{VFD / VFP (Vision–Force–Depth / Vision–Force–Proprioception).}
\revise{The data contain 3{,}000 trajectories; with 48 samples per trajectory, this gives 144{,}000 samples.} \revise{We split the data at the trajectory level with $\texttt{seed}=42$ into training, validation, and test trajectories at a ratio of 80/10/10 (48 samples per trajectory). Validation data are used for early stopping and checkpoint selection, and test data only for the final evaluation; the TCK model is fitted on the training trajectories only.} 
Each episode has \revise{length $48$}. For time step $t$, we form a fixed history window of length $L$ for force (default $L{=}32$) and read targets at $t{+}1$.
RGB images are center-cropped to $128{\times}128$, normalized by ImageNet statistics (mean $[0.485,0.456,0.406]$, std $[0.229,0.224,0.225]$). 
Depth is stored as $(128,128,1)$, normalized by mean $0.5$/std $0.5$, and used as single-channel tensors.
Force–torque histories are truncated to the last $L$ steps. The resulting tensor has shape $[b{\times}L{\times}6]$. 
Proprioception is parsed from the first $7$ pose components (end effector position/orientation) in the loader and returned as $[b{\times}7]$ at the current step.

\emph{Tasks.}
For \textbf{VFD}, we follow the main text: \revise{(1)} next-step end-effector orientation regression (4D) \revise{and (2) trajectory-pair discrimination (binary classification of whether two samples come from the same trajectory)}, using $(\text{RGB}_t$, $\text{F/T}_{t-L+1:t}$, $\text{Depth}_t)$ as inputs.
For \textbf{VFP}, we perform next-step contact prediction (binary) using $(\text{RGB}_t$, $\text{F/T}_{t-L+1:t}$, $\text{Proprio}_t)$; class balance is enforced by uniform sampling across trajectories. 
\begin{figure}[h!]
    \centering
    \includegraphics[width=\columnwidth]{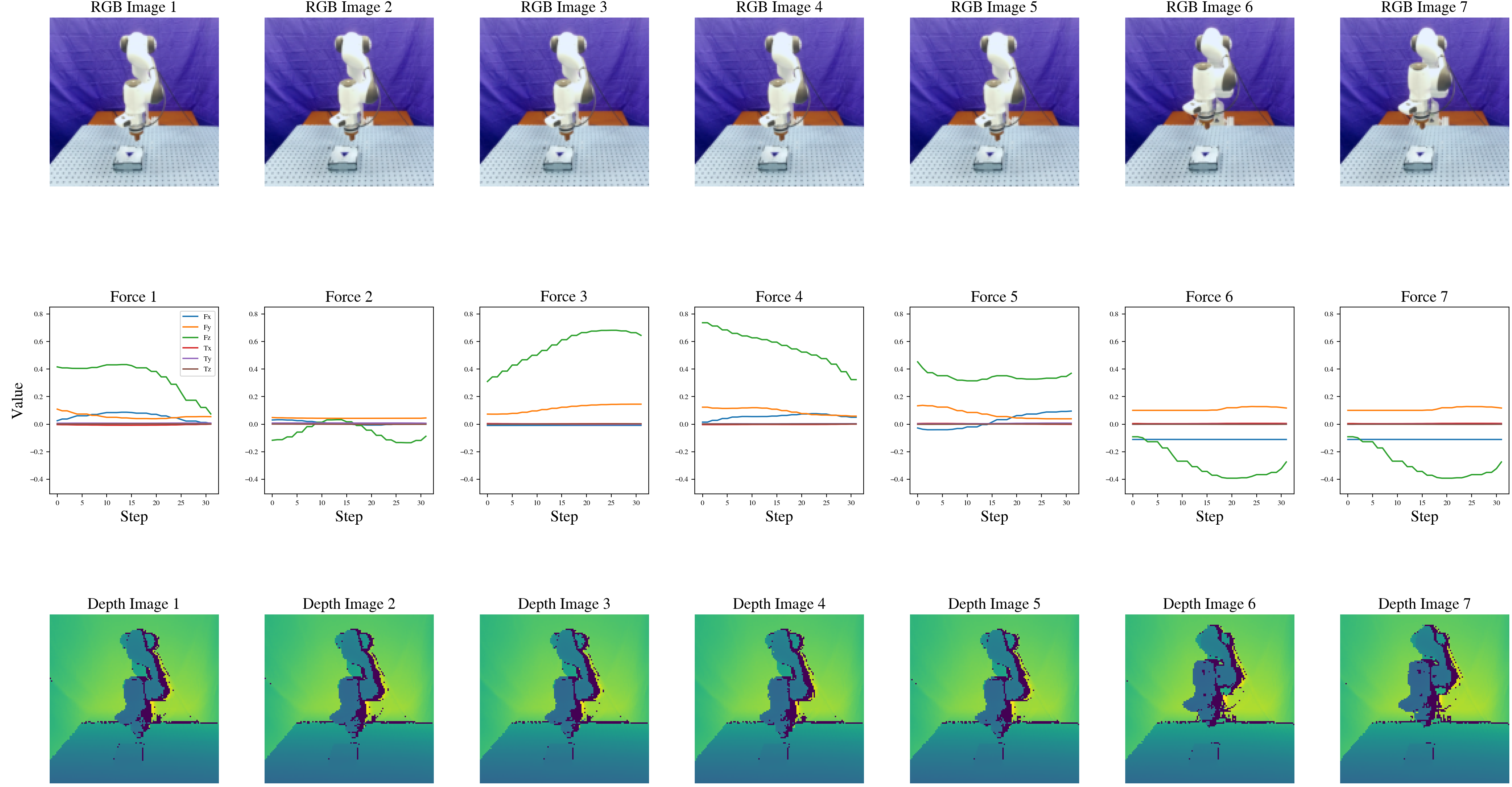}
    \caption{\textbf{VFD overview.} Synchronized windows of RGB images, force–torque signals (last $L{=}32$ steps), and depth camera images. Images are center-cropped to $128{\times}128$ and normalized. Depth images are normalized with mean/std $0.5$.}
    \label{supp:fig:dataset}
\end{figure}

\paragraph{MuJoCo Push.}
A planar pushing task with a Franka Panda arm interacting with a puck. 
\revise{The dataset contains 1{,}310 trajectories of 240 time steps, of which 1{,}000 are used for training, 10 for validation, and 300 for testing.}
\emph{Image modality:} we use sequences of grayscale frames. Each sample contains a length $S=32$ subsequence of $32{\times}32$ frames, forming tensors of shape $(B,S,1,32,32)$.
\revise{The force modality concatenates the current force--torque signal and the contact flag, forming tensors of shape $(B,7)$, and the end-effector position modality forms $(B,3)$.}

\paragraph{VAT (Vision–Audio–Tactile).}
We assemble object-level triplets from per-class folders. 
\revise{The data cover 20 objects with 425, 75, and 75 samples per object and modality for training, validation, and testing (8{,}500, 1{,}500, and 1{,}500 in total); 12 tuples whose visual image is blank are excluded. The tactile observations are the GelSight images rendered by ObjectFolder~2.0 ($160{\times}120$), and each audio clip lasts 3\,s at 44.1\,kHz.}
We use predefined train / validate / test directory structures over a fixed object list. Labels for the retrieval tasks are integer-encoded. 
Visual and tactile images are resized to $246{\times}246$ and normalized by ImageNet statistics.
Audio is loaded at its native sampling rate; at test time, the raw waveform is truncated to \texttt{TARGET\_LENGTH} $=132{,}300$ samples. The final shape of the tensors is \revise{$(B,132300)$.}

\emph{Task.} Cross-modal retrieval with relevance at the object identity level; we report direction-specific MAP on the test set.

\paragraph{VIP (Vision--IMU--Proprioception).}
To complement benchmark and simulation-based evaluations, we further assess UniOMA on a real-world robot-centric dataset collected on a physical 6-DoF robotic manipulator.
This dataset is designed to evaluate whether the learned multimodal structural alignment generalizes to noisy \revise{and} embodiment-specific sensing commonly encountered in real robotic systems.

\revise{\emph{Data collection.}
The robot is a commercial 6-DoF serial manipulator with a reach of 900\,mm, operated without payload and without contact.
Each trajectory is a reaching motion toward a sampled end-effector target, executed by a trained reaching policy that commands joint increments of at most $10^\circ$ per step.
The initial joint configuration is sampled uniformly from $[\pm180^\circ,\pm30^\circ,\pm90^\circ,\pm180^\circ,\pm180^\circ,\pm180^\circ]$, so that the trajectories cover varied initial configurations and reaching targets.
All data were collected in a single workspace under constant indoor lighting.

\emph{Scale and split.}
The dataset contains 1{,}000 trajectories of 100 steps recorded at 5\,Hz (20\,s each), i.e., 100{,}000 synchronized samples.
We split it at the trajectory level into training, validation, and test trajectories at a ratio of 80/10/10 (80{,}000/10{,}000/10{,}000 samples); training and test trajectories are drawn from the same distribution, and there are no held-out conditions.}

\emph{Modalities and preprocessing.}
Each sample consists of three synchronized modalities:
(1) a third-person RGB image captured from a fixed external camera \revise{on a tripod about 2\,m from the robot at $320{\times}240$}, resized to $64{\times}64$ and normalized using ImageNet statistics \revise{(the end-effector is not visible in every frame)};
(2) proprioceptive signals given by the current joint angles of the 6-DoF arm, represented as a $[b{\times}6]$ vector;
and (3) inertial measurements (IMU) from \revise{the six joint modules, each reporting three linear and three angular accelerations}, forming tensors of shape $[b{\times}6{\times}6]$.
\revise{Each of the 36 inertial channels is standardized with the mean and variance of the training trajectories.}
\revise{The three modalities are hardware-triggered and recorded as single-step readings at the 5\,Hz control frequency, without temporal windowing.}

\emph{Task: end-effector position regression.}
Given the multimodal input $(\text{RGB}_t, \text{Proprio}_t, \text{IMU}_t)$, the model is trained to predict the current Cartesian position of the end-effector (regression, 3D).
\revise{The ground-truth end-effector position is measured by an optical motion-capture system from markers rigidly attached to the end-effector and is expressed in a table-fixed frame; it is not computed from the joint encoders.}
We report the \revise{mean Euclidean distance between the predicted and measured end-effector positions ($\times10^{-2}$\,m)} on the held-out test set, as summarized in Table III of the main paper.

\begin{figure}[h!]
    \centering
    \includegraphics[width=0.7\columnwidth]{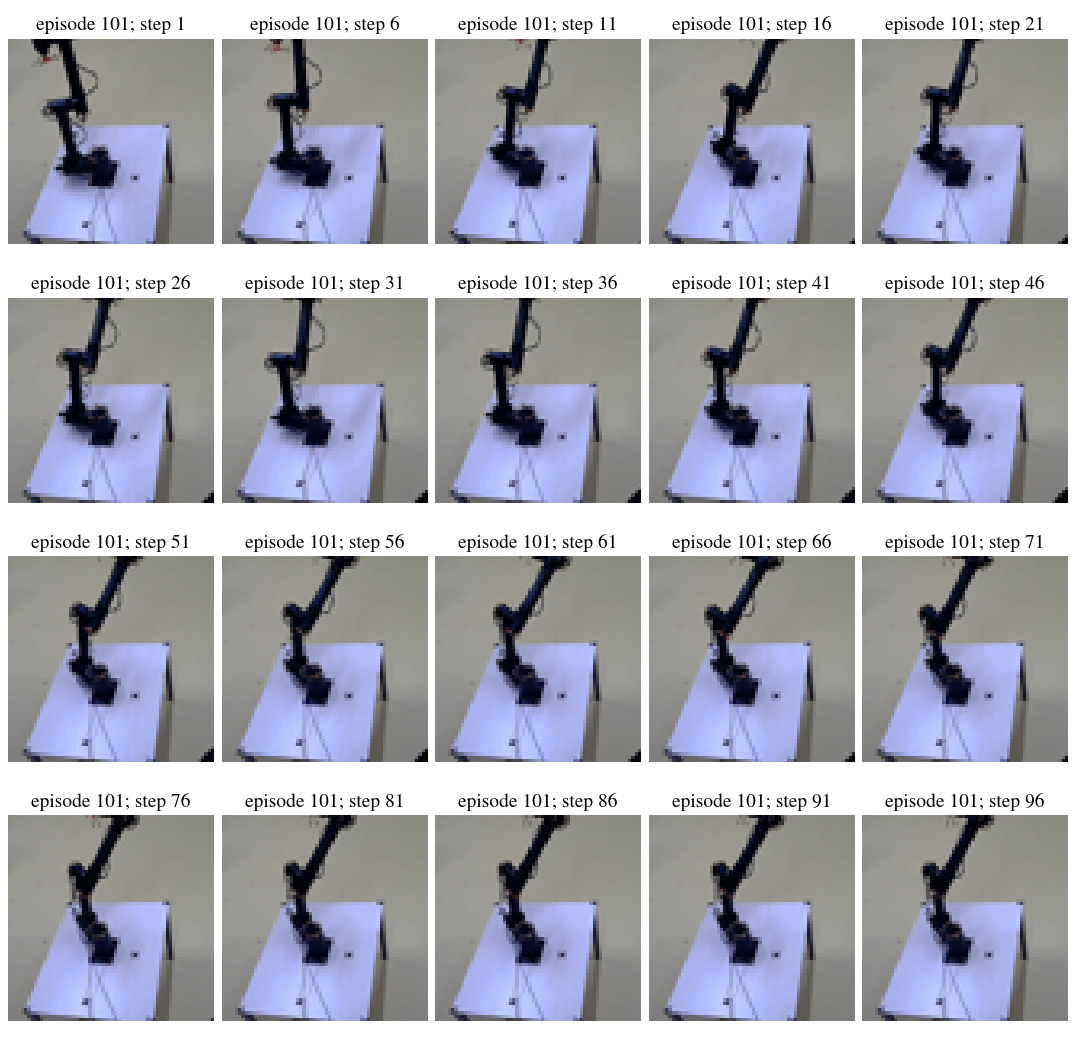}
    \caption{\textbf{VIP dataset overview.} Example RGB observations from the real-world robot dataset.
    Images are captured by a third-person camera while the robot executes \revise{reaching} trajectories.
    These visual observations are synchronized with joint proprioception and IMU signals.}
    \label{supp:fig:vip_dataset}
\end{figure}

\paragraph{MultiBench 4--7 Modality Scalability.}
To evaluate whether UniOMA scales beyond three modalities, we construct an
additional downstream classification benchmark using a multi-sensor subset of
the MultiBench dataset. This setting allows us to progressively increase the
number of modalities and test whether the learned structural consensus remains
stable as modality count grows.

\emph{Modalities.}
We select seven heterogeneous sensing streams commonly used in robotic
manipulation: 
\begin{align*}
    &\text{Vision (RGB)}: [3{\times}128{\times}128], \qquad
    \text{Depth}: [1{\times}128{\times}128], \\
    &\text{Force--Torque}: [6], \quad
    \text{Proprioception}: [7],\quad
    \text{Action}: [d_a], \\
    &\text{Contact state}: [1],\quad
    \text{Optical Flow}: [2{\times}128{\times}128].
\end{align*}
For a given experiment with $M\in\{4,5,6,7\}$ modalities, we take the first $M$
modalities from this list. All modalities are independently normalized using
training-set statistics following the MultiBench protocol.

\emph{Task: trajectory-consistency classification.}
Given two multimodal or single-modal samples, the model must classify whether
they originate from the \textbf{same trajectory}. Positive pairs are sampled
from two timesteps of the same trajectory; negatives are sampled across distinct
trajectories. This task directly evaluates whether embeddings preserve the
trajectory-level structure across multiple modalities.

\section{Additional Experiments}

\subsection{\revise{Structural Metrics and Cross-modal Transfer}}
\label{supp:supp:structural}
\begin{figure}[h!]
\centering
\includegraphics[width=0.8\textwidth]{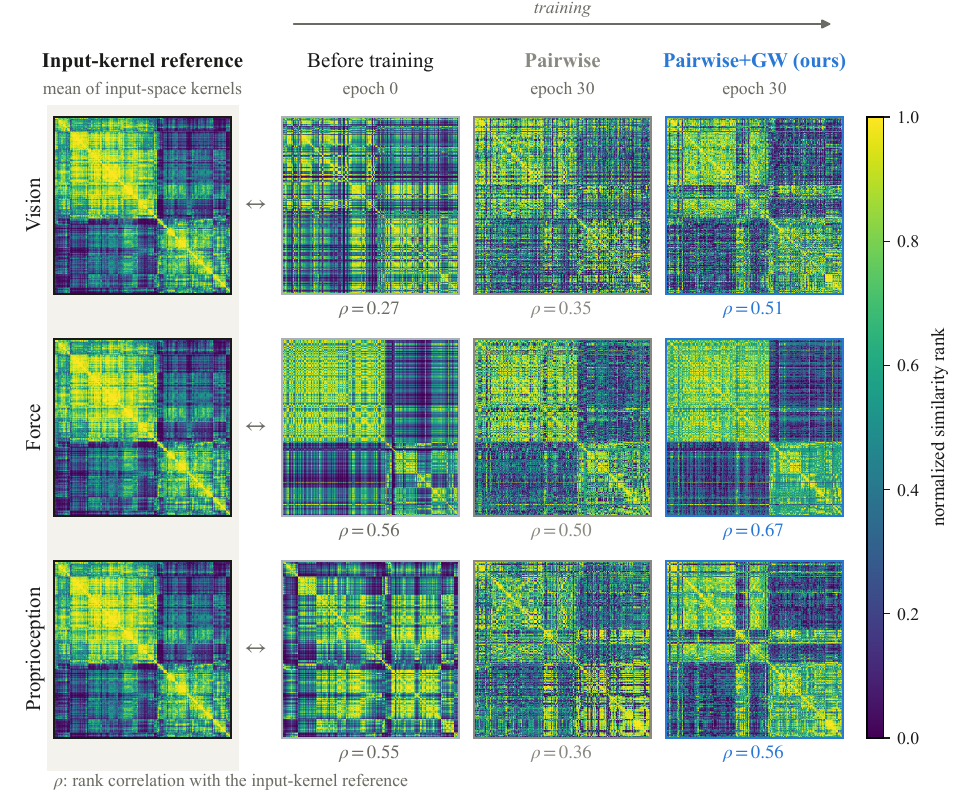}
\caption{\revise{\textbf{Similarity structure before and after training on VFP.} Each row is one modality. Left: the \revise{input-kernel reference $\bar{\mathbf K}$} (mean of the input-space kernels) of 256 held-out test samples, repeated in every row. Right: the embedding-space similarity matrix of that modality at epoch 0 and after 30 epochs of Pairwise training or training with the GW term; $\rho$ is the rank correlation with \revise{$\bar{\mathbf K}$}. All matrices share one sample order and show ranks of pairwise similarity.}}
\label{supp:fig:structure-before-after}
\end{figure}

\revise{All models of Table~V of the main paper (including the relational controls and the direct pairwise GW baseline) and of the weight analyses in Sec.~IV-F of the main paper (mispaired images; learned vs.\ uniform weights) are trained on VFD and VFP for a fixed 30 epochs and evaluated at the final checkpoint, over five seeds (42--46). 
The structural measures in Table~V of the main paper use 256 fixed test samples: the frames at time steps 0, 12, 24, and 36 of 64 test trajectories. For each modality, the embedding-space kernel is an RBF kernel with the median bandwidth on the normalized projected embeddings. 
\revise{As the evaluation reference we use the arithmetic mean of the input-space kernels of the same samples,} $\bar{\mathbf{K}}=\frac{1}{M}\sum_{m}\mathbf{K}^{(m)}_{\mathbf{x}}$ \revise{(the input-kernel reference), not the GW barycenter $\mathbf{C}^*_{\mathbf{x}}$ used in training; $\bar{\mathbf K}$} does not depend on the trained encoders \revise{or the modality weights} and is the same for all methods. Each measure is averaged over the modalities.}
\begin{itemize}
\item \revise{\textbf{Rank correlation:} Spearman correlation between the upper-triangular off-diagonal entries of the two matrices.}
\item \revise{\textbf{Neighborhood preservation (kNN-10):} the average fraction of each sample's 10 nearest neighbors (excluding itself) that the two matrices share.}
\item \revise{\textbf{GW distance:} the squared-loss GW distance to the input-kernel reference.}
\item \revise{\textbf{Clustering consistency (ARI):} the adjusted Rand index between eight-cluster spectral clusterings of the embedding kernel and of \revise{$\bar{\mathbf K}$}.}
\end{itemize}
\revise{The training curves of the GW distance (Fig.~6 of the main paper) evaluate the same samples and \revise{reference $\bar{\mathbf K}$} at epochs 0, 1, 2, 5, 10, 20, and 30 over three seeds. Fig.~\ref{supp:fig:structure-before-after} shows the similarity matrices on VFP before and after training, from one training seed; their rows and columns are ordered by hierarchical clustering of \revise{$\bar{\mathbf K}$} and show ranks of pairwise similarity, so that colors are comparable across panels and do not depend on the kernel scale.}

\begin{figure}[h!]
\centering
\includegraphics[width=0.9\textwidth]{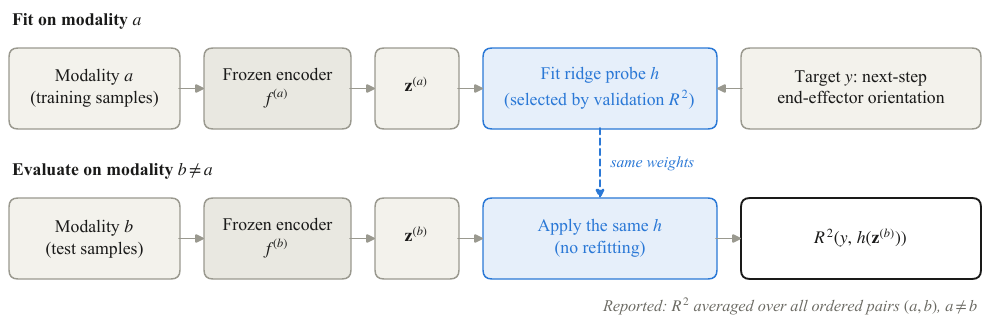}
\caption{\revise{\textbf{Cross-modal transfer protocol.} A ridge probe for the next-step end-effector orientation is fitted on the frozen projected embeddings of modality $a$ and applied with the same weights to the test embeddings of another modality $b$. A high $R^2$ indicates that the two modalities place the samples consistently in the shared embedding space.}}
\label{supp:fig:transfer-protocol}
\end{figure}

\revise{For cross-modal transfer, the projected embeddings are frozen. A ridge probe for next-step end-effector orientation is fitted on the training embeddings of one modality, with its regularization selected from $\{10^{-6},\dots,10^{-1}\}$ by validation $R^2$ on the same modality, and is then applied without refitting to the test embeddings of every other modality. We report the $R^2$ averaged over all ordered modality pairs. Fig.~\ref{supp:fig:transfer-protocol} illustrates the protocol.}

\subsection{Hyper-parameter Analysis}
\label{supp:appendix:hparam_analysis}

\paragraph{Number of GW barycenter iterations \(T_{\max}\).}
Let \(\mathbf{C}^{(t)}\) be the consensus at inner-loop iteration \(t\) in Alg.~3 of the main paper. We monitor the relative Frobenius change
\(\Delta_t=\|\mathbf{C}^{(t)}-\mathbf{C}^{(t-1)}\|_F/\|\mathbf{C}^{(t-1)}\|_F\)
and the trace objective
\begin{equation}
\sum_m \lambda_m\,\mathrm{tr}\!\big(\mathbf{C}^{(t)\top}\mathbf{T}^{(m)\top}\mathbf{K}^{(m)}_{\mathbf{x}}\mathbf{T}^{(m)}\big).
\end{equation}
Both stabilize rapidly; after \(t{=}5\) further changes are negligible (\(\Delta_t<10^{-3}\)).
We therefore fix \(T_{\max}=5\) for all reported results.

\begin{figure}[h!]
    \centering
    \includegraphics[width=0.8\columnwidth]{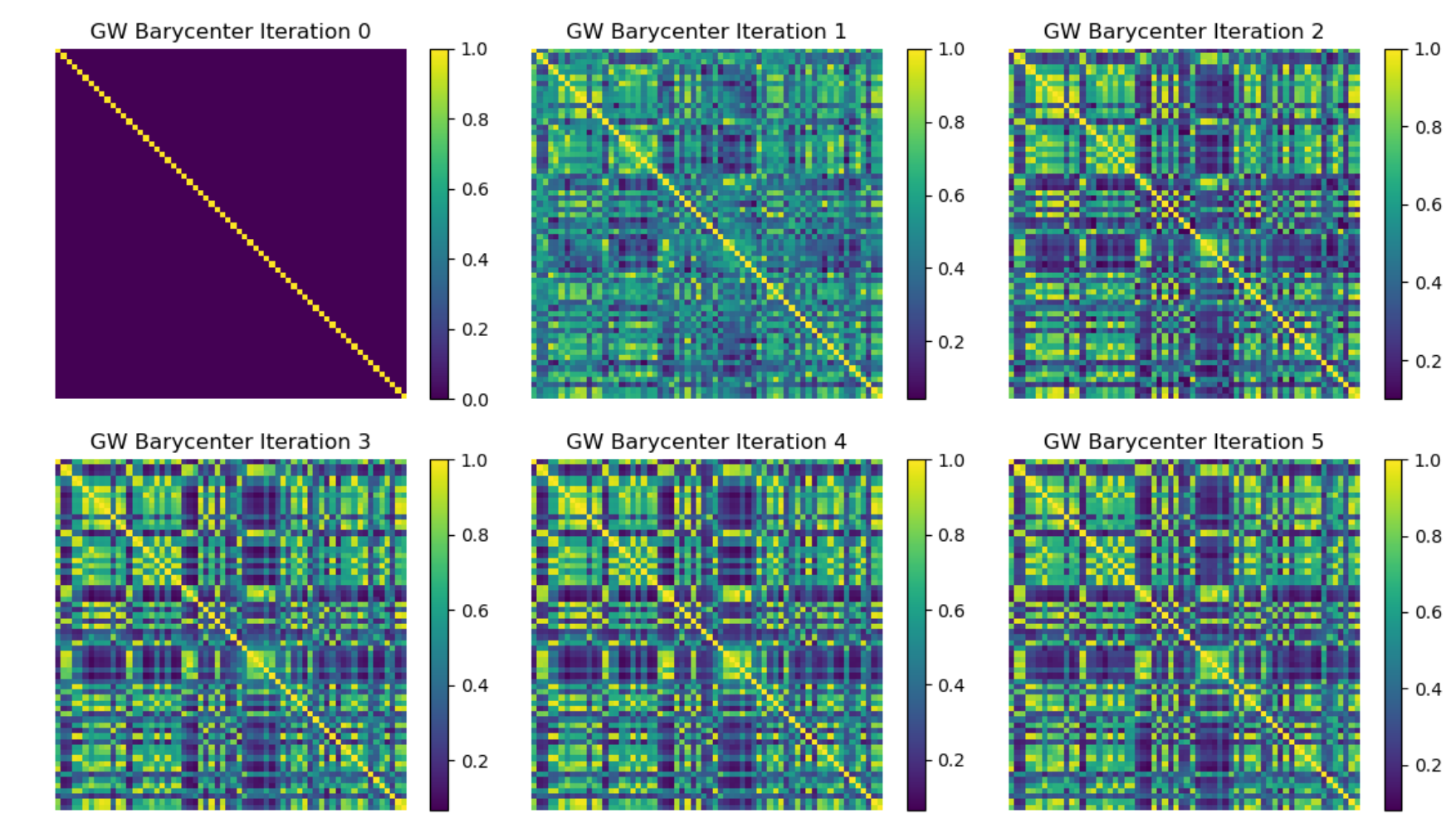}
    \caption{GW barycenter inner-loop: structural consensus across iterations \(t\) in Alg.~3 of the main paper.
    By \(t{=}5\), both geometry and objective are effectively stable; we therefore choose \(T_{\max}{=}5\) (batch size 64).}
    \label{supp:fig:bc}
\end{figure}

\subsection{Ablation: Hyper-parameter Effects}
\label{supp:appendix:hparam_ablation}
We empirically ablate \revise{two} key hyper-parameters of UniOMA---the GW regularization weight \revise{$\alpha$} and the number of barycenter iterations $T_{\max}$.
Table~\ref{supp:tab:hyper-ablation} summarizes results on the VFD classification task (Top-1 accuracy,
mean~$\pm$~std over 10 seeds). Performance is stable across a broad range of values around our
default settings; too small \revise{$\alpha$} under-utilizes structural alignment and too large
\revise{$\alpha$} marginally over-regularizes the embeddings. The solver iteration number $T_{\max}$ shows
a clear plateau around $T_{\max}=5$, confirming that a small number of GW barycenter iterations
is sufficient in practice.

\begin{table}[t]
\centering
\caption{
Ablation of UniOMA hyper-parameters on the VFD
classification task (Top-1 accuracy, \%, mean~$\pm$~std over 10 seeds).
We vary the GW weight $\alpha$ and the number of
barycenter iterations $T_{\max}$ around the default settings. UniOMA is robust across a wide range of values; our default
choices (in \textbf{bold}) lie near the center of each stable regime.
}
\label{supp:tab:hyper-ablation}
\small
\begin{tabular}{lcc}
\toprule
\multicolumn{3}{c}{\textbf{GW weight $\alpha$}} \\
\midrule
$\alpha = 200$   & $92.12 \pm 0.02$ & \\
$\alpha = 500$   & $92.32 \pm 0.02$ & \\
$\boldsymbol{\alpha = 1000}$  & $92.44 \pm 0.02$ & \\
$\alpha = 2000$  & $92.40 \pm 0.04$ & \\
$\alpha = 5000$  & $92.28 \pm 0.05$ & \\
\midrule
\multicolumn{3}{c}{\textbf{Barycenter iterations $T_{\max}$}} \\
\midrule
$T_{\max} = 3$    & $92.05 \pm 0.03$ & \\
$T_{\max} = 4$    & $92.27 \pm 0.03$ & \\
$\boldsymbol{T_{\max} = 5}$    & $92.44 \pm 0.02$ & \\
$T_{\max} = 6$    & $92.45 \pm 0.02$ & \\
$T_{\max} = 7$    & $92.44 \pm 0.02$ & \\
\bottomrule
\end{tabular}
\end{table}

\subsection{Additional Qualitative Visualizations}
\label{supp:app:vis}

\begin{figure}[t]
\centering
\includegraphics[width=\linewidth]{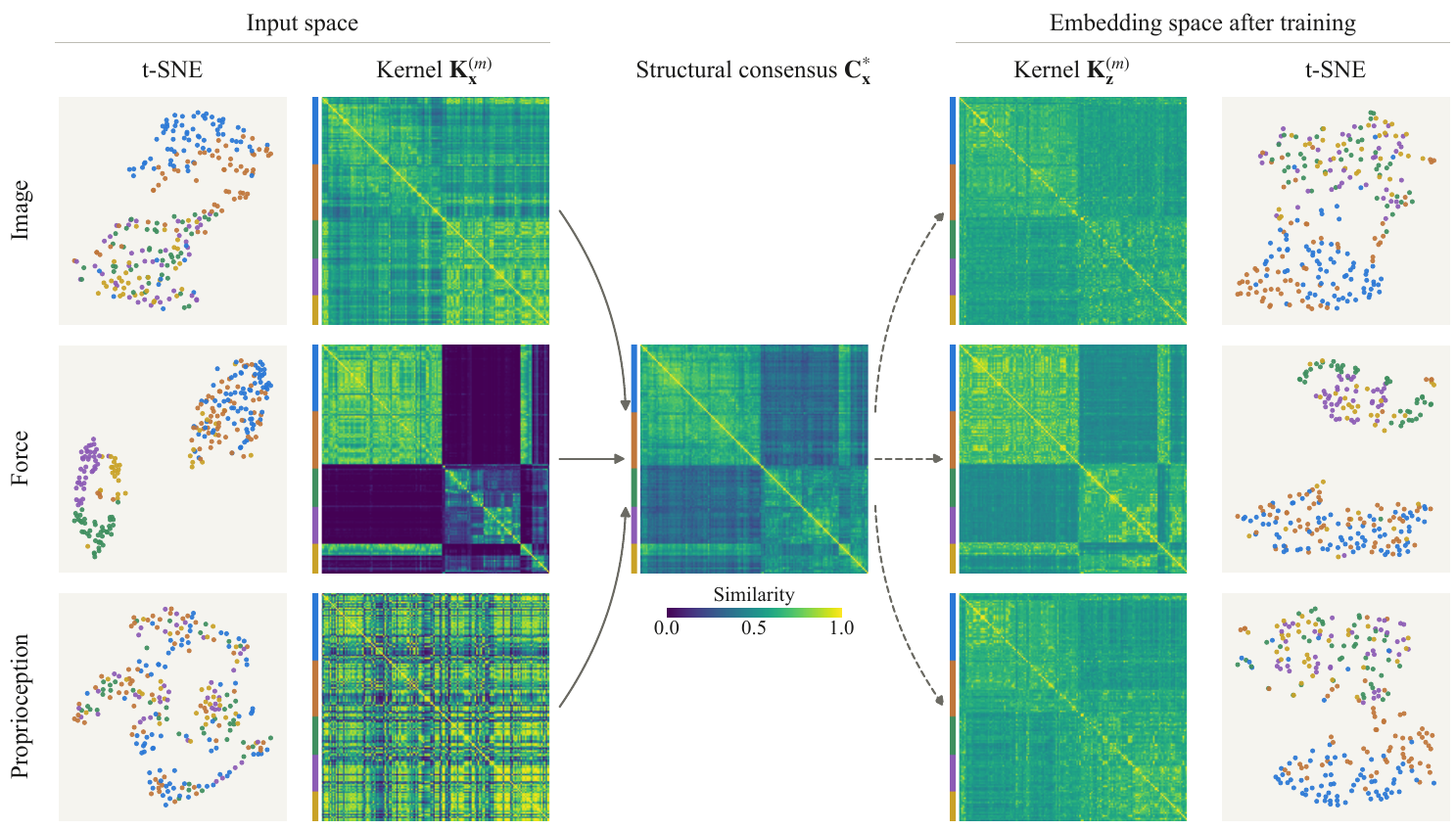}
\caption{\revise{Qualitative visualizations of UniOMA on the VFP benchmark. We choose a mini-batch with size 256 to illustrate the interpretation of the structural consensus and the aligned embeddings. Rows correspond to the modalities (image, force, and proprioception).
\textbf{First column:} t-SNE visualizations of the input modalities, showing clear sub-cluster structures (colors indicate groups of samples that are similar in the consensus).
\textbf{Second column:} Input-space similarity matrices $\mathbf{K}^{(m)}_{\mathbf{x}}$, showing block structure and modality-specific artifacts.
\textbf{Middle:} The structural consensus $\mathbf{C}^{*}_{\mathbf{x}}$, which preserves the shared block structure while smoothing modality-specific noise.
\textbf{Fourth column:} Embedding-space similarity matrices $\mathbf{K}^{(m)}_{\mathbf{z}}$ after training, which show the block structure of the consensus for all three modalities.
\textbf{Last column:} t-SNE of the learned embeddings, where the groups form coherent clusters across modalities, indicating successful structural alignment beyond pairwise correspondence.
All matrices share the same sample order, marked by the color strip on their left.}}
\label{supp:fig:vis}
\end{figure}

For the benefit of the reader, we provide further qualitative visualizations of the
structural alignment induced by UniOMA. Fig.~\ref{supp:fig:vis} summarizes \revise{five}
aspects on \revise{VFP}: \revise{(a) a t-SNE embedding of the input space of each modality;} \revise{(b)} input-space similarity matrices for each
modality; \revise{(c)} the corresponding \revise{structural consensus}; \revise{(d) embedding-space similarity matrices after training; and (e)} a t-SNE embedding
of the learned representations.
Together, these views illustrate how UniOMA preserves shared structure while
remaining interpretable.

\subsection{\revise{Entropy Regularization of the Modality Weights}}
\label{supp:supp:entropy}
\begin{figure}[h!]
\centering
\includegraphics[width=0.9\textwidth]{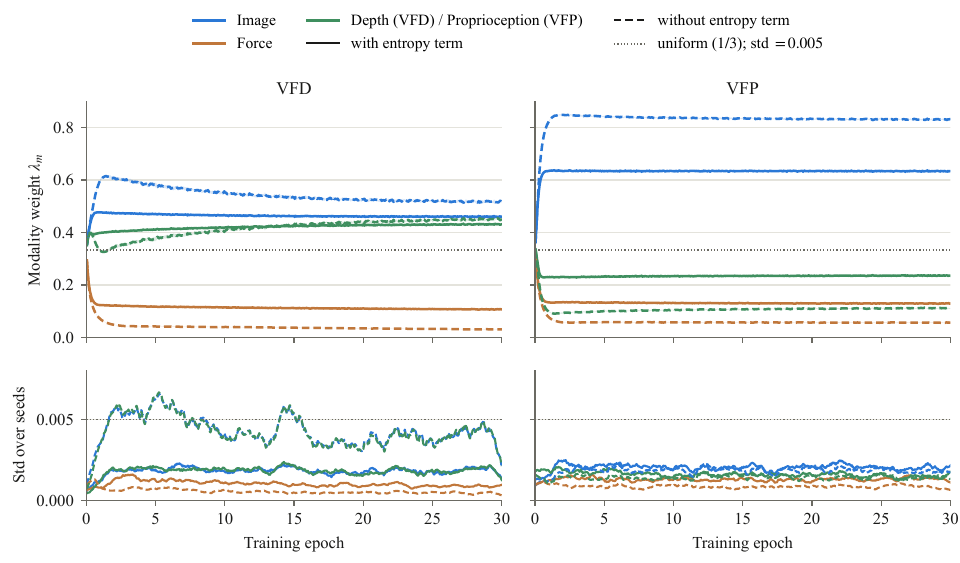}
\caption{\revise{\textbf{Modality weights during training with and without the entropy term.} Same runs as Table~\ref{supp:tab:entropy-ablation} (VFD and VFP, 30 epochs). Top: lines show the mean and bands $\pm1$ standard deviation over five seeds; the dotted line marks uniform weights. Bottom: the standard deviation over the five seeds (running mean over one epoch); the dotted line marks 0.005.}}
\label{supp:fig:weights-training}
\end{figure}

\revise{The training objective includes the entropy term $\beta\sum_{m}\lambda_m\log(\lambda_m+\varepsilon)$ (Sec.~III-D of the main paper).
Table~\ref{supp:tab:entropy-ablation} compares the learned modality weights on VFD and VFP with and without this term after 30 epochs.
Without the entropy term, the largest weight increases from 0.46 to 0.52 on VFD and from 0.63 to 0.83 on VFP, and the force weight decreases from 0.11 to 0.03 and from 0.13 to 0.06.
The term therefore reduces weight concentration.
Fig.~\ref{supp:fig:weights-training} shows the weights during training in the same runs. 
With the entropy term, the weights leave the uniform initialization within the first epoch and stay within 0.02 of their final values after 6 epochs on VFD and 1 epoch on VFP. 
Without the term, the vision weight rises further, increases over 0.6 early in training, and stabilizes at 0.52 on VFD.}

\begin{table}[h!]
\centering
\caption{\revise{\textbf{Final modality weights with and without the entropy term.} Mean $\pm$ std over five seeds after 30 epochs.}}
\label{supp:tab:entropy-ablation}
\small
{\color{revisionblue}
\begin{tabular}{llcc}
\toprule
Dataset & Modality & With entropy & Without entropy \\
\midrule
\multirow{3}{*}{VFD} & Vision & 0.460{\scriptsize$\pm$0.001} & 0.518{\scriptsize$\pm$0.001} \\
 & Force & 0.108{\scriptsize$\pm$0.001} & 0.032{\scriptsize$\pm$0.000} \\
 & Depth & 0.432{\scriptsize$\pm$0.001} & 0.451{\scriptsize$\pm$0.001} \\
\midrule
\multirow{3}{*}{VFP} & Vision & 0.634{\scriptsize$\pm$0.002} & 0.832{\scriptsize$\pm$0.001} \\
 & Force & 0.130{\scriptsize$\pm$0.001} & 0.057{\scriptsize$\pm$0.001} \\
 & Proprioception & 0.236{\scriptsize$\pm$0.002} & 0.111{\scriptsize$\pm$0.001} \\
\bottomrule
\end{tabular}
}
\end{table}

\subsection{\revise{Learning Curves of Peg Insertion}}
\label{supp:supp:peg-curves}
\begin{figure}[h!]
\centering
\includegraphics[width=0.9\textwidth]{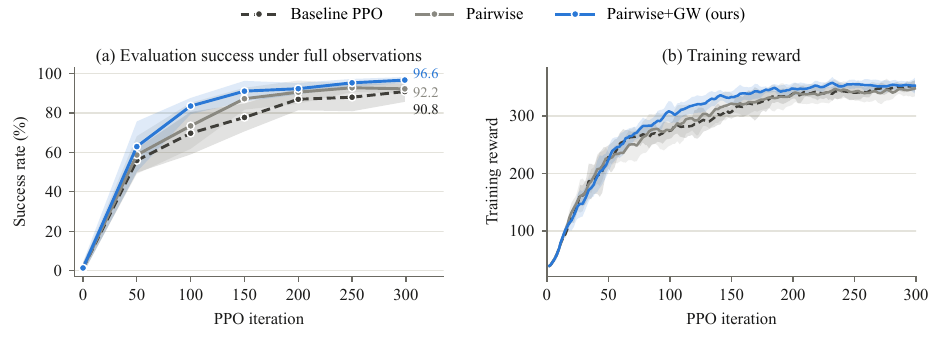}
\caption{\revise{\textbf{Peg-insertion learning curves under full observations.} (a) Success rate of the deterministic policy at the saved checkpoints; the numbers give the final values. (b) Training reward per PPO iteration, smoothed over five iterations. Lines show the mean and bands $\pm1$ standard deviation over five seeds.}}
\label{supp:fig:peg-curves}
\end{figure}

\revise{Fig.~\ref{supp:fig:peg-curves} shows the learning progress of the peg-insertion task under full observations (Table~IV of the main paper), over five seeds.
The success rate is evaluated with the deterministic policy on 512 episodes at iterations 0, 50, 100, 150, 200, and 250 and at the final iteration; the training reward is logged at every PPO iteration.
Pairwise+GW has the highest mean success rate at every checkpoint from iteration 50 onward and ends at 96.6\%, compared with 92.2\% for Pairwise and 90.8\% for Baseline PPO; at the end of training, the training rewards of the three methods are similar.}

\subsection{\revise{Information-Retention Analysis}}
\label{supp:supp:posthoc}
{\color{revisionblue}
A natural concern with structural alignment is that enforcing a shared geometry may suppress information from a more informative modality.
We examine this in \revise{VFD and} VFP by treating image embeddings as the information-rich modality $Z_R$ and force embeddings as the information-poor modality $Z_P$.
We freeze the trained Pairwise/UniOMA encoders and fit lightweight probes $\hat f_\phi$ from $Z_R$, $Z_P$, and $[Z_R,Z_P]$ to \revise{two targets $Y$: the next-step end-effector orientation (four-dimensional, ridge regression) and the binary contact label (logistic regression)}.
\revise{For the orientation, we use} the held-out Gaussian residual \revise{measure}~\cite{supp:10.5555/2981345.2981371,supp:poole2019variational}:
\begin{equation}
\widehat I_{\mathrm{probe}}(Z;Y)=\max\!\left\{0,\,\frac{1}{2}\log_2\frac{\det(\widehat\Sigma_Y)}{\det(\widehat\Sigma_{Y-\hat f_\phi(Z)})}\right\},
\label{supp:eq:probe-info}
\end{equation}
where $\widehat\Sigma_{Y}$ and $\widehat\Sigma_{Y-\hat f_\phi(Z)}$ denote the covariance matrices of the target and the prediction residual, respectively.
\revise{This measure is a mutual-information lower bound only for a Gaussian target, so we read it as task information that a linear probe can decode rather than as $I(Z;Y)$.

For contact, we report $\widehat H(Y)-\widehat{\mathrm{CE}}$ on the test split, where $\widehat H(Y)$ is the empirical entropy of the binary contact label and $\widehat{\mathrm{CE}}=-\frac{1}{n}\sum_{i=1}^{n}\log_2 q_\phi(y_i\mid z_i)$ is the cross-entropy of the logistic probe $q_\phi$ on the $n$ test samples. This quantity estimates the variational lower bound $I(Z;Y)\ge H(Y)-\mathrm{CE}$~\cite{supp:10.5555/2981345.2981371} on held-out data, and it is at most $\widehat H(Y)=0.955$ bits.
Both quantities are instances of the bound $I(Z;Y)\ge H(Y)-\mathrm{CE}$ with a decoder matched to the target: a Gaussian decoder for the continuous orientation, which gives Eq.~\eqref{supp:eq:probe-info} when $H(Y)$ is replaced by its Gaussian value, and a Bernoulli (logistic) decoder for the binary contact label.}

Here, ``rich information'' refers to task-relevant predictive information in the embedding, rather than total information in the raw signal.
\revise{Table~\ref{supp:tab:posthoc-information} shows that, compared with Pairwise, UniOMA increases the image-only and joint orientation measures by $1.6\%$ on VFP and by $0.8$--$0.9\%$ on VFD, and changes the contact measures by at most $1.5\%$.}
This suggests that alignment with the information-poor force modality does not degrade the information-rich image modality.
}

\begin{table*}[h!]
\centering
\caption{
\textbf{Task-information retention on \revise{VFD and} VFP.}
\revise{GW alignment does not reduce task information in the image features: compared with Pairwise, UniOMA increases orientation information in image features by 0.8--1.6\% and changes the contact information of each feature set by at most 1.5\%.}
We report the information (in bits) about the target that a probe recovers from frozen encoder features: the Gaussian residual measure (Eq.~\eqref{supp:eq:probe-info}) for the next-step end-effector orientation, and the empirical $\widehat H(Y)-\widehat{\mathrm{CE}}$ for the binary contact label, which is at most $\widehat H(Y)=0.955$.
We fit the probes on the training trajectories, regularize on validation, and evaluate once on the test split (mean $\pm$ std over 5 seeds); $\Delta$ is the relative change with respect to Pairwise.
\colorbox{gray!10}{Gray columns} denote our method.}
\label{supp:tab:posthoc-information}
\small
\resizebox{0.8\textwidth}{!}{
\color{revisionblue}
\begin{tabular}{l l c >{\columncolor{gray!10}}c c c >{\columncolor{gray!10}}c c}
\toprule
\multirow{2}{*}{Dataset} & \multirow{2}{*}{Features} &
\multicolumn{3}{c}{Next-step end-effector orientation (bits)} &
\multicolumn{3}{c}{Contact (bits, at most 0.955)} \\
\cmidrule(lr){3-5}\cmidrule(lr){6-8}
& & Pairwise & Ours & $\Delta$ & Pairwise & Ours & $\Delta$ \\
\midrule
\multirow{3}{*}{VFD} & Image & 11.60{\scriptsize$\pm$0.04} & 11.69{\scriptsize$\pm$0.04} & $+0.8\%$ & 0.549{\scriptsize$\pm$0.009} & 0.541{\scriptsize$\pm$0.003} & $-1.5\%$ \\
 & Force & 0.59{\scriptsize$\pm$0.01} & 0.62{\scriptsize$\pm$0.04} & $+5.1\%$ & 0.701{\scriptsize$\pm$0.004} & 0.698{\scriptsize$\pm$0.008} & $-0.4\%$ \\
 & Image + force & 11.71{\scriptsize$\pm$0.03} & 11.81{\scriptsize$\pm$0.05} & $+0.9\%$ & 0.718{\scriptsize$\pm$0.004} & 0.715{\scriptsize$\pm$0.008} & $-0.4\%$ \\
\midrule
\multirow{3}{*}{VFP} & Image & 12.18{\scriptsize$\pm$0.06} & 12.37{\scriptsize$\pm$0.07} & $+1.6\%$ & 0.548{\scriptsize$\pm$0.005} & 0.544{\scriptsize$\pm$0.002} & $-0.7\%$ \\
 & Force & 0.59{\scriptsize$\pm$0.04} & 0.62{\scriptsize$\pm$0.04} & $+4.9\%$ & 0.701{\scriptsize$\pm$0.005} & 0.699{\scriptsize$\pm$0.006} & $-0.4\%$ \\
 & Image + force & 12.31{\scriptsize$\pm$0.06} & 12.51{\scriptsize$\pm$0.07} & $+1.6\%$ & 0.719{\scriptsize$\pm$0.003} & 0.713{\scriptsize$\pm$0.005} & $-0.8\%$ \\
\bottomrule
\end{tabular}
}
\end{table*}

\subsection{\revise{Choice of the Pre-trained Feature Extractor}}
\label{supp:supp:teacher}
\revise{The input-space kernel of the image modality is built from frozen ViT-B/16 features (Sec.~\ref{supp:appendix:x}). To test whether the structural gains depend on this choice, we replaced ViT-B/16 by DINOv2 ViT-B/14~\cite{supp:oquab2024dinov2} or an ImageNet-pretrained ResNet-18~\cite{supp:he2016deep}, retrained our method on VFD and VFP (30 epochs, five seeds), and evaluated the structural measures of Sec.~\ref{supp:supp:structural} against the same ViT-based \revise{input-kernel reference} as in Table~V of the main paper. Table~\ref{supp:tab:teacher-swap} shows that the gains over Pairwise are preserved with both feature extractors, and ResNet-18 features give slightly higher rank correlation and lower GW distance than the default. The structural alignment therefore does not rely on a particular pre-trained encoder.}

\begin{table}[h!]
\centering
\caption{\revise{\textbf{Structural alignment with different feature extractors for the image input-space kernel.} All rows are evaluated against the ViT-based \revise{input-kernel reference} of Table~V of the main paper on the same 256 held-out samples (mean $\pm$ std over five seeds).}}
\label{supp:tab:teacher-swap}
\small
{\color{revisionblue}
\begin{tabular}{@{}lcccc@{}}
\toprule
Method & Rank correlation\,$\uparrow$ & kNN-10\,$\uparrow$ & GW dist.\,$\downarrow$ {\scriptsize($\times10^{-2}$)} & ARI\,$\uparrow$ \\
\midrule
\multicolumn{5}{@{}l}{\textbf{VFD}} \\
Pairwise & 0.318{\scriptsize$\pm$0.045} & 0.205{\scriptsize$\pm$0.006} & 2.78{\scriptsize$\pm$0.10} & 0.146{\scriptsize$\pm$0.008} \\
Ours, ViT-B/16 (default) & 0.481{\scriptsize$\pm$0.042} & 0.219{\scriptsize$\pm$0.009} & 2.59{\scriptsize$\pm$0.04} & 0.163{\scriptsize$\pm$0.010} \\
Ours, DINOv2 ViT-B/14 & 0.480{\scriptsize$\pm$0.028} & 0.223{\scriptsize$\pm$0.004} & 2.52{\scriptsize$\pm$0.05} & 0.165{\scriptsize$\pm$0.012} \\
Ours, ResNet-18 & 0.504{\scriptsize$\pm$0.030} & 0.224{\scriptsize$\pm$0.005} & 2.50{\scriptsize$\pm$0.06} & 0.157{\scriptsize$\pm$0.021} \\
\midrule
\multicolumn{5}{@{}l}{\textbf{VFP}} \\
Pairwise & 0.392{\scriptsize$\pm$0.035} & 0.246{\scriptsize$\pm$0.006} & 3.25{\scriptsize$\pm$0.06} & 0.150{\scriptsize$\pm$0.011} \\
Ours, ViT-B/16 (default) & 0.575{\scriptsize$\pm$0.020} & 0.257{\scriptsize$\pm$0.007} & 2.70{\scriptsize$\pm$0.05} & 0.164{\scriptsize$\pm$0.007} \\
Ours, DINOv2 ViT-B/14 & 0.574{\scriptsize$\pm$0.014} & 0.255{\scriptsize$\pm$0.002} & 2.51{\scriptsize$\pm$0.10} & 0.162{\scriptsize$\pm$0.025} \\
Ours, ResNet-18 & 0.606{\scriptsize$\pm$0.010} & 0.259{\scriptsize$\pm$0.005} & 2.46{\scriptsize$\pm$0.06} & 0.153{\scriptsize$\pm$0.013} \\
\bottomrule
\end{tabular}
}
\end{table}

\revise{\textbf{Tactile input-space kernel.}
Tactile images differ substantially from natural RGB images, so we also replaced the ViT-B/16 features of the tactile modality on VAT, while keeping the audio and visual kernels unchanged, with two frozen encoders: an ImageNet-supervised ResNet-50 (global average-pooled 2048-dimensional features; $224{\times}224$ inputs with ImageNet normalization) and the tactile-specific encoder Sparsh~\cite{supp:higuera2024sparsh} (\texttt{sparsh-dino-base}, a ViT-B on two stacked background-subtracted tactile frames, mean of the patch tokens; since VAT touches are single images, the frame is repeated and the mean training touch image serves as the background).
The tactile input-space kernel is an RBF kernel with the median heuristic on these features (Sec.~\ref{supp:appendix:nn}), and retrieval is evaluated as in Table~I of the main paper.
Table~\ref{supp:tab:tactile-swap} shows that retrieval remains above Pairwise in all three directions with both encoders; ResNet-50 features give the highest Vis$\to$Tact and Tact$\to$Aud MAP, and the tactile-specific Sparsh features do not improve over the generic encoders.}

\begin{table}[h!]
\centering
\caption{\revise{\textbf{Cross-modal retrieval on VAT with different feature extractors for the tactile input-space kernel} (MAP, mean $\pm$ std). The Pairwise and ViT-B/16 rows are those of Table~I of the main paper.}}
\label{supp:tab:tactile-swap}
\small
{\color{revisionblue}
\begin{tabular}{@{}lccc@{}}
\toprule
Method & Vis$\to$Aud\,$\uparrow$ & Vis$\to$Tact\,$\uparrow$ & Tact$\to$Aud\,$\uparrow$ \\
\midrule
Pairwise & 0.25{\scriptsize$\pm$0.07} & 0.41{\scriptsize$\pm$0.11} & 0.10{\scriptsize$\pm$0.01} \\
Ours, ViT-B/16 (default) & 0.36{\scriptsize$\pm$0.05} & 0.60{\scriptsize$\pm$0.03} & 0.12{\scriptsize$\pm$0.02} \\
Ours, ResNet-50 (ImageNet) & 0.32{\scriptsize$\pm$0.04} & 0.65{\scriptsize$\pm$0.03} & 0.15{\scriptsize$\pm$0.02} \\
Ours, Sparsh & 0.31{\scriptsize$\pm$0.06} & 0.56{\scriptsize$\pm$0.10} & 0.14{\scriptsize$\pm$0.04} \\
\bottomrule
\end{tabular}
}
\end{table}

\clearpage

%% file: main.bbl
\begin{thebibliography}{10}
\providecommand{\url}[1]{#1}
\csname url@samestyle\endcsname
\providecommand{\newblock}{\relax}
\providecommand{\bibinfo}[2]{#2}
\providecommand{\BIBentrySTDinterwordspacing}{\spaceskip=0pt\relax}
\providecommand{\BIBentryALTinterwordstretchfactor}{4}
\providecommand{\BIBentryALTinterwordspacing}{\spaceskip=\fontdimen2\font plus
\BIBentryALTinterwordstretchfactor\fontdimen3\font minus \fontdimen4\font\relax}
\providecommand{\BIBforeignlanguage}[2]{{%
\expandafter\ifx\csname l@#1\endcsname\relax
\typeout{** WARNING: IEEEtran.bst: No hyphenation pattern has been}%
\typeout{** loaded for the language `#1'. Using the pattern for}%
\typeout{** the default language instead.}%
\else
\language=\csname l@#1\endcsname
\fi
#2}}
\providecommand{\BIBdecl}{\relax}
\BIBdecl

\bibitem{Rombach_2022_CVPR}
R.~Rombach, A.~Blattmann, D.~Lorenz, P.~Esser, and B.~Ommer, ``High-resolution image synthesis with latent diffusion models,'' in \emph{Proceedings of the IEEE/CVF Conference on Computer Vision and Pattern Recognition (CVPR)}, June 2022, pp. 10\,684--10\,695.

\bibitem{Mirza2014}
M.~Mirza and S.~Osindero, ``Conditional generative adversarial nets,'' arXiv preprint arXiv:1411.1784, 2014.

\bibitem{Acosta2022}
J.~N. Acosta, G.~J. Falcone, P.~Rajpurkar, and E.~J. Topol, ``Multimodal biomedical ai,'' \emph{Nature Medicine}, vol.~28, no.~9, pp. 1773--1784, Sep. 2022.

\bibitem{9000872}
D.~Feng, C.~Haase-Schütz, L.~Rosenbaum, H.~Hertlein, C.~Gläser, F.~Timm, W.~Wiesbeck, and K.~Dietmayer, ``Deep multi-modal object detection and semantic segmentation for autonomous driving: Datasets, methods, and challenges,'' \emph{IEEE Transactions on Intelligent Transportation Systems}, vol.~22, no.~3, pp. 1341--1360, 2021.

\bibitem{Steyaert2023}
S.~Steyaert, M.~Pizurica, D.~Nagaraj, P.~Khandelwal, T.~Hernandez-Boussard, A.~J. Gentles, and O.~Gevaert, ``Multimodal data fusion for cancer biomarker discovery with deep learning,'' \emph{Nature Machine Intelligence}, vol.~5, no.~4, pp. 351--362, Apr. 2023.

\bibitem{lee2019making}
M.~A. Lee, Y.~Zhu, K.~Srinivasan, P.~Shah, S.~Savarese, L.~Fei-Fei, A.~Garg, and J.~Bohg, ``Making sense of vision and touch: Self-supervised learning of multimodal representations for contact-rich tasks,'' in \emph{2019 International conference on robotics and automation (ICRA)}.\hskip 1em plus 0.5em minus 0.4em\relax IEEE, 2019, pp. 8943--8950.

\bibitem{Calandra2017}
R.~Calandra, A.~Owens, M.~Upadhyaya, W.~Yuan, J.~Lin, E.~H. Adelson, and S.~Levine, ``The feeling of success: Does touch sensing help predict grasp outcomes?'' 2017.

\bibitem{he2020momentum}
K.~He, H.~Fan, Y.~Wu, S.~Xie, and R.~Girshick, ``Momentum contrast for unsupervised visual representation learning,'' in \emph{Proceedings of the IEEE/CVF conference on computer vision and pattern recognition}, 2020, pp. 9729--9738.

\bibitem{chen2020simple}
T.~Chen, S.~Kornblith, M.~Norouzi, and G.~Hinton, ``A simple framework for contrastive learning of visual representations,'' in \emph{International conference on machine learning}.\hskip 1em plus 0.5em minus 0.4em\relax PMLR, 2020, pp. 1597--1607.

\bibitem{10.5555/3495724.3497510}
J.-B. Grill, F.~Strub, F.~Altch\'{e}, C.~Tallec, P.~H. Richemond, E.~Buchatskaya, C.~Doersch, B.~A. Pires, Z.~D. Guo, M.~G. Azar, B.~Piot, K.~Kavukcuoglu, R.~Munos, and M.~Valko, ``Bootstrap your own latent a new approach to self-supervised learning,'' in \emph{Proceedings of the 34th International Conference on Neural Information Processing Systems}, ser. NIPS '20.\hskip 1em plus 0.5em minus 0.4em\relax Red Hook, NY, USA: Curran Associates Inc., 2020.

\bibitem{chen2021exploring}
X.~Chen and K.~He, ``Exploring simple siamese representation learning,'' in \emph{Proceedings of the IEEE/CVF conference on computer vision and pattern recognition}, 2021, pp. 15\,750--15\,758.

\bibitem{Oord2018}
A.~v.~d. Oord, Y.~Li, and O.~Vinyals, ``Representation learning with contrastive predictive coding,'' arXiv preprint arXiv:1807.03748, 2018.

\bibitem{radford2021learning}
A.~Radford, J.~W. Kim, C.~Hallacy, A.~Ramesh, G.~Goh, S.~Agarwal, G.~Sastry, A.~Askell, P.~Mishkin, J.~Clark \emph{et~al.}, ``Learning transferable visual models from natural language supervision,'' in \emph{International conference on machine learning}.\hskip 1em plus 0.5em minus 0.4em\relax PMLR, 2021, pp. 8748--8763.

\bibitem{girdhar2023imagebind}
R.~Girdhar, A.~El-Nouby, Z.~Liu, M.~Singh, K.~V. Alwala, A.~Joulin, and I.~Misra, ``Imagebind: One embedding space to bind them all,'' in \emph{Proceedings of the IEEE/CVF conference on computer vision and pattern recognition}, 2023, pp. 15\,180--15\,190.

\bibitem{chen2023vast}
S.~Chen, H.~Li, Q.~Wang, Z.~Zhao, M.~Sun, X.~Zhu, and J.~Liu, ``Vast: A vision-audio-subtitle-text omni-modality foundation model and dataset,'' \emph{Advances in Neural Information Processing Systems}, vol.~36, pp. 72\,842--72\,866, 2023.

\bibitem{zhu2023languagebind}
B.~Zhu, B.~Lin, M.~Ning, Y.~Yan, J.~Cui, H.~Wang, Y.~Pang, W.~Jiang, J.~Zhang, Z.~Li \emph{et~al.}, ``Languagebind: Extending video-language pretraining to n-modality by language-based semantic alignment,'' \emph{arXiv preprint arXiv:2310.01852}, 2023.

\bibitem{luo2022clip4clip}
H.~Luo, L.~Ji, M.~Zhong, Y.~Chen, W.~Lei, N.~Duan, and T.~Li, ``Clip4clip: An empirical study of clip for end to end video clip retrieval and captioning,'' \emph{Neurocomputing}, vol. 508, pp. 293--304, 2022.

\bibitem{Nair2022}
S.~Nair, A.~Rajeswaran, V.~Kumar, C.~Finn, and A.~Gupta, ``R3m: A universal visual representation for robot manipulation,'' 2022.

\bibitem{karamcheti2023voltron}
S.~Karamcheti, S.~Nair, A.~S. Chen, T.~Kollar, C.~Finn, D.~Sadigh, and P.~Liang, ``Language-driven representation learning for robotics,'' in \emph{Robotics: Science and Systems (RSS)}, 2023.

\bibitem{10.5555/3524938.3525859}
T.~Wang and P.~Isola, ``Understanding contrastive representation learning through alignment and uniformity on the hypersphere,'' in \emph{Proceedings of the 37th International Conference on Machine Learning}, ser. ICML'20.\hskip 1em plus 0.5em minus 0.4em\relax JMLR.org, 2020.

\bibitem{10.5555/3600270.3601550}
W.~Liang, Y.~Zhang, Y.~Kwon, S.~Yeung, and J.~Zou, ``Mind the gap: understanding the modality gap in multi-modal contrastive representation learning,'' in \emph{Proceedings of the 36th International Conference on Neural Information Processing Systems}, ser. NIPS '22.\hskip 1em plus 0.5em minus 0.4em\relax Red Hook, NY, USA: Curran Associates Inc., 2022.

\bibitem{Kraskov2004}
A.~Kraskov, H.~Stögbauer, and P.~Grassberger, ``Estimating mutual information,'' \emph{Physical Review E}, vol.~69, no.~6, p. 066138, Jun. 2004.

\bibitem{poole2019variational}
B.~Poole, S.~Ozair, A.~Van Den~Oord, A.~Alemi, and G.~Tucker, ``On variational bounds of mutual information,'' in \emph{International conference on machine learning}, 2019, pp. 5171--5180.

\bibitem{Cover2005}
T.~M. Cover and J.~A. Thomas, \emph{Elements of Information Theory}.\hskip 1em plus 0.5em minus 0.4em\relax Wiley, Apr. 2005.

\bibitem{Kinney2014}
J.~B. Kinney and G.~S. Atwal, ``Equitability, mutual information, and the maximal information coefficient,'' \emph{Proceedings of the National Academy of Sciences}, vol. 111, no.~9, pp. 3354--3359, Feb. 2014.

\bibitem{Sermanet2017TimeContrastiveNS}
P.~Sermanet, C.~Lynch, Y.~Chebotar, J.~Hsu, E.~Jang, S.~Schaal, and S.~Levine, ``Time-contrastive networks: Self-supervised learning from video,'' \emph{2018 IEEE International Conference on Robotics and Automation (ICRA)}, pp. 1134--1141, 2017.

\bibitem{pmlr-v211-guo23b}
M.~Guo, Y.~Jiang, A.~E. Spielberg, J.~Wu, and K.~Liu, ``Benchmarking rigid body contact models,'' in \emph{Proceedings of The 5th Annual Learning for Dynamics and Control Conference}, ser. Proceedings of Machine Learning Research, N.~Matni, M.~Morari, and G.~J. Pappas, Eds., vol. 211.\hskip 1em plus 0.5em minus 0.4em\relax PMLR, 15--16 Jun 2023, pp. 1480--1492.

\bibitem{Lee2020}
M.~A. Lee, B.~Yi, R.~Mart{\'\i}n-Mart{\'\i}n, S.~Savarese, and J.~Bohg, ``Multimodal sensor fusion with differentiable filters,'' in \emph{2020 IEEE/RSJ International Conference on Intelligent Robots and Systems (IROS)}, IEEE.\hskip 1em plus 0.5em minus 0.4em\relax IEEE, Oct. 2020, pp. 10\,444--10\,451.

\bibitem{10.5555/897831}
G.~Welch and G.~Bishop, ``An introduction to the kalman filter,'' USA, Tech. Rep., 1995.

\bibitem{peyre2016gromov}
G.~Peyr{\'e}, M.~Cuturi, and J.~Solomon, ``Gromov-wasserstein averaging of kernel and distance matrices,'' in \emph{International conference on machine learning}.\hskip 1em plus 0.5em minus 0.4em\relax PMLR, 2016, pp. 2664--2672.

\bibitem{Gong2022}
F.~Gong, Y.~Nie, and H.~Xu, ``Gromov-wasserstein multi-modal alignment and clustering,'' in \emph{Proceedings of the 31st ACM International Conference on Information \& Knowledge Management}, ser. CIKM ’22.\hskip 1em plus 0.5em minus 0.4em\relax ACM, Oct. 2022, pp. 603--613.

\bibitem{10.5555/3454287.3454289}
J.~Lu, D.~Batra, D.~Parikh, and S.~Lee, \emph{ViLBERT: pretraining task-agnostic visiolinguistic representations for vision-and-language tasks}.\hskip 1em plus 0.5em minus 0.4em\relax Red Hook, NY, USA: Curran Associates Inc., 2019.

\bibitem{Li2019}
L.~H. Li, M.~Yatskar, D.~Yin, C.-J. Hsieh, and K.-W. Chang, ``Visualbert: A simple and performant baseline for vision and language,'' arXiv preprint arXiv:1908.03557, 2019.

\bibitem{pmlr-v139-jia21b}
C.~Jia, Y.~Yang, Y.~Xia, Y.-T. Chen, Z.~Parekh, H.~Pham, Q.~Le, Y.-H. Sung, Z.~Li, and T.~Duerig, ``Scaling up visual and vision-language representation learning with noisy text supervision,'' in \emph{Proceedings of the 38th International Conference on Machine Learning}, ser. Proceedings of Machine Learning Research, M.~Meila and T.~Zhang, Eds., vol. 139.\hskip 1em plus 0.5em minus 0.4em\relax PMLR, 18--24 Jul 2021, pp. 4904--4916.

\bibitem{tian2020contrastive}
Y.~Tian, D.~Krishnan, and P.~Isola, ``Contrastive multiview coding,'' in \emph{European conference on computer vision}.\hskip 1em plus 0.5em minus 0.4em\relax Springer, 2020, pp. 776--794.

\bibitem{akbari2021vatt}
H.~Akbari, L.~Yuan, R.~Qian, W.-H. Chuang, S.-F. Chang, Y.~Cui, and B.~Gong, ``Vatt: Transformers for multimodal self-supervised learning from raw video, audio and text,'' \emph{Advances in neural information processing systems}, vol.~34, pp. 24\,206--24\,221, 2021.

\bibitem{alayrac2020self}
J.-B. Alayrac, A.~Recasens, R.~Schneider, R.~Arandjelovi{\'c}, J.~Ramapuram, J.~De~Fauw, L.~Smaira, S.~Dieleman, and A.~Zisserman, ``Self-supervised multimodal versatile networks,'' \emph{Advances in neural information processing systems}, vol.~33, pp. 25--37, 2020.

\bibitem{chen2021multimodal}
B.~Chen, A.~Rouditchenko, K.~Duarte, H.~Kuehne, S.~Thomas, A.~Boggust, R.~Panda, B.~Kingsbury, R.~Feris, D.~Harwath \emph{et~al.}, ``Multimodal clustering networks for self-supervised learning from unlabeled videos,'' in \emph{Proceedings of the IEEE/CVF international conference on computer vision}, 2021, pp. 8012--8021.

\bibitem{liu2024valor}
J.~Liu, S.~Chen, X.~He, L.~Guo, X.~Zhu, W.~Wang, and J.~Tang, ``Valor: Vision-audio-language omni-perception pretraining model and dataset,'' \emph{IEEE Transactions on Pattern Analysis and Machine Intelligence}, 2024.

\bibitem{huang2023clover}
J.~Huang, Y.~Li, J.~Feng, X.~Wu, X.~Sun, and R.~Ji, ``Clover: Towards a unified video-language alignment and fusion model,'' in \emph{Proceedings of the IEEE/CVF Conference on Computer Vision and Pattern Recognition}, 2023, pp. 14\,856--14\,866.

\bibitem{mai2022hybrid}
S.~Mai, Y.~Zeng, S.~Zheng, and H.~Hu, ``Hybrid contrastive learning of tri-modal representation for multimodal sentiment analysis,'' \emph{IEEE Transactions on Affective Computing}, vol.~14, no.~3, pp. 2276--2289, 2022.

\bibitem{moon2022imu2clip}
S.~Moon, A.~Madotto, Z.~Lin, A.~Dirafzoon, A.~Saraf, A.~Bearman, and B.~Damavandi, ``Imu2clip: Multimodal contrastive learning for imu motion sensors from egocentric videos and text,'' \emph{arXiv preprint arXiv:2210.14395}, 2022.

\bibitem{shvetsova2022everything}
N.~Shvetsova, B.~Chen, A.~Rouditchenko, S.~Thomas, B.~Kingsbury, R.~S. Feris, D.~Harwath, J.~Glass, and H.~Kuehne, ``Everything at once-multi-modal fusion transformer for video retrieval,'' in \emph{Proceedings of the ieee/cvf conference on computer vision and pattern recognition}, 2022, pp. 20\,020--20\,029.

\bibitem{xue2022clip}
H.~Xue, Y.~Sun, B.~Liu, J.~Fu, R.~Song, H.~Li, and J.~Luo, ``Clip-vip: Adapting pre-trained image-text model to video-language representation alignment,'' \emph{arXiv preprint arXiv:2209.06430}, 2022.

\bibitem{guzhov2022audioclip}
A.~Guzhov, F.~Raue, J.~Hees, and A.~Dengel, ``Audioclip: Extending clip to image, text and audio,'' in \emph{ICASSP 2022-2022 IEEE International Conference on Acoustics, Speech and Signal Processing (ICASSP)}.\hskip 1em plus 0.5em minus 0.4em\relax IEEE, 2022, pp. 976--980.

\bibitem{saporta2024contrasting}
A.~Saporta, A.~M. Puli, M.~Goldstein, and R.~Ranganath, ``Contrasting with symile: Simple model-agnostic representation learning for unlimited modalities,'' \emph{Advances in Neural Information Processing Systems}, vol.~37, pp. 56\,919--56\,957, 2024.

\bibitem{cicchetti2024gramian}
G.~Cicchetti, E.~Grassucci, L.~Sigillo, and D.~Comminiello, ``Gramian multimodal representation learning and alignment,'' \emph{arXiv preprint arXiv:2412.11959}, 2024.

\bibitem{dufumier2024align}
B.~Dufumier, J.~Castillo-Navarro, D.~Tuia, and J.-P. Thiran, ``What to align in multimodal contrastive learning?'' \emph{arXiv preprint arXiv:2409.07402}, 2024.

\bibitem{Cicchetti2025}
G.~Cicchetti, E.~Grassucci, and D.~Comminiello, ``A triangle enables multimodal alignment beyond cosine similarity,'' arXiv preprint arXiv:2509.24734, 2025.

\bibitem{Zhu2024}
S.~Zhu and D.~Luo, \emph{Enhancing Multi-modal Contrastive Learning via Optimal Transport-Based Consistent Modality Alignment}.\hskip 1em plus 0.5em minus 0.4em\relax Springer Nature Singapore, Nov. 2024, pp. 157--171.

\bibitem{Shridhar2020}
M.~Shridhar, J.~Thomason, D.~Gordon, Y.~Bisk, W.~Han, R.~Mottaghi, L.~Zettlemoyer, and D.~Fox, ``Alfred: A benchmark for interpreting grounded instructions for everyday tasks,'' in \emph{2020 IEEE/CVF Conference on Computer Vision and Pattern Recognition (CVPR)}.\hskip 1em plus 0.5em minus 0.4em\relax IEEE, Jun. 2020.

\bibitem{Brohan2022}
A.~Brohan \emph{et~al.}, ``{RT-1}: Robotics transformer for real-world control at scale,'' arXiv preprint arXiv:2212.06817, 2022.

\bibitem{10.5555/3618408.3618748}
D.~Driess \emph{et~al.}, ``{PaLM-E}: An embodied multimodal language model,'' in \emph{Proceedings of the 40th International Conference on Machine Learning}, ser. Proceedings of Machine Learning Research, vol. 202.\hskip 1em plus 0.5em minus 0.4em\relax PMLR, 23--29 Jul 2023, pp. 8469--8488.

\bibitem{Kim2024}
M.~J. Kim \emph{et~al.}, ``{OpenVLA}: An open-source vision-language-action model,'' arXiv preprint arXiv:2406.09246, 2024.

\bibitem{octo_2023}
D.~Ghosh \emph{et~al.}, ``{Octo}: An open-source generalist robot policy,'' in \emph{Proceedings of Robotics: Science and Systems}, Delft, Netherlands, July 2024.

\bibitem{Intelligence2025}
{Physical Intelligence}, K.~Black \emph{et~al.}, ``{$\pi_{0.5}$}: A vision-language-action model with open-world generalization,'' arXiv preprint arXiv:2504.16054, 2025.

\bibitem{wojcik2024case}
J.~Wojcik, J.~Jiang, J.~Wu, and S.~Luo, ``A case study on visual-audio-tactile cross-modal retrieval,'' in \emph{2024 IEEE/RSJ International Conference on Intelligent Robots and Systems (IROS)}.\hskip 1em plus 0.5em minus 0.4em\relax IEEE, 2024, pp. 12\,472--12\,478.

\bibitem{Dutta2024}
A.~Dutta, E.~Burdet, and M.~Kaboli, ``Visuo-tactile based predictive cross modal perception for object exploration in robotics,'' in \emph{2024 IEEE International Symposium on Robotic and Sensors Environments (ROSE)}.\hskip 1em plus 0.5em minus 0.4em\relax Chemnitz, Germany: IEEE, 2024.

\bibitem{pmlr-v155-zambelli21a}
M.~Zambelli, Y.~Aytar, F.~Visin, Y.~Zhou, and R.~Hadsell, ``Learning rich touch representations through cross-modal self-supervision,'' in \emph{Proceedings of the 2020 Conference on Robot Learning}, ser. Proceedings of Machine Learning Research, J.~Kober, F.~Ramos, and C.~Tomlin, Eds., vol. 155.\hskip 1em plus 0.5em minus 0.4em\relax PMLR, 16--18 Nov 2021, pp. 1415--1425.

\bibitem{villani2008optimal}
C.~Villani \emph{et~al.}, \emph{Optimal transport: old and new}.\hskip 1em plus 0.5em minus 0.4em\relax Springer, 2008, vol. 338.

\bibitem{korotin2022neural}
A.~Korotin, D.~Selikhanovych, and E.~Burnaev, ``Neural optimal transport,'' \emph{arXiv preprint arXiv:2201.12220}, 2022.

\bibitem{korotin2022kernel}
A.~Korotin, D.~Selikhanovych, and E.~Burnaev, ``Kernel neural optimal transport,'' \emph{arXiv preprint arXiv:2205.15269}, 2022.

\bibitem{arjovsky2017wasserstein}
M.~Arjovsky, S.~Chintala, and L.~Bottou, ``Wasserstein generative adversarial networks,'' in \emph{International conference on machine learning}.\hskip 1em plus 0.5em minus 0.4em\relax PMLR, 2017, pp. 214--223.

\bibitem{Mikalsen2018}
K.~{\O}. Mikalsen, F.~M. Bianchi, C.~Soguero-Ruiz, and R.~Jenssen, ``Time series cluster kernel for learning similarities between multivariate time series with missing data,'' \emph{Pattern Recognition}, vol.~76, pp. 569--581, Apr. 2018.

\bibitem{Liang2021}
P.~P. Liang, Y.~Lyu, X.~Fan, Z.~Wu, Y.~Cheng, J.~Wu, L.~Chen, P.~Wu, M.~A. Lee, Y.~Zhu, R.~Salakhutdinov, and L.-P. Morency, ``Multibench: Multiscale benchmarks for multimodal representation learning,'' arXiv preprint arXiv:2107.07502, 2021.

\bibitem{Todorov2012}
E.~Todorov, T.~Erez, and Y.~Tassa, ``Mujoco: A physics engine for model-based control,'' in \emph{2012 IEEE/RSJ International Conference on Intelligent Robots and Systems}.\hskip 1em plus 0.5em minus 0.4em\relax IEEE, Oct. 2012.

\bibitem{gao2022objectfolder}
R.~Gao, Z.~Si, Y.-Y. Chang, S.~Clarke, J.~Bohg, L.~Fei-Fei, W.~Yuan, and J.~Wu, ``Objectfolder 2.0: A multisensory object dataset for sim2real transfer,'' in \emph{Proceedings of the IEEE/CVF conference on computer vision and pattern recognition}, 2022, pp. 10\,598--10\,608.

\bibitem{NVIDIA2025}
{NVIDIA}, M.~Mittal, P.~Roth \emph{et~al.}, ``Isaac lab: A gpu-accelerated simulation framework for multi-modal robot learning,'' 2025.

\bibitem{Schulman2017}
J.~Schulman, F.~Wolski, P.~Dhariwal, A.~Radford, and O.~Klimov, ``Proximal policy optimization algorithms,'' 2017, arXiv:1707.06347.

\end{thebibliography}

\begin{thebibliography}{10}
\providecommand{\url}[1]{#1}
\csname url@samestyle\endcsname
\providecommand{\newblock}{\relax}
\providecommand{\bibinfo}[2]{#2}
\providecommand{\BIBentrySTDinterwordspacing}{\spaceskip=0pt\relax}
\providecommand{\BIBentryALTinterwordstretchfactor}{4}
\providecommand{\BIBentryALTinterwordspacing}{\spaceskip=\fontdimen2\font plus
\BIBentryALTinterwordstretchfactor\fontdimen3\font minus \fontdimen4\font\relax}
\providecommand{\BIBforeignlanguage}[2]{{%
\expandafter\ifx\csname l@#1\endcsname\relax
\typeout{** WARNING: IEEEtran.bst: No hyphenation pattern has been}%
\typeout{** loaded for the language `#1'. Using the pattern for}%
\typeout{** the default language instead.}%
\else
\language=\csname l@#1\endcsname
\fi
#2}}
\providecommand{\BIBdecl}{\relax}
\BIBdecl

\bibitem{supp:Mikalsen2018}
K.~{\O}. Mikalsen, F.~M. Bianchi, C.~Soguero-Ruiz, and R.~Jenssen, ``Time series cluster kernel for learning similarities between multivariate time series with missing data,'' \emph{Pattern Recognition}, vol.~76, pp. 569--581, Apr. 2018.

\bibitem{supp:Dosovitskiy2020}
A.~Dosovitskiy, L.~Beyer, A.~Kolesnikov, D.~Weissenborn, X.~Zhai, T.~Unterthiner, M.~Dehghani, M.~Minderer, G.~Heigold, S.~Gelly, J.~Uszkoreit, and N.~Houlsby, ``An image is worth 16x16 words: Transformers for image recognition at scale,'' 2020.

\bibitem{supp:gong21ast}
Y.~Gong, Y.-A. Chung, and J.~Glass, ``{AST}: Audio spectrogram transformer,'' in \emph{Proc. Interspeech 2021}, 2021, pp. 571--575.

\bibitem{supp:JMLR:v22:20-451}
\BIBentryALTinterwordspacing
R.~Flamary, N.~Courty, A.~Gramfort, M.~Z. Alaya, A.~Boisbunon, S.~Chambon, L.~Chapel, A.~Corenflos, K.~Fatras, N.~Fournier, L.~Gautheron, N.~T. Gayraud, H.~Janati, A.~Rakotomamonjy, I.~Redko, A.~Rolet, A.~Schutz, V.~Seguy, D.~J. Sutherland, R.~Tavenard, A.~Tong, and T.~Vayer, ``Pot: Python optimal transport,'' \emph{Journal of Machine Learning Research}, vol.~22, no.~78, pp. 1--8, 2021. [Online]. Available: \url{http://jmlr.org/papers/v22/20-451.html}
\BIBentrySTDinterwordspacing

\bibitem{supp:Bonneel2011}
N.~Bonneel, M.~van~de Panne, S.~Paris, and W.~Heidrich, ``Displacement interpolation using lagrangian mass transport,'' \emph{ACM Transactions on Graphics}, vol.~30, no.~6, pp. 1--12, Dec. 2011.

\bibitem{supp:Gong2022}
F.~Gong, Y.~Nie, and H.~Xu, ``Gromov-wasserstein multi-modal alignment and clustering,'' in \emph{Proceedings of the 31st ACM International Conference on Information \& Knowledge Management}, ser. CIKM ’22.\hskip 1em plus 0.5em minus 0.4em\relax ACM, Oct. 2022, pp. 603--613.

\bibitem{supp:10.5555/2981345.2981371}
D.~Barber and F.~Agakov, ``The im algorithm: a variational approach to information maximization,'' in \emph{Proceedings of the 17th International Conference on Neural Information Processing Systems}, ser. NIPS'03.\hskip 1em plus 0.5em minus 0.4em\relax Cambridge, MA, USA: MIT Press, 2003, p. 201–208.

\bibitem{supp:poole2019variational}
B.~Poole, S.~Ozair, A.~Van Den~Oord, A.~Alemi, and G.~Tucker, ``On variational bounds of mutual information,'' in \emph{International conference on machine learning}, 2019, pp. 5171--5180.

\bibitem{supp:oquab2024dinov2}
M.~Oquab, T.~Darcet, T.~Moutakanni, H.~Vo, M.~Szafraniec, V.~Khalidov, P.~Fernandez, D.~Haziza, F.~Massa, A.~El-Nouby \emph{et~al.}, ``{DINOv2}: Learning robust visual features without supervision,'' \emph{Transactions on Machine Learning Research}, 2024.

\bibitem{supp:he2016deep}
K.~He, X.~Zhang, S.~Ren, and J.~Sun, ``Deep residual learning for image recognition,'' in \emph{Proceedings of the IEEE Conference on Computer Vision and Pattern Recognition}, 2016, pp. 770--778.

\bibitem{supp:higuera2024sparsh}
C.~Higuera, A.~Sharma, C.~K. Bodduluri, T.~Fan, P.~Lancaster, M.~Kalakrishnan, M.~Kaess, B.~Boots, M.~Lambeta, T.~Wu, and M.~Mukadam, ``Sparsh: Self-supervised touch representations for vision-based tactile sensing,'' in \emph{Proceedings of the Conference on Robot Learning (CoRL)}, 2024.

\end{thebibliography}
